\documentclass[lettersize,journal]{IEEEtran}

\usepackage{mathtools}   
\usepackage{amssymb}     
\usepackage{amsthm}      
\usepackage{amsfonts}

\usepackage{array}
\usepackage{booktabs}
\usepackage{multirow}

\usepackage{graphicx}
\usepackage{subcaption}
\usepackage{wrapfig}
\usepackage{stfloats}

\usepackage{algorithm}
\usepackage{algpseudocode}

\usepackage{textcomp}
\usepackage{verbatim}
\usepackage{microtype}
\usepackage{nicefrac}
\usepackage{xcolor}
\usepackage{hyperref}
\usepackage{url}
\usepackage{cite}

\def\BibTeX{{\rm B\kern-.05em{\sc i\kern-.025em b}\kern-.08em
		T\kern-.1667em\lower.7ex\hbox{E}\kern-.125emX}}
\usepackage{balance}
\begin{document}
	\title{Enhancing the Effectiveness and Durability of Backdoor Attacks in Federated Learning through Maximizing Task Distinction}
	\author{Zhaoxin Wang, 		
		Handing~Wang,~\IEEEmembership{Senior Member,~IEEE,}
		Cong~Tian,
		and~Yaochu~Jin,~\IEEEmembership{Fellow,~IEEE}
		\thanks{This work was supported in part by the National Natural Science Foundation of China (No. 62376202). (\textit{Corresponding author: Handing Wang})
			
			Z. Wang and H. Wang are with School of Artificial Intelligence, Xidian University, Xi'an 710071, China. (e-mail: wangzhaoxin@stu.xidian.edu.cn, hdwang@xidian.edu.cn).
			
			C. Tian is with School of Computer Science and Technology, Xidian University, Xi'an 710071, China. (e-mail: ctian@mail.xidian.edu.cn).
			
			Y. Jin is with the Trustworthy and General Artificial Intelligence Laboratory, School of Engineering, Westlake University, Hangzhou 310030, China. (Email: jinyaochu@westlake.edu.cn)
		}
	}
	
	\markboth{Journal of \LaTeX\ Class Files,~Vol.~18, No.~9, September~2020}%
	{How to Use the IEEEtran \LaTeX \ Templates}
	
	\maketitle
	
	\begin{abstract}
		Federated learning allows multiple participants to collaboratively train a central model without sharing their private data. However, this distributed nature also exposes new attack surfaces. In particular, backdoor attacks allow attackers to implant malicious behaviors into the global model while maintaining high accuracy on benign inputs. Existing attacks usually rely on fixed patterns or adversarial perturbations as triggers, which tightly couple the main and backdoor tasks. This coupling makes them vulnerable to dilution by honest updates and limits their persistence under federated defenses. In this work, we propose an approach to decouple the backdoor task from the main task by dynamically optimizing the backdoor trigger within a min-max framework. The inner layer maximizes the performance gap between poisoned and benign samples, ensuring that the contributions of benign users have minimal impact on the backdoor. The outer process injects the adaptive triggers into the local model. We evaluate our method on both computer vision and natural language tasks, and compare it with six backdoor attack methods under six defense algorithms. Experimental results show that our method achieves good attack performance and can be easily integrated into existing backdoor attack techniques. 
	\end{abstract}


	\section{Introduction} \label{Introduction}	
	\IEEEPARstart{F}{ederated} learning (FL)~\cite{mcmahan2017communication} has emerged as a promising paradigm for privacy-preserving machine learning \cite{xue2023differentially,chen2023privacy, yazdinejad2024robust}, enabling a central model to be collaboratively trained across distributed clients without direct data sharing. In each communication round, selected participants train local models on their private datasets and upload updates to a central server for aggregation~\cite{konevcny2016federated, mcmahan2016federated, yang2019federated}. While this framework reduces the risks of raw data leakage, it simultaneously enlarges the attack surface of the learning system, introducing both privacy and security threats~\cite{lyu2020threats, kairouz2021advances, rodriguez2023survey}. 
		
	Among the various threats~\cite{gu2017badnets, szegedy2013intriguing, shokri2017membership, liu2025token, zhu2019deep}, backdoor attacks are particularly concerning~\cite{gu2017badnets, he2023backdoor, fang2023vulnerability}. Unlike evasion attacks that occur at inference time, backdoor attacks poison the training process. Compromised participants inject malicious patterns into their local updates, which then propagate into the global model through aggregation. As the server cannot inspect client data and secure aggregation protocols~\cite{cramer2015secure, bonawitz2017practical, gong2023multi}, it further obfuscates local updates, making it extremely challenging to detect such malicious modifications. A successfully backdoored model behaves normally on benign inputs but produces malicious outputs on inputs containing specific triggers, posing severe risks in practical applications such as healthcare diagnostics, autonomous driving, or financial decision-making.  

    Early studies established the feasibility of backdoor insertion in FL. BadNets demonstrated the basic concept of trigger-based poisoning in centralized settings~\cite{gu2017badnets}, and Bagdasaryan \textit{et al.} adapted this threat to FL, showing that semantic backdoors can be particularly effective~\cite{bagdasaryan2020backdoor}. Subsequent work explored distributed trigger decomposition~\cite{xie2019dba} and edge-case triggers that induce misclassification~\cite{wang2020attack}. However, many of these attacks require either a substantial proportion of malicious clients or explicit scaling of submitted updates to dominate aggregation. The assumptions make backdoor fragile in the presence of robust aggregation rules and clipping defenses~\cite{blanchard2017machine,pillutla2022robust,sun2019can,nguyen2022flame,gong2023agramplifier, zhang2024flpurifier}. 
	
    Recent works begin to address the durability of backdoors under constrained attacker capabilities. For example, Neurotoxin~\cite{zhang2022neurotoxin} targets parameters that are updated infrequently to increase longevity, IBA~\cite{nguyen2024iba} uses generative strategies for more adaptive triggers, and A3FL~\cite{zhang2024a3fl} considers the discrepancies between the local and global models to implant more effective backdoors. While these efforts extend attack durability, they largely rely on fixed-pattern triggers or adversarial-perturbation mechanisms that do not explicitly optimize for separation between the benign and backdoor objectives. Intuitively, a trigger that behaves like an adversarial perturbation may merely push the model across decision boundaries, but does not guarantee the formation of an independent backdoor objective that resists dilution by subsequent benign updates.
	
	The current research has taken steps toward durable attacks, but important aspects remain underexplored. In particular, prior studies less explored two critical properties. The first is durability in a long-term sense, whether the implanted backdoor continues to be effective after malicious clients cease contributing updates. The second is integrability, whether an attack mechanism can be combined with other strategies to amplify its effectiveness. Overlooking these properties risks underestimating the threats posed by backdoors in FL and may give a false sense of security regarding the effectiveness of existing defenses.
	
	In this work, we propose enhancing the \textbf{E}ffectiveness and \textbf{D}urability of \textbf{B}ackdoor \textbf{A}ttacks (EDBA) through maximizing the distinction between the main and backdoor tasks. This maximization ensures a clear separation between the main and backdoor tasks, preventing the weights submitted by other normal participants from influencing the backdoor task, thereby enhancing the ASR and the durability of backdoors. Specifically, we establish a dynamic trigger optimization mechanism, as the optimal trigger pattern that can separate the backdoor and the main task changes as the model updates. This mechanism can be formulated within a min-max framework, where the trigger pattern is updated during the inner maximization process and the backdoor trigger is implanted in the outer minimization phase. In summary, our contributions are:

	\begin{itemize}
		\item We propose dynamically optimizing the trigger pattern within a min-max backdoor attack framework where the maximization phase focuses on trigger generation to enhance the distinction between the backdoor and the main task. The minimization phase aims at backdoor injection, employing the generated trigger to train a backdoored local model.
		\item We employ gradient ascent to design triggers that effectively separate the main and backdoor tasks. In computer vision tasks, we directly optimize trigger values to minimize the cosine similarity between the model outputs for poisoned and clean samples, while in natural language processing tasks, we focus on optimizing the trigger patterns.
		\item Experimental results demonstrate that our backdoor attack achieves a high ASR and durability while maintaining the main task accuracy, and the proposed dynamic trigger generation method can be easily integrated into the existing backdoor attack methods.
	\end{itemize}

    This paper proceeds as follows: Section~\ref{sec:related_work} reviews the related literature. Section~\ref{sec:threat_model} formalizes the threat model and attacker assumptions. Section~\ref{sec:method} presents the EDBA framework and its instantiations for vision and NLP tasks. Section~\ref{experiment} details experimental setup and results, including robustness and durability evaluations. Section~\ref{sec:discussion} discusses security implications and potential defenses. Finally, Section~\ref{conclusion} concludes with limitations and future work.

	\section{Related Work} \label{sec:related_work}
	\paragraph{Federated Learning.}
	Federated learning, as a decentralized learning method, ensures that participants collaboratively train a joint model safety and efficiency~\cite{li2023federated, tan2022towards, karimireddy2020scaffold, zhu2019multi}. Generally, the FL training framework follows three main steps:
	
	1. Model Distribution: The central server selects a subset of participants $S \subset {1,2, \ldots, N}$ for the current communication, and distributes the current global model $G^t$ to the selected participants $S$.
	
	2. Local Model Training: The selected participants $i \subset S$ train their local models $L_i^{t+1}$ using their own data $D_i$. After that, they upload their updated model parameters or gradients $L_i^{t+1}-G^t$ to the server.
	
	3. Model Aggregation: The server uses aggregation algorithms to update the global model with the gradients or parameters submitted by the participants, as in FedAvg~\cite{mcmahan2017communication}, where:
	\begin{equation}
		G^{t+1}=G^t+\frac{1}{|S|} \sum_{i \subset S}\left(L_i^{t+1}-G^t\right),
	\end{equation}
	where $|S|$ represents the number of selected participants.
	
	\paragraph{Backdoor Attacks in FL.}
	Backdoor attacks in FL involve attackers uploading malicious parameters to poison the central global model~\cite{tolpegin2020data, bagdasaryan2020backdoor, wang2020attack, baruch2019little, lyu2023poisoning, fang2023vulnerability}. The compromised model performs well on benign samples but follows the attackers' intentions when it processes inputs with triggers. BadNets~\cite{gu2017badnets} first demonstrates injecting a specific pixel pattern trigger during the training process can easily backdoor the deep neural networks. Subsequently, Bagdasaryan \textit{et al.}~\cite{bagdasaryan2020backdoor} show that the global model can inherit these poisoned parameters through the aggregation process in FL. They further suggest using semantic backdoors instead of pixel pattern backdoors and scaling the submitted model parameters to increase the backdoor ASR of backdoor attacks in FL. Neuroxin~\cite{zhang2022neurotoxin} extends the duration of backdoor attacks by injecting backdoor tasks into the model parameters with minimal updates. CerP~\cite{lyu2023poisoning} optimize the trigger and the model weights with $L_2$ regularization to minimize the bias. IBA~\cite{nguyen2024iba} employs adversarial perturbations as triggers and selectively poisons specific neurons to preserve the attack's efficacy. While these variants significantly enhance backdoor attacks, they fail to devise a trigger pattern that can effectively separate the main and the backdoor tasks. 
	
	\paragraph{Defense in FL.}
	Defense strategies in FL aim to eliminate the impact of malicious attackers, and these defenses can implemented during various phases of FL~\cite{lyu2022privacy}. Before the aggregation phase, implementing some detecting defense algorithms is challenging as the FL server does not have access to local private data~\cite{huang2019neuroninspect, hou2021mitigating, nasr2018comprehensive}. During the aggregation process, defenses~\cite{liu2021federaser, rieger2022deepsight, fung2018mitigating, nguyen2022flame, panda2022sparsefed, fereidooni2023freqfed} focus on reducing the influence of potential attackers. NDC~\cite{sun2019can} employs a norm clipping to limit large model updates, mitigating the impact of attackers uploading scaled malicious parameter weights. Krum~\cite{blanchard2017machine} calculates the Euclidean distance between the uploaded weights and selects the smallest one for updating the global model. Similarly, RFA~\cite{pillutla2022robust} aggregates local models using their geometric median. The defenses after the aggregation phase typically operate by identifying and removing potential backdoors in the model. However, a limitation of this approach is that the central server requires access to some training data to implement these defenses~\cite{wang2019neural, liu2018fine,li2024backdoorindicator}.

	\section{Threat Model} \label{sec:threat_model}
	\subsection{Attacker Ability and Knowledge} 
	\textbf{Attacker Ability.}
	We follow the assumptions in the previous work~\cite{bagdasaryan2020backdoor, xie2019dba, zhang2024a3fl, nguyen2024iba}, where attackers have complete control over certain malicious participants. Specifically, attackers can access the training data of those compromised participants and manipulate their training hyperparameters, such as the learning rate and the number of local training epochs. The attacker has bounded compute and communication resources per client (no unrealistic unlimited scaling). 
	
	\textbf{Knowledge.} We consider a black-box setting. Adversaries do not know the FL protocol (client selection, aggregation rule), and they do not have access to other clients' private data. We assume the adversaries are unaware of the exact internal defenses deployed on the server (e.g., clipping thresholds), though we evaluate robustness under a range of plausible defenses.

    \subsection{Adversary Objectives.}
    The primary objective of attackers is to inject backdoors into the central global model, ensuring that the model behaves as the attackers' intentions for any inputs containing specific triggers, while maintaining good performance on benign inputs, $i.e.$, high accuracy on both the backdoor and the main task. Given the expected backdoor output $P$, a successful backdoored model parameters $w_i$ follows:
    \begin{equation}
        \small
        w_i^* = \arg\max_{w_i}\sum_{j \in D_{p}^i} \mathbb{I}\left(G^{t}(x_j^i) = P\right) + \sum_{j \in D_{c}^i} \mathbb{I}\left(G^{t}(x_j^i) = y_j\right),
        \small
    \end{equation}
    where  $\mathbb{I}$ represents an indicator function that is equal to 1 when a certain condition is true and 0 otherwise, $x$ denotes the training data, $y$ represents its corresponding label, $D_{p}$ represents the poisoned dataset, $D_{c}$ represents the clean dataset. Here, $D_{p}^i \cup D_{c}^i=D_i$. Besides the high ASR of the backdoors, attackers also focus on the durability of these backdoors, meaning that the malicious modifications should persist in the model even if the compromised participants cease uploading malicious parameters.
	
	\section{Methodology} \label{sec:method}
	The existing backdoor attacks typically employ fixed trigger patterns or leverage adversarial perturbations to generate triggers. However, neither method can effectively separate the backdoor task from the main task. Achieving this separation requires that, regardless of the neural network’s predictions for the given sample, the poisoned sample should produce a distinct outcome rather than merely an incorrect output similar to an adversarial example.
	
	In this work, we propose a dynamic min-max backdoor attack framework. Specifically, we dynamically optimize the backdoor trigger, regarding that the optimal trigger pattern may differ across various training phases. Within this min-max framework, we optimize the trigger in the inner maximization process, and then inject the backdoor into the local model in the outer layer. To better illustrate our attack framework, we first introduce the threat model, followed by the processes of trigger generation in computer vision and natural language processing tasks, and backdoor injection. 
	
	\subsection{Trigger Generation on Computer Vision Tasks}
	Unlike other backdoor attacks, which typically employ static trigger patterns~\cite{gu2017badnets, bagdasaryan2020backdoor, alam2022perdoor}, our approach advocates that triggers should be dynamically updated as the FL process progresses.  For the currently calculated optimal trigger pattern $T^*$ for separating the main task and the backdoor task, if the trigger changes significantly compared to the previous $T$, we update the backdoor trigger when:
	\begin{equation}
		\label{convergence}
		\|   T^* - T        \| \geq \epsilon,
	\end{equation}
	where $\epsilon$ is the convergence condition.
	
	In addition, an effective backdoor trigger should make the poisoned sample and the clean sample produce completely different predictions on the target model even if the model cannot correctly classify the original sample. Therefore, compared with using adversarial perturbations to make triggers that make the sample misclassified. The formulation of the EDBA trigger optimization problem is as follows:
	\begin{equation}
		\label{Trigger}
		T^* = \arg \max_{T}\sum_{(x,y)\sim D} d\left(f_{\theta}(x+T),f_{\theta}(x)\right),
	\end{equation}
	where $x$ represents the input image data, $y$ is the corresponding label, $T$ denotes the dynamically generated image trigger, $f_{\theta}(x)$ indicates the logits output of the deep neural network, and $d$ is the distance metric. 
	
	This formulation aims to create a distinct separation between the behavior of the main task and that induced by the backdoor, enhancing the efficacy of the backdoor under the federated setting. We use cosine similarity as the distance metric to generate the trigger $T$ in Eq.(\ref{Trigger}). The updating mechanism can be expressed as follows:
	\begin{equation}
		\label{Trigger_generation}
		\begin{gathered}
			T^{t+1} = T^t - \alpha \cdot  \nabla_T L_{\text{cos}}(m_p, m_b) , \\[1mm]
			m_p = f_{\theta}(x + T^t),  \\[1mm]
			m_b = f_{\theta}(x),
		\end{gathered}
	\end{equation}
	where $\alpha$ is the learning rate for the trigger, the $\nabla_T$ represents the gradient of trigger $T$ and $L_{cos}$ is the cosine similarity loss.

    \begin{algorithm}[!]
    \caption{Workflow of the EDBA in Computer Vision Tasks}
    \label{algorithm}
    \begin{algorithmic}[1]
        \State \textbf{Input:} Global model $G$ with parameters $\theta$, dataset $D_i$, model learning rate $\beta$, training epoch $E$, attack learning rate $\alpha$, trigger generation epoch $E_t$, previous trigger $T_{ar}$
        \State $\theta^{0} \leftarrow \theta$
        \If{the first attack}
        \State $T^0 \leftarrow U[0, 1]$ \Comment{Initialize trigger randomly}
        \Else
        \State $T^0 \leftarrow T_{ar}$ 
        \EndIf
        \For{$\{\boldsymbol{x}, \boldsymbol{y}\} \sim D_i$}
        \State $m_b = G(x)$
        \For{$t = 1$ to $E_t$}
        \State $m_p = G(x + T^{t-1})$ \Comment{Updating trigger}
        \State $T^{t} = T^{t-1} - \alpha \cdot  \nabla_T L_{\text{cos}}(m_p, m_b) $
        \EndFor
        \If{$\|T^*-T^0\|>\epsilon$}
        \State $T_{ar} = T^t$
        \EndIf
        \EndFor
        \For{epoch = 1 to $E$}
        \State \Comment{Partition the data into poisoned and clean subsets}
        \State $D_p \leftarrow \text{random\_select}(\frac{1}{10} \times |D_i|, D_i)$
        \State $D_c \leftarrow D_i - D_p$
        
        \For{$\{{x}, {y}\} \sim D_p$}
        \State $x \leftarrow x + T_{ar}$
        \State $y \leftarrow y_p$
        \EndFor
        
        \State $\theta \leftarrow \theta - \beta \nabla \rho(\theta)$  \Comment{Injecting backdoor with Eq.(\ref{training})}
        \EndFor
        \State Upload $\theta - \theta^{0}$ to the server
    \end{algorithmic}
    \end{algorithm}

    \subsection{Trigger Generation on Natural Language Processing Tasks}
    Unlike the computer vision tasks the pixel can be optimized with the gradient and directly appended to the original data as in Eq.(\ref{Trigger_generation}). In natural language processing tasks, the data is often encoded as a sequence of discrete tokens $\boldsymbol{X}=\left\{x_1, x_2, \cdots, x_n\right\}$ and the trigger replaces the original tokens as $\boldsymbol{X_{Tr}}=\left\{x_1, tigger_1, \cdots, x_n\right\}$. The trigger token cannot be optimized according to the gradient directly. Therefore, to maximize the separation between the main task and the backdoor task, it is crucial to determine the replacement pattern of the trigger tokens, i.e., the placement position within the sequence. The choice of replacement positions significantly impacts the success rate of backdoor injection. For example, a scattered replacement pattern is less likely to disrupt the original sentence's semantics, thereby preserving the accuracy of the main task, whereas a continuous token replacement pattern is more likely to alter the sentence's meaning. 
	
	We select the trigger position according to the position importance ranking. We preset the trigger length (i.e., the number of replacement tokens) and sequentially replace the original tokens with the placeholders, selecting the position with the highest score $S_{i}$ with Eq. (\ref{score}) for replacement.
	\begin{equation}
		\label{score}
		S_{i} = 1 - Cos\left(F(X),F(X_{\backslash i})\right),
	\end{equation}
	where $F({X})$ represents the prediction score, ${X}_{\backslash i}$ represents the token sequence with trigger replacement at position $i$, $S_i$ represents the importance score of position $i$. When the token at position $i$ is replaced with the placeholder, a larger $S_i$ change indicates that the current position $i$ is important.

        \begin{algorithm}[!]
    \caption{Workflow of the EDBA in Natural Language Processing Tasks}
    \label{algorithm2}
    \begin{algorithmic}[1]
        \Require Global model $G$ with parameters $\theta$, dataset $D_i$, backdoor label $Y_p$, model learning rate $\beta$, training epoch $E$, trigger length $M$, candidate position $K$
        \State $\theta^{0} \leftarrow \theta$
        
        \For{$i \gets 1$ to $K$}
        \State Calculate $S_i$ with Eq.~(\ref{score})
        \EndFor
        
        \State Position $P \leftarrow$ Top-$M$ in $\{ S_i \}_{i=1}^{K}$
        
        \State $D_p \leftarrow \text{random\_select}\,\!\bigl(\tfrac{1}{10} \times |D_i|,\, D_i\bigr)$
        \State $D_c \leftarrow D_i \setminus D_p$
        
        \For{$epoch \gets 1$ to $E$}
        \ForAll{$\{X, Y\}$ in $D_p$}
        \State $X^{Tr} \leftarrow X$ with replacement by $Triggert$ at Position $P$
        \State $Y \leftarrow Y_p$
        \EndFor
        \State $\theta \leftarrow \theta - \beta \nabla \rho(\theta)$  \Comment{Injecting backdoor with Eq.(\ref{training})}
        \EndFor
        
        \State Upload $\theta - \theta^{0}$ to the server
    \end{algorithmic}
    \end{algorithm}
	
	\subsection{Backdoor Injection} \label{backdoor_inj}
	In the backdoor injection phase, we first train a backdoored local model with the malicious participants’ private data. Subsequently, these compromised participants submit the backdoored model parameters to the central server for aggregation. Similar to other attack methods, we adopt Frobenius norm $\| \theta - \theta^g\|_F$ to constrain change between the poison backdoored model and the global model to avoid the potential weights clipping or scaling defenses, where $\theta^g$ represents the model parameter of the global model. The training process for local backdoored models can be described as:
	
	\begin{equation}
		\label{training}
		\begin{gathered}
			\min_{\theta}\,\rho(\theta), \quad \text{where } \rho(\theta)= \\[4pt]
			\frac{1}{|D^{i}|} \left(
			\sum_{j\in D^{i}_{p}} L(\theta,x^{i}_{j},y^{i}_{j})
			+\sum_{j\in D^{i}_{c}} L(\theta,x^{i}_{j},y^{i}_{j})
			\right)
			+\gamma\,\|\theta-\theta^{g}\| .
		\end{gathered}
	\end{equation}
	
	Here, $\theta$ is the parameters of the backdoored model, $|D^i|$ denotes the number of samples in training data $D$ of participant $i$, and $L$ represents the loss function, $\gamma$ is the regularization factor. The dataset $D_{c}^i$ includes the clean data samples, while the poisoned dataset $D_{p}^i$ comprises clean data samples that have been modified by embedding triggers. The union $D_{p}^i \cup D_{c}^i=D_i$ form the complete dataset $D_i$. 
	
	It is crucial to craft the poisoned dataset $D_{p}^i$, in computer vision tasks, we craft triggers with Eq.(\ref{Trigger_generation}) and attach them to the clean examples. In natural language processing tasks, we first obtain the position importance rank with Eq.(\ref{score}) and choose the trigger positions according to the scores. We select handcrafted rare words from the vocabulary as the trigger tokens to ensure the effectiveness of the backdoor. These rare words are then used to replace the original tokens at the selected positions, thereby crafting the poisoned dataset. 
	
	In summary, combined with Eq.(\ref{Trigger}) and Eq.(\ref{training}), the entire backdoor attack method can be formalized as a min-max problem:
	\begin{equation}
		\begin{gathered}
			\min_{\theta} \rho(\theta), \; 	\text{where} \;
			\rho(\theta) = \\[4pt]
            \mathbb{E}_{(x,y) \sim D}\!\Bigl[\max_{T} L_{cos}(\theta, x+T, x)\Bigr].
		\end{gathered}
	\end{equation}

	For a better understanding of the training process, the detailed description of the computer vision task is presented in Algorithm \ref{algorithm}. The natural language processing task is presented in Algorithm \ref{algorithm2}.

    \section{Experimental Results} \label{experiment}
	In this section, we present a comprehensive experimental results to evaluate the effectiveness of the proposed EDBA in comparison to other federated backdoor attack algorithms under different defense methods. We conduct experiments on image classification and semantic analysis these two tasks under two different experimental settings including fixed-pool and fixed-frequency two scenarios. Experiments are conducted on an NVIDIA RTX 4090 GPU and the code will be released upon publication.

    \begin{table*}[htbp]  \centering  \caption{Task and parameters description.}    
    \resizebox{0.8\textwidth}{!}{
        \begin{tabular}{ccccc}    
            \toprule    
            \toprule    Dataset & Model & Local learning rate/E & Poison learning rate/Ep & Poison ratio \\    
            \midrule    
            MNIST & ResNet18 & 0.01/12 & 0.05/2 & 5/64 \\    
            CIFAR10 & ResNet18 & 0.01/12 & 0.05/2 & 5/64 \\    
            CIFAR100 & ResNet18 & 0.01/12 & 0.05/2 & 10/64 \\    
            Tiny-ImageNet & ResNet18 & 0.01/12 & 0.05/2 & 20/64 \\    
            Yelp-Review & Transformer & 0.0002/2 & 0.0005/2 & 3/12 \\    
            IMDB & Transformer & 0.005/2 & 0.0001/2 & 10/64 \\    
            \bottomrule    
            \bottomrule    
        \end{tabular}% 
    }
    \label{Parameters}%
    \end{table*}%
	
	\subsection{Experimental Settings}
	\subsubsection{Datasets and Models}
	\paragraph{Computer Vision.}  For this task, we evaluate the performance of our method on MNIST~\cite{lecun1995learning}, CIFAR10, CIFAR100~\cite{AlexKrizhevsky2009LearningML} and Tiny-ImageNet~\cite{deng2009imagenet} datasets. The MNIST dataset contains 60,000 training examples and 10,000 testing examples of handwritten digits. Each of the ten digit classes contains 6000 training examples centered in a 28x28 image. The CIFAR10 dataset consists of 50,000 images across 10 classes, with 5000 images per class. CIFAR100 contains 100 classes with 500 images each class, and the image is 3 $\times$ 32 $\times$ 32. Tiny-ImageNet contains 100,000 images of 200 classes (500 for each class), and each image is 64 $\times$ 64 $\times$ 3. Our base model is ResNet18~\cite{KaimingHe2016IdentityMI}.
	\paragraph{Natural Language Processing.} For natural language processing tasks, we choose sentiment analysis to evaluate the performance of our method. IMDB \cite{maas2011learning} is a binary sentiment analysis dataset consisting of 50,000 reviews, and half of them are used to test. The Yelp reviews full star dataset~\cite{zhang2015character} consists of 650,000 training samples and 50,000 testing samples for each review star from 1 to 5. In this task, we use transformer~\cite{vaswani2017attention} as the base model, combined with the BERT pre-training paradigm~\cite{devlin2019bert} and fine-tune on the selected dataset.
	\subsubsection{Attack Scenario and Backdoor Task}
	We evaluate the algorithms' effectiveness under fixed-frequency and fixed-pool these two attack scenarios with IID and Non-IID data distribution these two federated settings. In the fixed-frequency scenario~\cite{wang2020attack}, only one compromised client participates in the training for each $f$ round, and the fixed-pool attack scenario involves a certain number of malicious attackers mixed among users, with clients randomly selected from these users for communication. We simulate heterogeneous data partitioning by Dirichlet distribution sampling~\cite{minka2000estimating} with different hyperparameter $\alpha$, which $\operatorname{Dir}_K(0.5)$ for MNIST and CIFAR, $\operatorname{Dir}_K(0.01)$ for Tiny-ImageNet.
	
	\subsubsection{Compared Methods}
	We choose BadNets~\cite{gu2017badnets}, Scaling~\cite{bagdasaryan2020backdoor}, Neuroxin~\cite{zhang2022neurotoxin}, IBA~\cite{nguyen2024iba}, Chameleon~\cite{dai2023chameleon} and A3FL~\cite{zhang2024a3fl} these six backdoor attack methods as comparison and evaluate the performance under NDC~\cite{sun2019can}, Krum~\cite{blanchard2017machine}, Multi-Krum~\cite{blanchard2017machine}, Median~\cite{yin2018byzantine}, FLAME~\cite{nguyen2022flame} and Freqfed~\cite{fereidooni2023freqfed} six defense methods.
    
    \subsubsection{Training Details}
    Following the previous work~\cite{xie2019dba, nguyen2024iba}, we utilize the Stochastic Gradient Descent (SGD) optimizer with a momentum of 0.9 and a weight decay of $5 \times 10^{-4}$ with $E$ local epochs, a local learning rate of $l_r$, and a batch size of $B$, poison ratio $r$, poison learning rate $l_p$, local training epochs $E$ and local poison training epochs $E_p$. The number of clients selected in each round is 10/200 and the trigger learning rate in Eq.(\ref{Trigger_generation}) is set to 0.1. All the training parameters including comparison attack and defense methods are summarized in Table~\ref{Parameters}.
    
    For EDBA, the trigger is dynamically optimized until convergence, where the threshold $\epsilon$ in Eq.~(\ref{convergence}) is set to $0.1 \cdot \|T^*\|_2$, the update step size $\alpha$ in Eq.~(\ref{Trigger_generation}) is 0.05, and the regularization coefficient $\gamma$ in Eq.~(\ref{training}) is 0.1. For baselines, we follow the default configurations in prior works: Scaling~\cite{bagdasaryan2020backdoor} amplifies malicious weights with a factor of 10; Neurotoxin~\cite{zhang2022neurotoxin} updates only the top-5\% of parameters per round and employs a fixed pixel trigger; A3FL~\cite{zhang2024a3fl} jointly optimizes an adversarial model with a PGD-style trigger; NDC~\cite{sun2019can} clips model updates with a norm threshold of 3; Krum~\cite{blanchard2017machine} in its multi-Krum variant selects 6 clients per round; and FLAME~\cite{nguyen2022flame} is implemented with clustering and Gaussian noise, where $\lambda=0.001$. 
    
    \subsubsection{Evaluation Metrics}
    We use the accuracy on the main task (MA) and the accuracy on the backdoor task (BA) as the primary evaluation metrics. In addition, we focus on the durability and the effectiveness of the injected backdoors. Durability refers to whether the ASR decreases as training progresses after the malicious attacker is removed. The effectiveness refers to the backdoor ASR with a fixed proportion of malicious attackers.
	
	\begin{figure}[ht]
		\centering
		\begin{subfigure}{0.23\textwidth}
			\includegraphics[width=\linewidth]{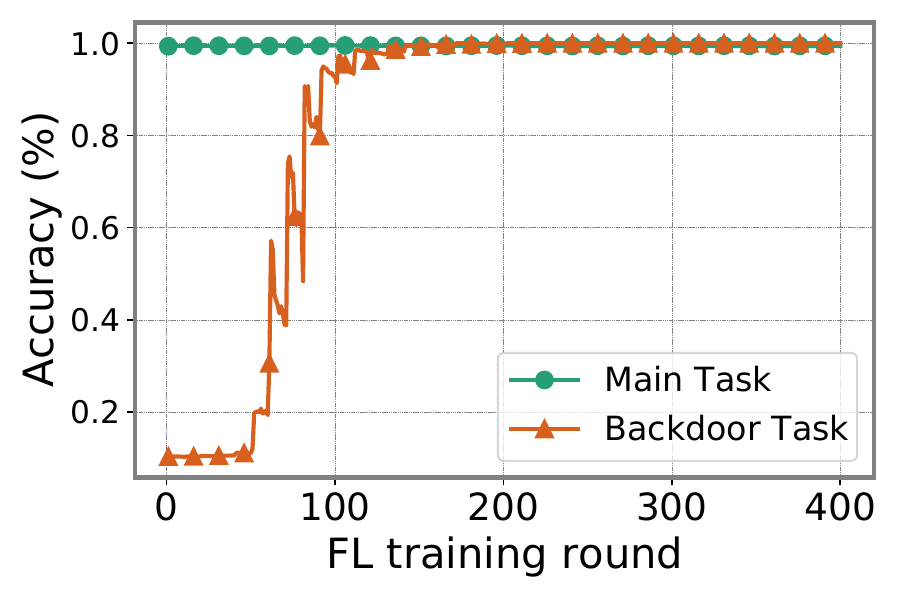}
			\caption{IID MNIST}
		\end{subfigure}
		\hfill
		\begin{subfigure}{0.23\textwidth}
			\includegraphics[width=\linewidth]{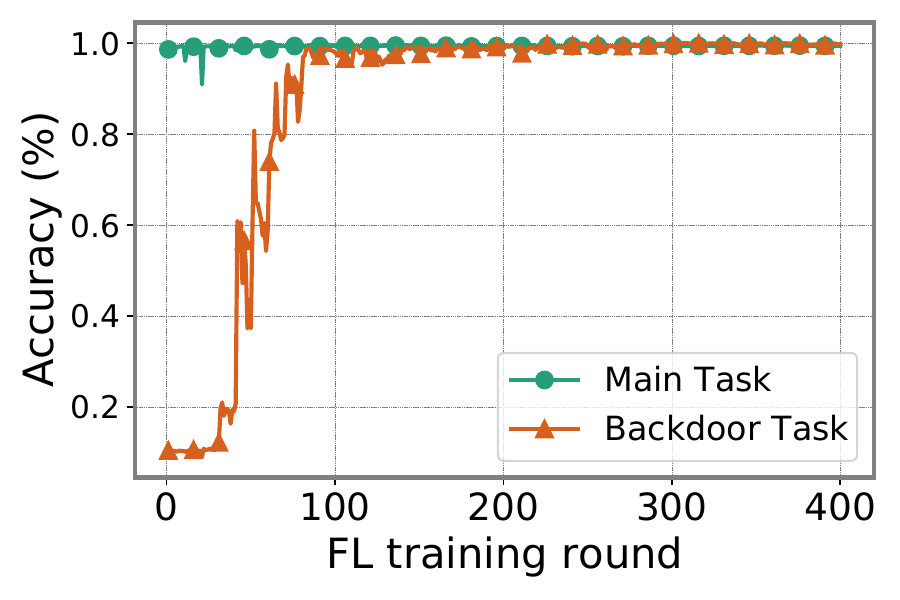}
			\caption{Non-IID MNIST}
		\end{subfigure}
		\hfill
		\begin{subfigure}{0.23\textwidth}
			\includegraphics[width=\linewidth]{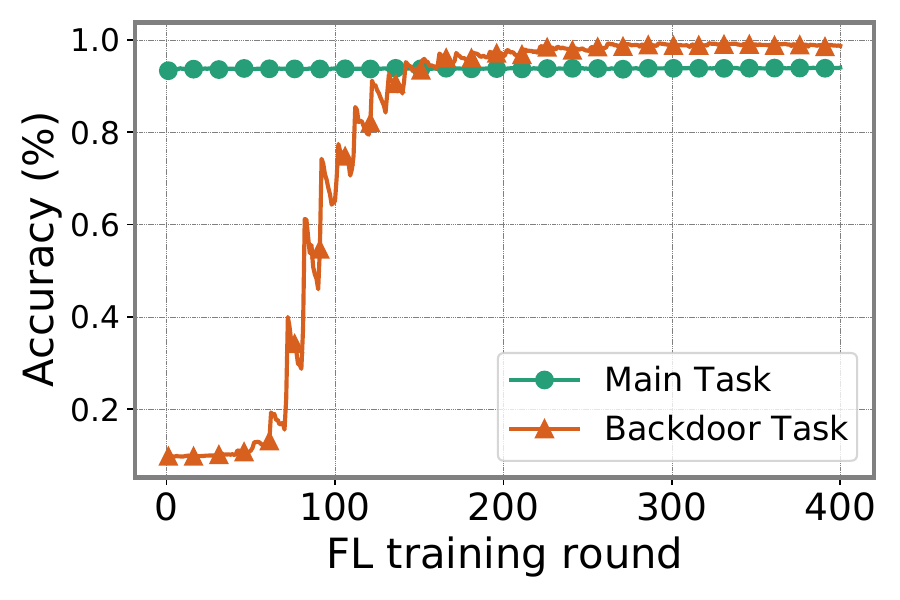}
			\caption{IID CIFAR10}
		\end{subfigure}
		\hfill
		\begin{subfigure}{0.23\textwidth}
			\includegraphics[width=\linewidth]{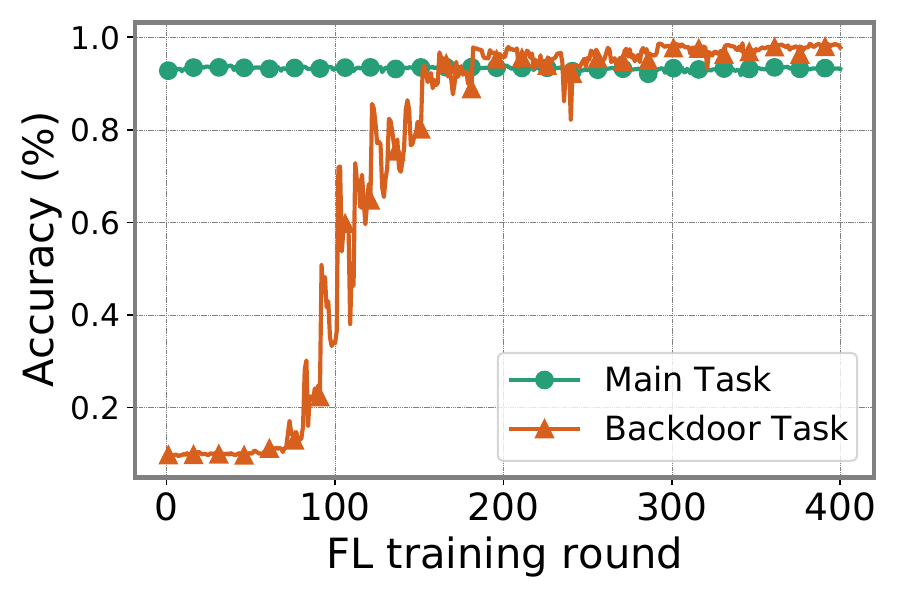}
			\caption{Non-IID CIFAR10}
		\end{subfigure}
		\begin{subfigure}{0.23\textwidth}
			\includegraphics[width=\linewidth]{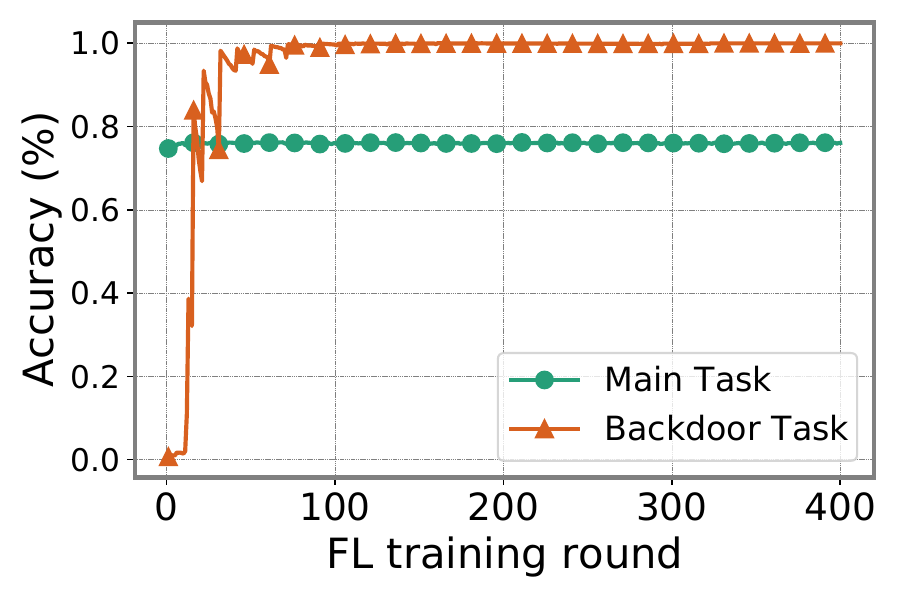}
			\caption{IID CIFAR100}
		\end{subfigure}
		\hfill
		\begin{subfigure}{0.23\textwidth}
			\includegraphics[width=\linewidth]{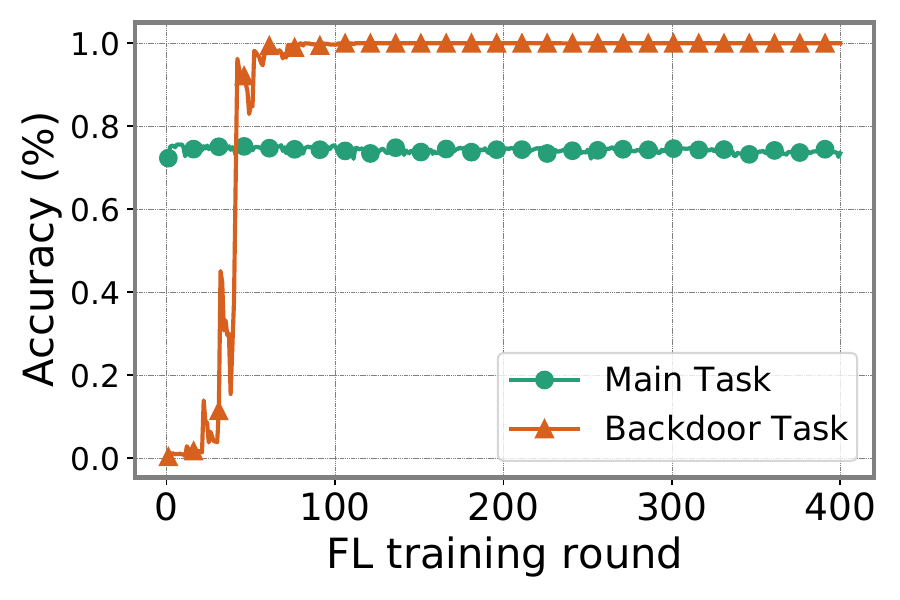}
			\caption{Non-IID CIFAR100}
		\end{subfigure}
		\hfill
		\begin{subfigure}{0.23\textwidth}
			\includegraphics[width=\linewidth]{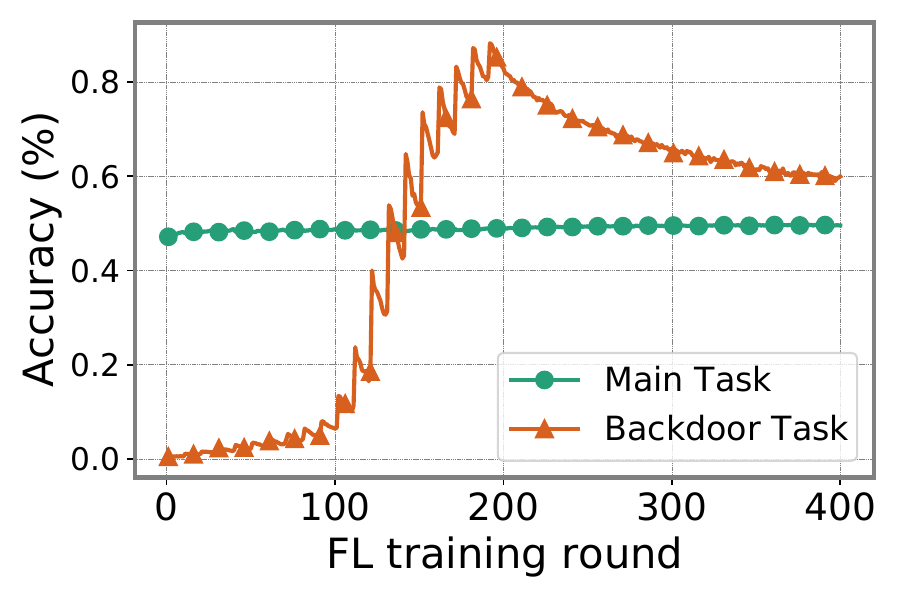}
			\caption{IID TINY}
		\end{subfigure}
		\hfill
		\begin{subfigure}{0.23\textwidth}
			\includegraphics[width=\linewidth]{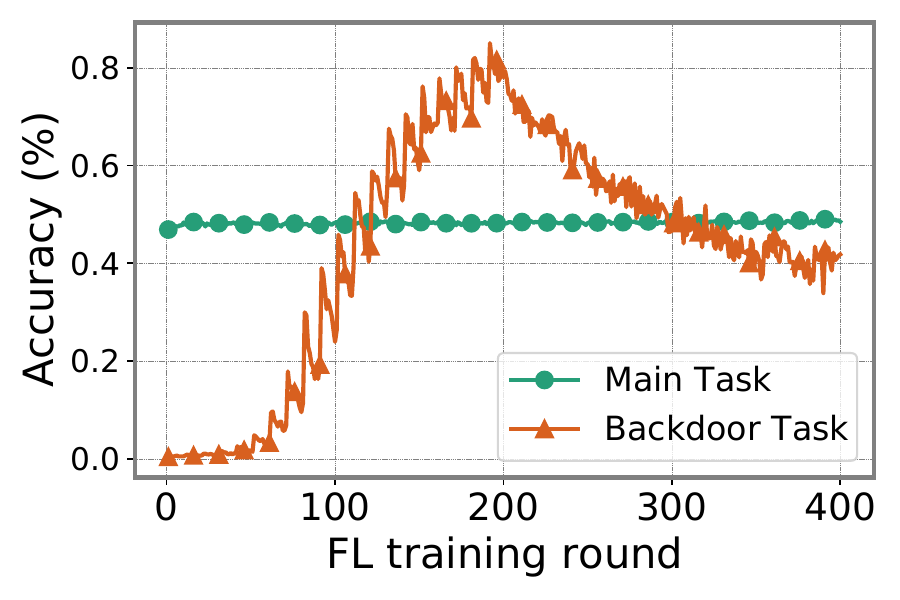}
			\caption{Non-IID TINY}
		\end{subfigure}
		\caption{Main task and backdoor task accuracy under the fixed-frequency attack scenario with Non-IID and IID setting.}
		\label{fixed-frequency}
	\end{figure}
	
    \subsection{Results on Image Classification Tasks}
    \paragraph{Fixed-frequency scenario.} 
    We first evaluate the effectiveness of EDBA under the fixed-frequency attack setting, where a single compromised client (out of 200 clients in total) is injected into the training process every 10 communication rounds. The attack persists for the first 200 rounds, and the target datasets include MNIST, CIFAR10, CIFAR100, and Tiny-ImageNet, all trained on a ResNet18 backbone. The MA and BA under both IID and Non-IID settings are reported in Fig.~\ref{fixed-frequency}.  
    
    The results demonstrate consistent trends. First, under the IID setting, EDBA achieves a high BA across all datasets while preserving MA close to the clean baseline, confirming that our dynamically optimized trigger can effectively implant backdoors without degrading the main task utility. Second, under the more challenging Non-IID setting, EDBA maintains strong performance, achieving 95.71\% BA on CIFAR10 and 90.87\% BA on Tiny-ImageNet, while MA remains unaffected. These results validate that our method can reliably separate the main and backdoor tasks even in heterogeneous client data distributions.  

    \paragraph{Fixed-pool scenario.} 
    To further evaluate the practicality of EDBA, we consider the fixed-pool scenario, where a subset of clients is compromised throughout training. This setting reflects realistic deployments such as federated mobile applications.
    
\begin{figure}[htbp]
    \centering
    \begin{subfigure}[b]{0.23\textwidth}
        \includegraphics[width=\linewidth]{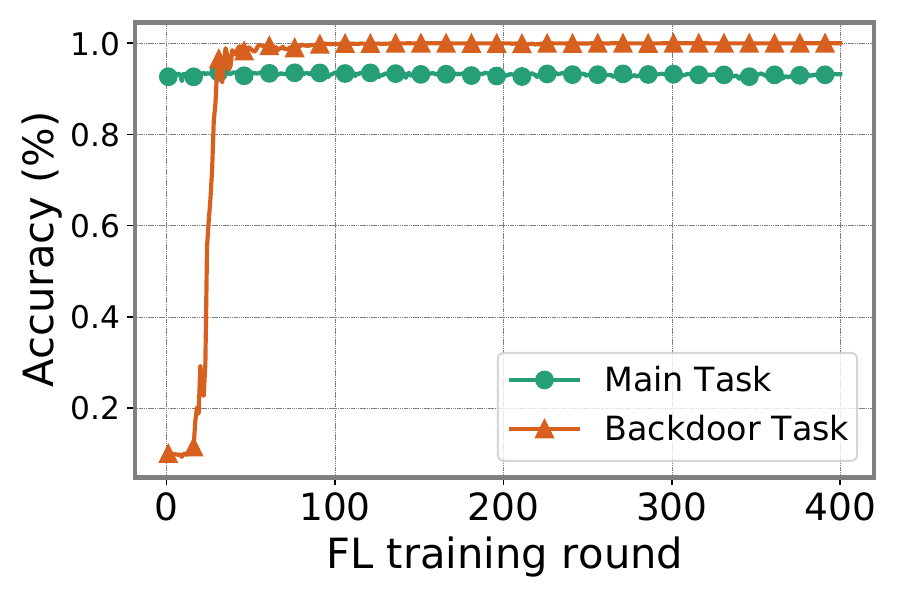}
        \caption{CIFAR10}
        \label{fixed-pool-cifar10}
    \end{subfigure}
    \centering 
    \begin{subfigure}[b]{0.23\textwidth}
        \includegraphics[width=\linewidth]{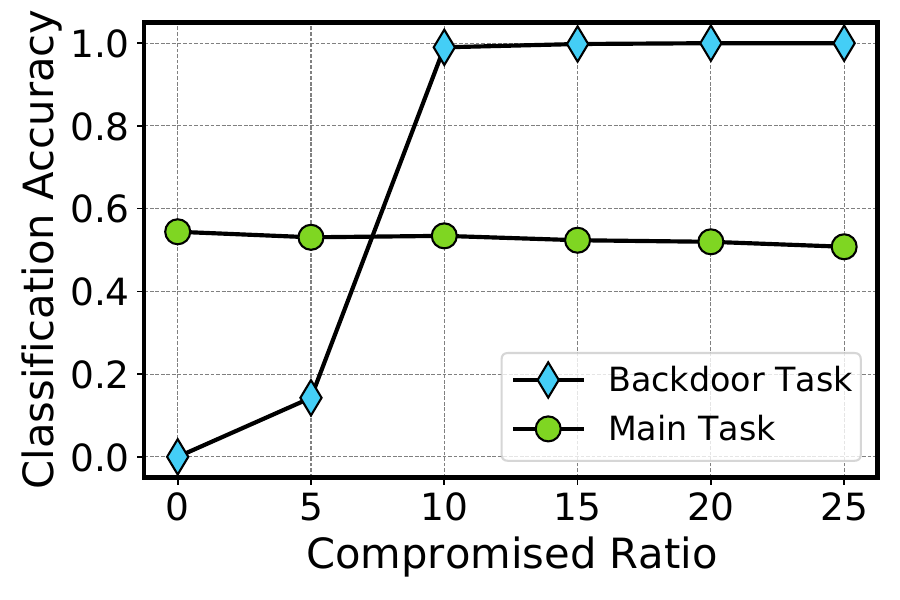}
        \caption{TINY-ImageNet}
        \label{ratio_tiny}
    \end{subfigure}
    \caption{The performance of EDBA under the fixed-pool scenario.}
    \label{}
    \end{figure}

    Fig.~\ref{fixed-pool-cifar10} presents the results on Non-IID CIFAR10 with only 5\% malicious clients. Despite the extremely small proportion of adversaries, EDBA achieves a high BA while maintaining the main MA indistinguishable from the clean baseline. This highlights that our method can implant persistent backdoors without requiring a large coalition of attackers. We further vary the compromised client ratio from 5\% to 25\% on Tiny-ImageNet, as shown in Fig.~\ref{ratio_tiny}. The results reveal that as the adversarial ratio increases, the injected backdoor becomes more effective, while MA remains stable. Moreover, even at the lowest attacker ratio, EDBA already yields a substantial BA, underscoring its efficiency. This demonstrates that FL systems cannot rely on the assumption that small fractions of compromised clients are harmless.

    \subsection{Results on Semantic Analysis Tasks}
\begin{figure}[htbp]
    \centering
    \begin{subfigure}[b]{0.23\textwidth}
        \includegraphics[width=\linewidth]{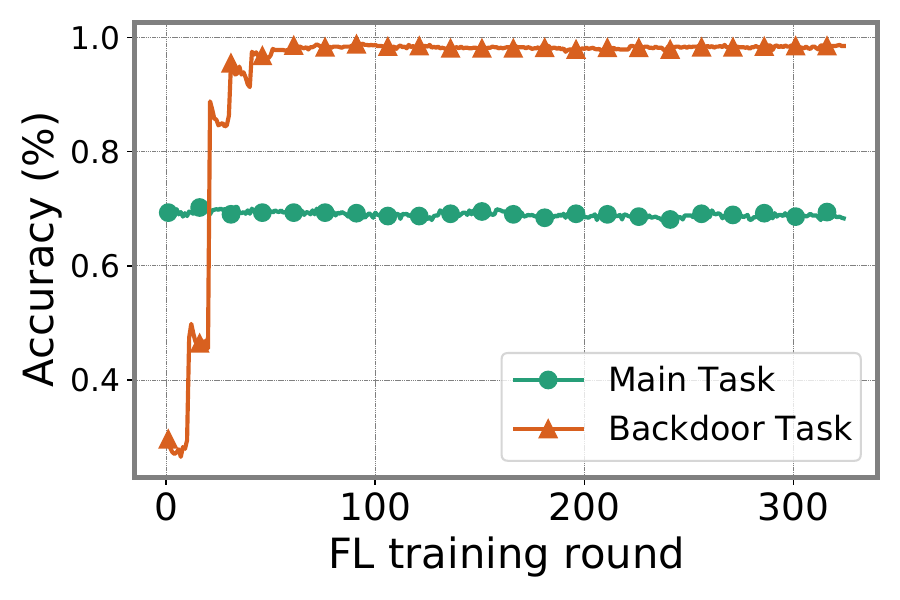}
        \caption{Frequency-Yelp}
        \label{nlp-fixed-frequency}
    \end{subfigure}
    \centering 
    \begin{subfigure}[b]{0.23\textwidth}
        \includegraphics[width=\linewidth]{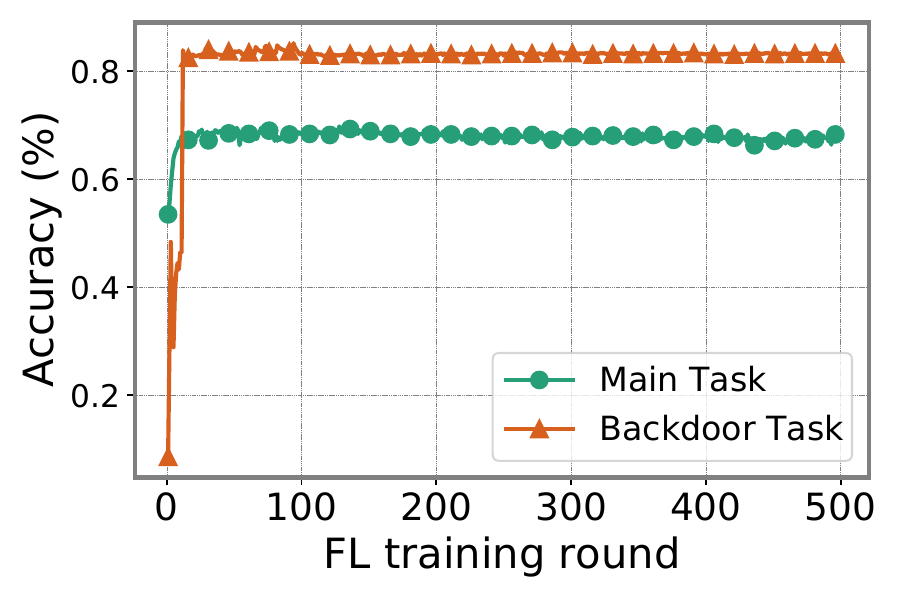}
        \caption{Pool-Yelp}
        \label{nlp-fixed-pool}
    \end{subfigure}
    \centering 
    \begin{subfigure}[b]{0.23\textwidth}
        \includegraphics[width=\linewidth]{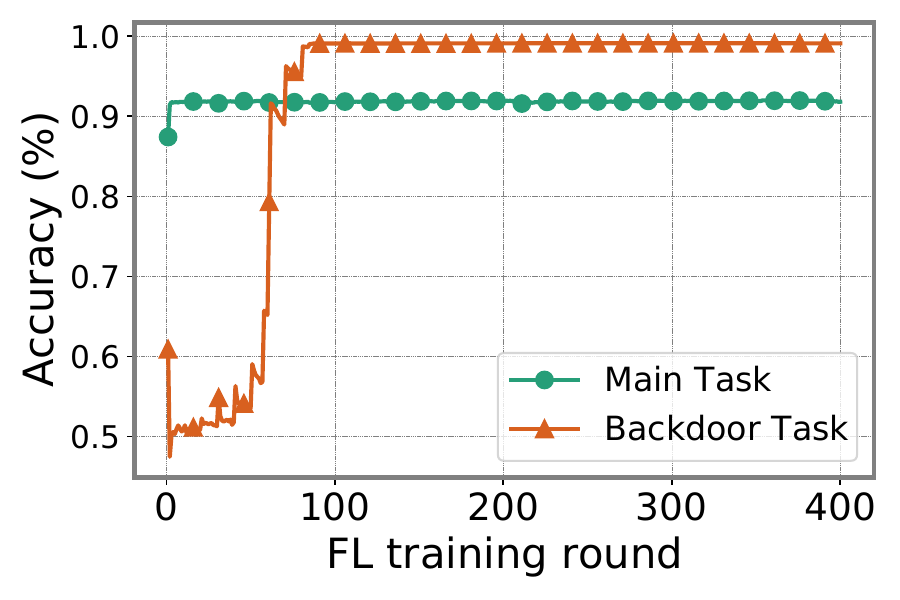}
        \caption{Frequency-IMDB}
        \label{nlp-fixed-frequency-IMDB}
    \end{subfigure}	
    \centering 
    \begin{subfigure}[b]{0.23\textwidth}
        \includegraphics[width=\linewidth]{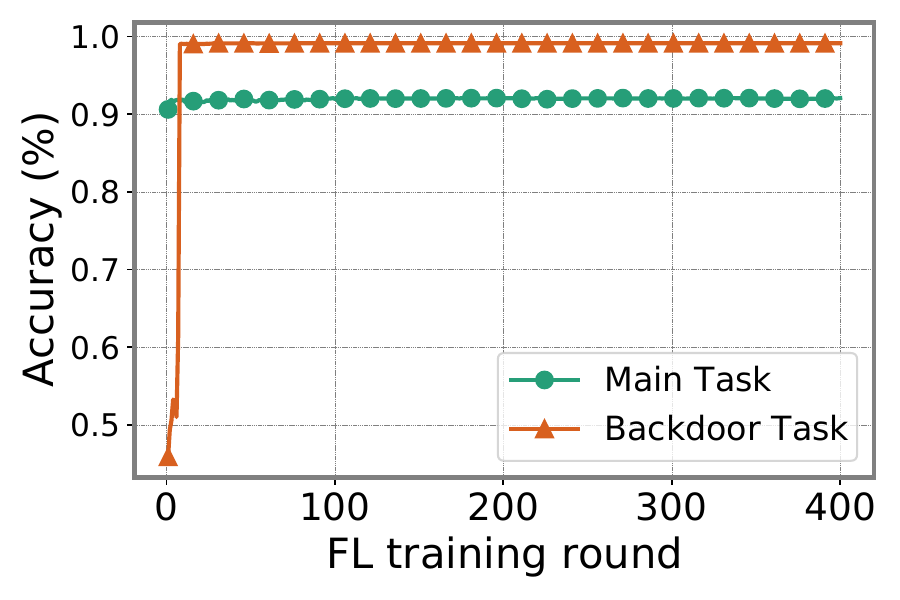}
        \caption{Pool-IMDB}
        \label{nlp-fixed-pool-IMDB}
    \end{subfigure}
    \caption{The performance of EDBA in semantic analysis task.}
    \label{Ratio}
    \end{figure}

    \paragraph{Fixed-frequency scenario.} 
    We evaluate EDBA in natural language processing tasks to verify its cross-task effectiveness. Specifically, we consider the fixed-frequency scenario where a compromised client participates every 10 rounds during the first 100 training epochs of a federated transformer model. As shown in Fig.~\ref{nlp-fixed-frequency}, EDBA rapidly implants trigger tokens in the Yelp dataset under the IID setting. The BA reaches nearly 100\% after only a few malicious updates. Even after the adversarial client is removed, the backdoor effect persists throughout subsequent training, confirming the durability of our attack in sequential language tasks. 

    \paragraph{Fixed-pool scenario.} 
    We further investigate the fixed-pool setting, where a small proportion of clients are compromised. The results on Yelp and IMDB datasets are shown in Figs.~\ref{nlp-fixed-pool} and~\ref{nlp-fixed-pool-IMDB}. With as few as 5\% malicious clients, EDBA achieves a high BA while preserving the main MA, demonstrating that only a limited attackers is sufficient to implant an effective backdoor. Moreover, similar to the vision tasks, the attack does not compromise MA, highlighting that the separation between benign and malicious objectives remains valid in semantic analysis tasks.  

        \begin{figure*}[!h]
    \centering
    \begin{subfigure}{0.16\textwidth}
        \includegraphics[width=\linewidth]{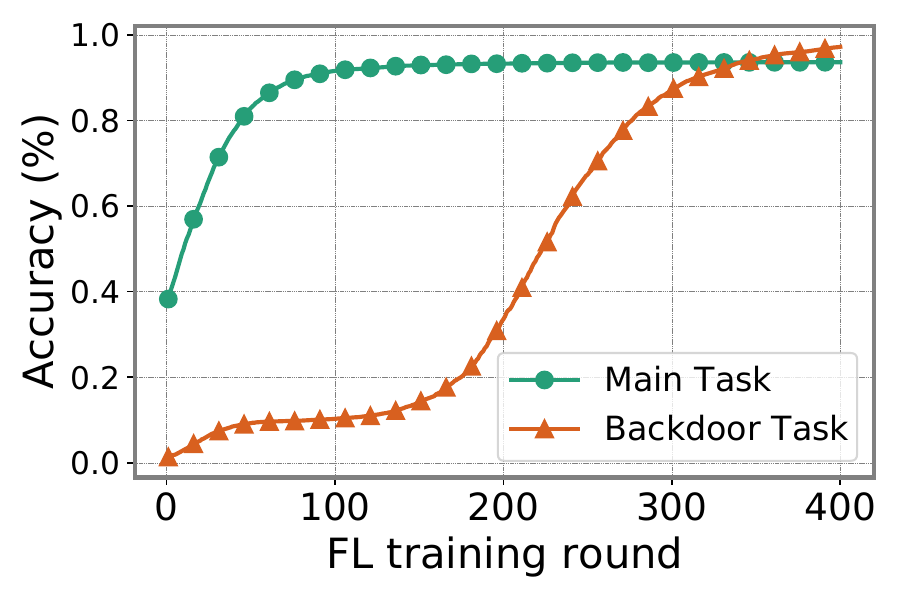}
        \caption{IID-NDC}
    \end{subfigure}
    \hfill
    \begin{subfigure}{0.16\textwidth}
        \includegraphics[width=\linewidth]{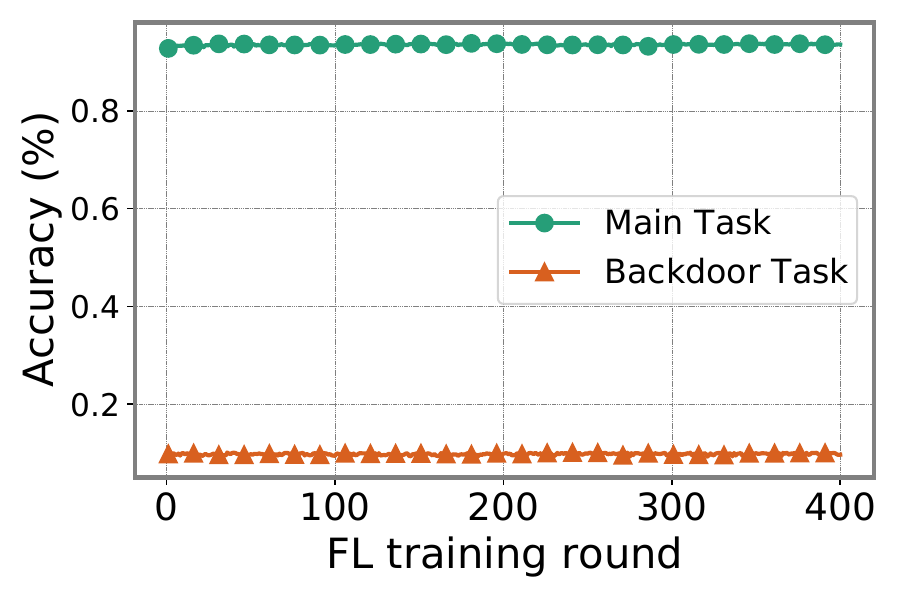}
        \caption{IID-Krum}
    \end{subfigure}
    \hfill
    \begin{subfigure}{0.16\textwidth}
        \includegraphics[width=\linewidth]{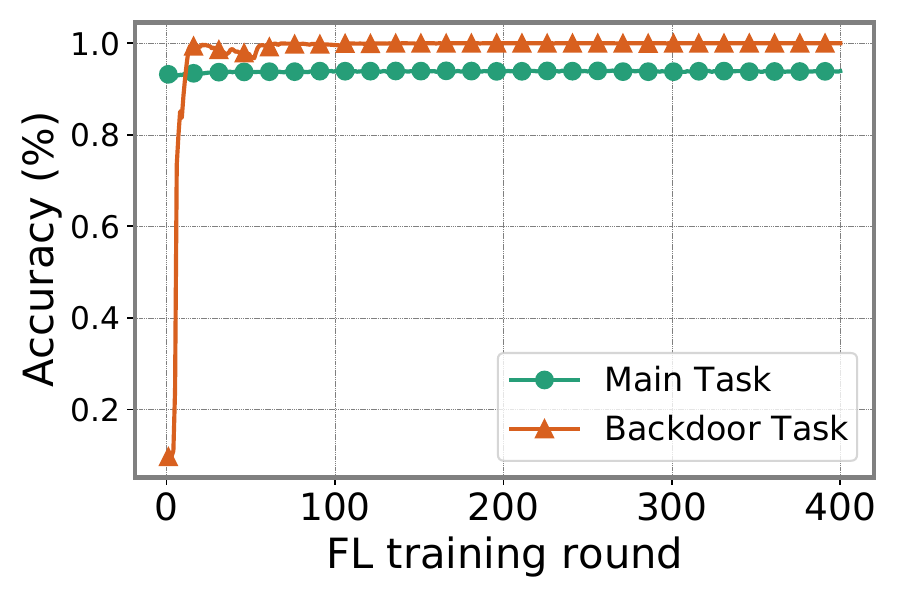}
        \caption{IID-MKrum}
    \end{subfigure}
    \hfill
    \begin{subfigure}{0.16\textwidth}
        \includegraphics[width=\linewidth]{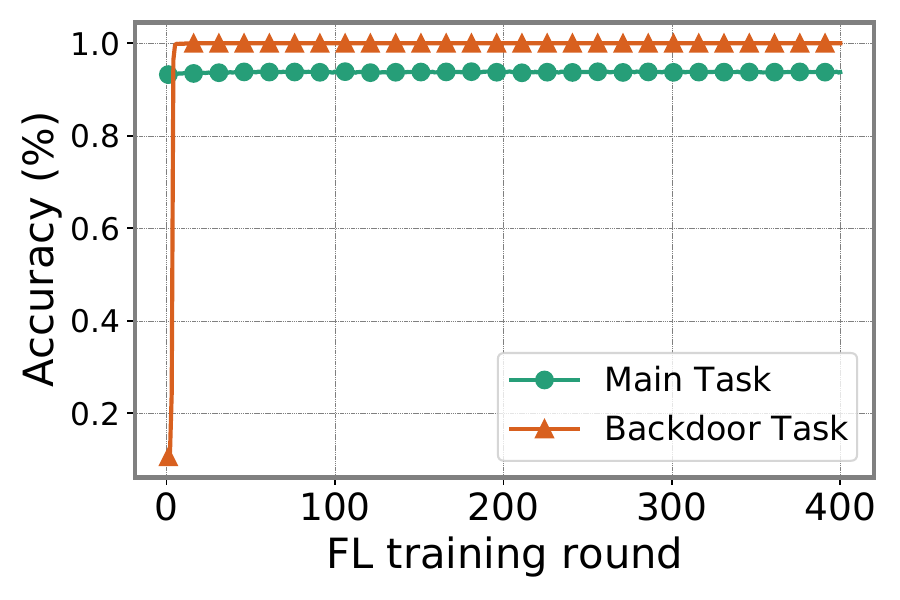}
        \caption{IID-Median}
    \end{subfigure}
    \hfill
    \begin{subfigure}{0.16\textwidth}
        \includegraphics[width=\linewidth]{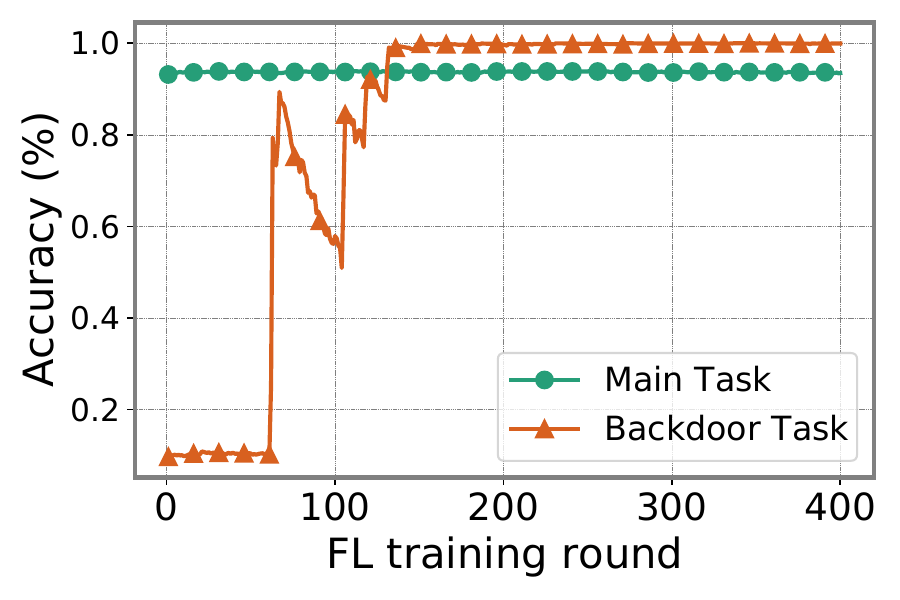}
        \caption{IID-FLAME}
    \end{subfigure}
    \hfill
    \begin{subfigure}{0.16\textwidth}
        \includegraphics[width=\linewidth]{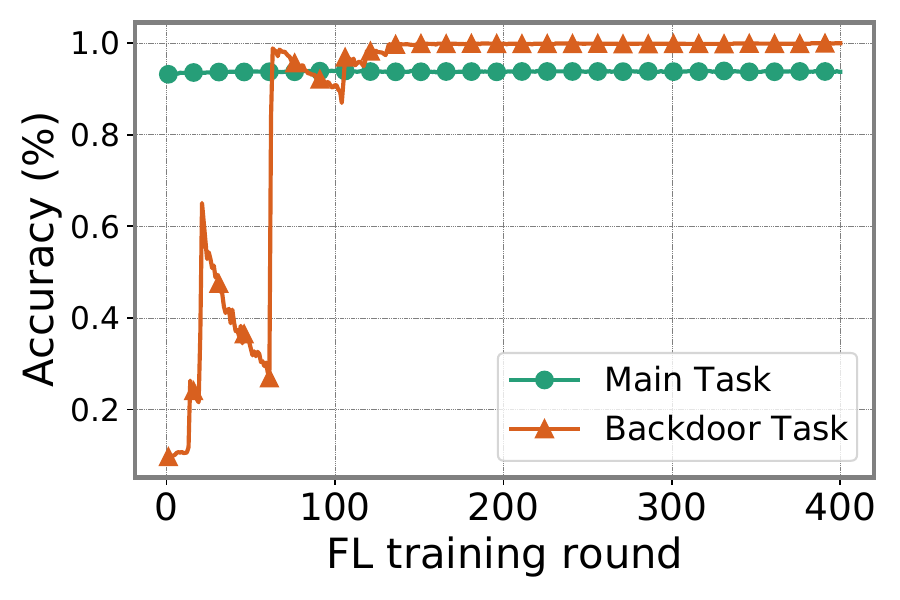}
        \caption{IID-FreqFed}
    \end{subfigure}
    \begin{subfigure}{0.16\textwidth}
        \includegraphics[width=\linewidth]{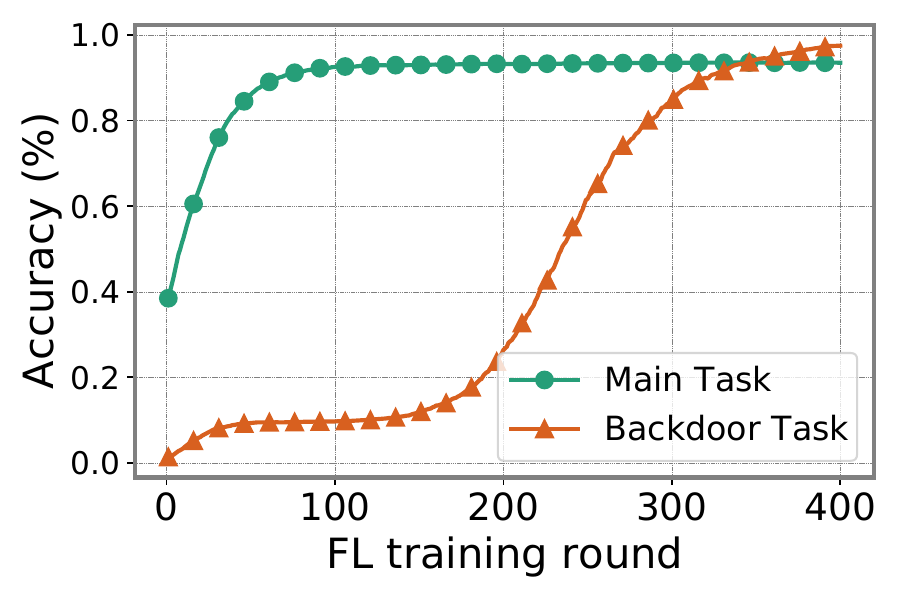}
        \caption{Non-NDC}
    \end{subfigure}
    \hfill
    \begin{subfigure}{0.16\textwidth}
        \includegraphics[width=\linewidth]{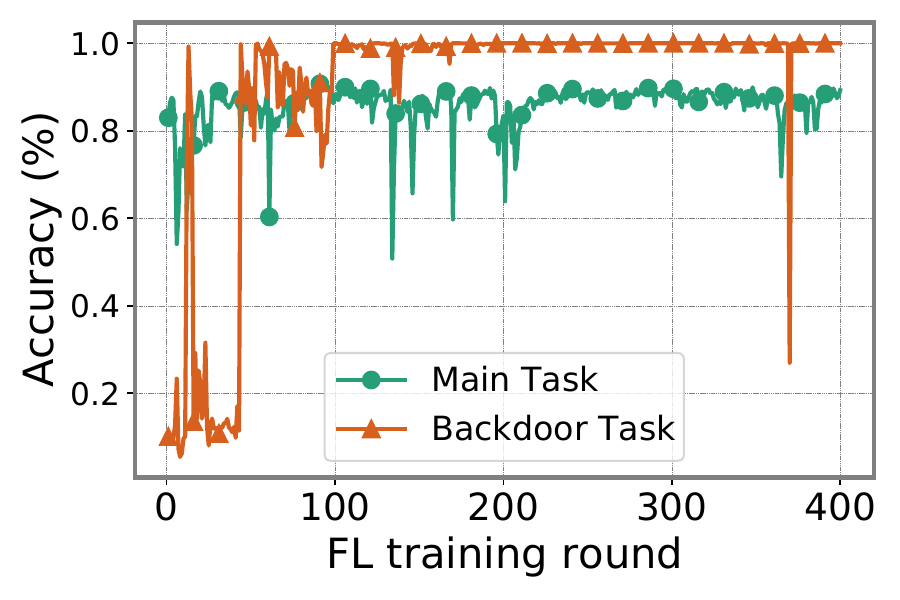}
        \caption{Non-Krum}
    \end{subfigure}
    \hfill
    \begin{subfigure}{0.16\textwidth}
        \includegraphics[width=\linewidth]{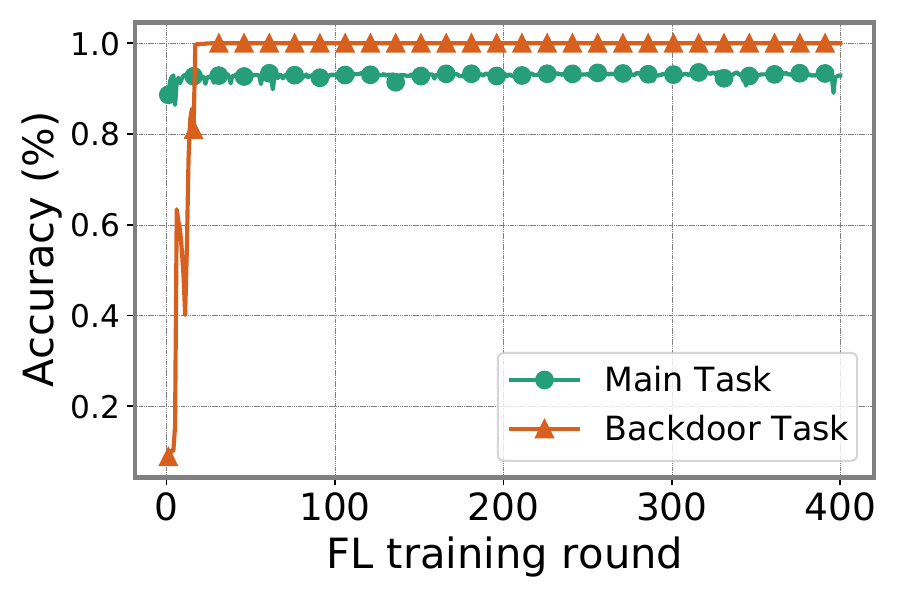}
        \caption{Non-MKrum}
    \end{subfigure}
    \hfill
    \begin{subfigure}{0.16\textwidth}
        \includegraphics[width=\linewidth]{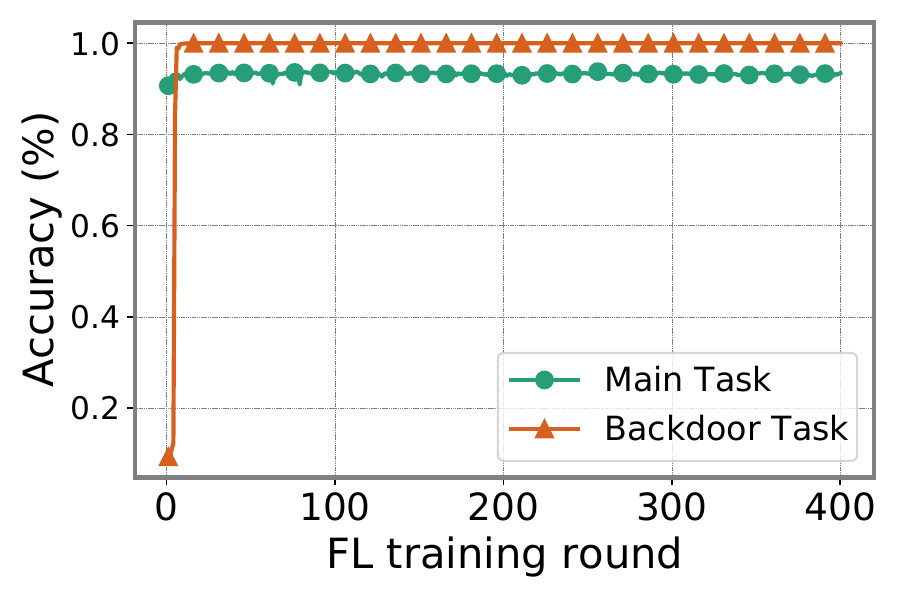}
        \caption{Non-Median}
    \end{subfigure}
    \hfill
    \begin{subfigure}{0.16\textwidth}
        \includegraphics[width=\linewidth]{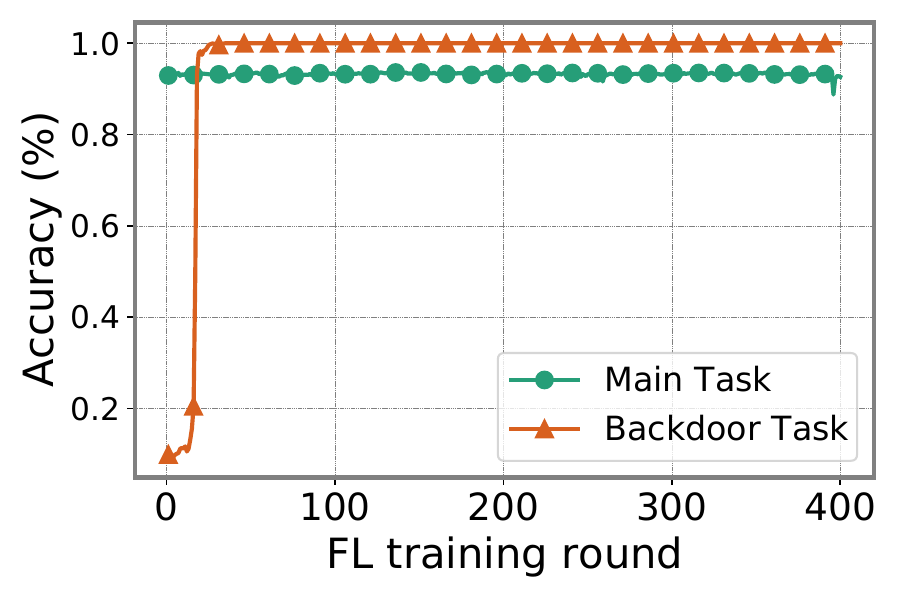}
        \caption{Non-FLAME}
    \end{subfigure}
    \hfill
    \begin{subfigure}{0.16\textwidth}
        \includegraphics[width=\linewidth]{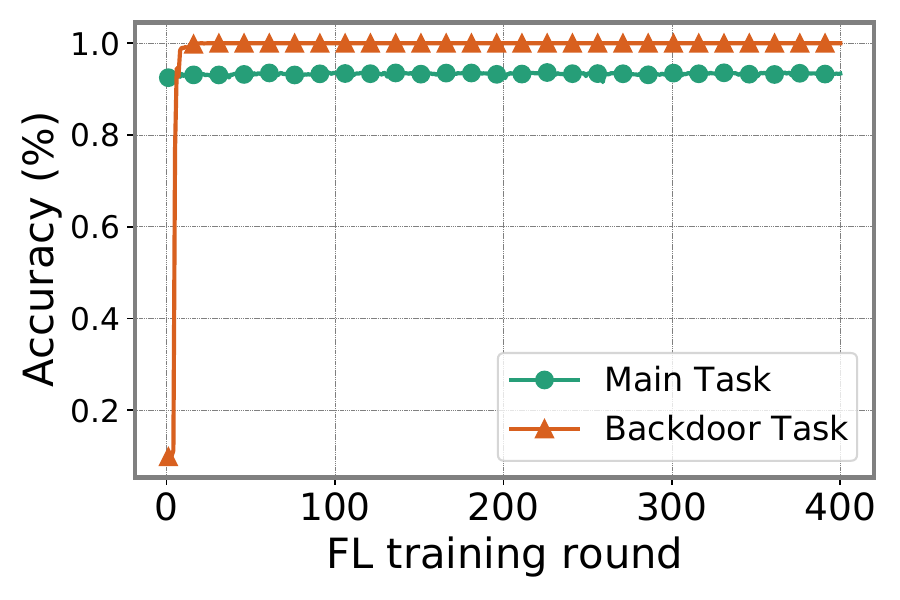}
        \caption{Non-FreqFed}
    \end{subfigure}
    \caption{The performance of EDBA in  under different defense methods with Non-IID setting and fixed-pool attack scenario.}
    \label{defense_EDBA_CIFAR10}
    \end{figure*}

        \begin{table*}[htbp]  
    \centering  
    \caption{Comparison with other attack methods under the different FL defenses.}    
    \resizebox{0.85\textwidth}{!}{
    \begin{tabular}{ccccccccc}    
        \toprule    
        \toprule    
        \multirow{2}[4]{*}{Defense} &
        \multirow{2}[4]{*}{Metric} & 
        \multicolumn{7}{c}{Method} \\
        \cmidrule{3-9}          &       & BadNets & Scaling & IBA   & Neuroxin & A3FL  & Chameleon & EDBA \\    
        \midrule    
        \multirow{2}[2]{*}{None} & MA    & 93.46 & 92.35 & 88.66 & 92.95 & 93.43 & 91.82 & 93.18 \\          & BA    & 9.43  & 100.00 & 99.42 & 95.15 & 100.00 & 97.47 & 100.00 \\    
        \midrule    
        \multirow{2}[2]{*}{NDC~\cite{sun2019can}} & MA    & 93.49 & 87.40 & 89.14 & 93.83 & 93.91 &    93.81   & 93.54 \\          & BA    & 3.03  & 10.31 & 99.50 & 10.09 & 99.95 &   95.31    & 99.75 \\   
        \midrule   
        \multirow{2}[2]{*}{Krum~\cite{blanchard2017machine}} & MA    & 43.79 & 92.97 & 86.58 & 90.44 & 85.07 & 87.53 & 89.98 \\          & BA    & 22.76 & 9.74  & 91.69 & 95.30 & 100.00 & 98.25 & 100.00 \\    
        \midrule    
        \multirow{2}[2]{*}{Multi-Krum~\cite{blanchard2017machine}} & MA    & 93.23 & 91.03 & 87.32 & 92.59 & 92.52 & 90.67 & 93.38 \\          & BA    & 5.67  & 100.00 & 99.87 & 95.02 & 100.00 & 98.27 & 100.00 \\    
        \midrule    
        \multirow{2}[2]{*}{Median~\cite{yin2018byzantine}} & MA    & 92.63 & 90.91 & 88.20 & 93.33 & 92.48 &    91.59   & 93.28 \\          & BA    & 10.43 & 100.00 & 99.89 & 92.56 & 100.00 &   97.91    & 99.84 \\    
        \midrule    
        \multirow{2}[2]{*}{FLAME~\cite{nguyen2022flame}} & MA    & 93.84 & 92.73 & 64.21 & 92.87 & 93.32 & 92.94 & 93.59 \\          & BA    & 9.27  & 10.14 & 21.28 & 94.54 & 100.00 & 97.38 & 100.00 \\    
        \midrule    
        \multirow{2}[2]{*}{FreqFed~\cite{fereidooni2023freqfed}} & MA    & 93.84 & 93.28 & 81.30 & 92.73 & 93.21 & 93.10 & 93.61 \\          & BA    & 9.64  & 98.04 & 17.04 & 95.41 & 100.00 & 97.62 & 100.00 \\    
        \bottomrule    
        \bottomrule    
    \end{tabular}%  
    }
    \label{defense}%
    \end{table*}%

    \begin{figure*}[ht]
    \centering
    \begin{subfigure}{0.19\textwidth}
    \includegraphics[width=\linewidth]{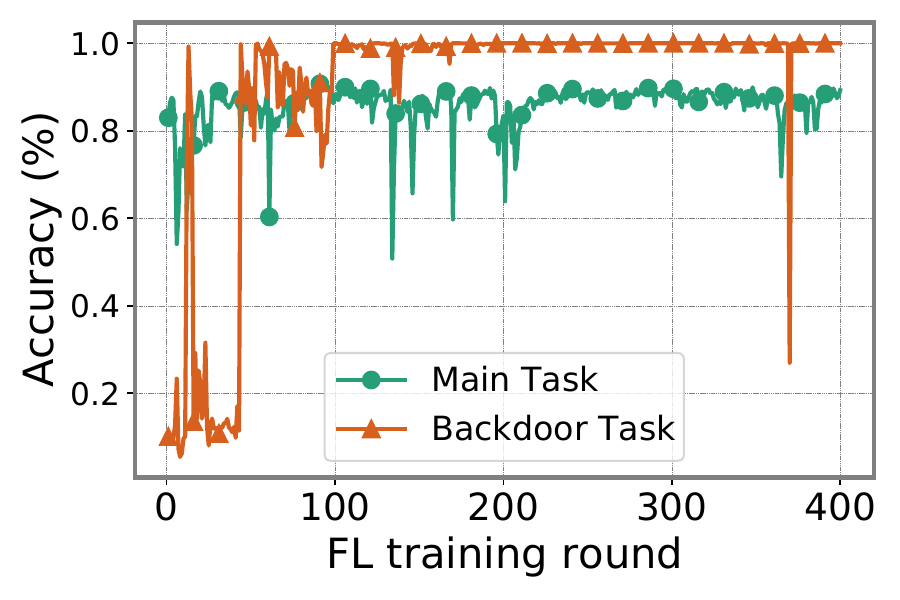}
    \caption{EDBA}
    \end{subfigure}
    \begin{subfigure}{0.19\textwidth}
    \includegraphics[width=\linewidth]{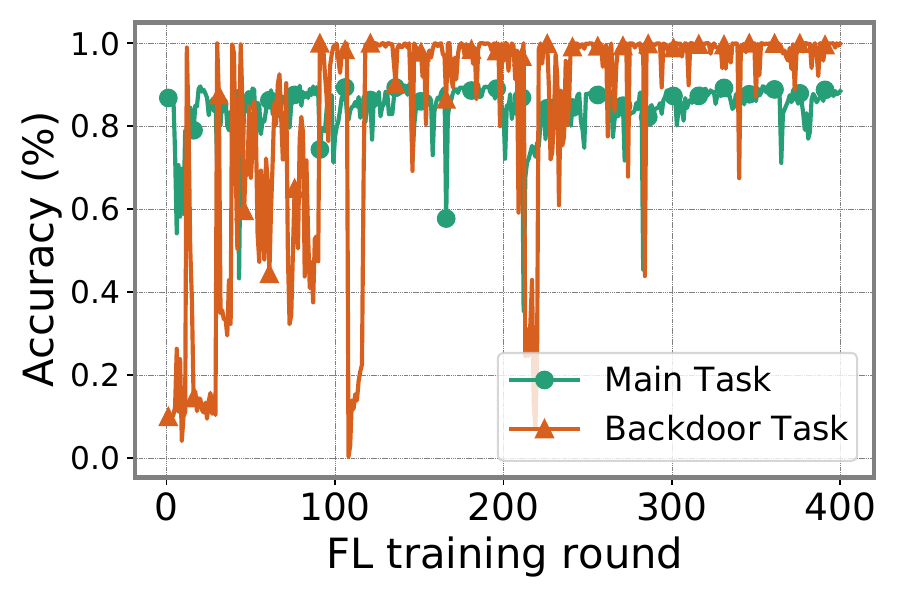}
    \caption{A3FL}
    \end{subfigure}
    \begin{subfigure}{0.19\textwidth}
    \includegraphics[width=\linewidth]{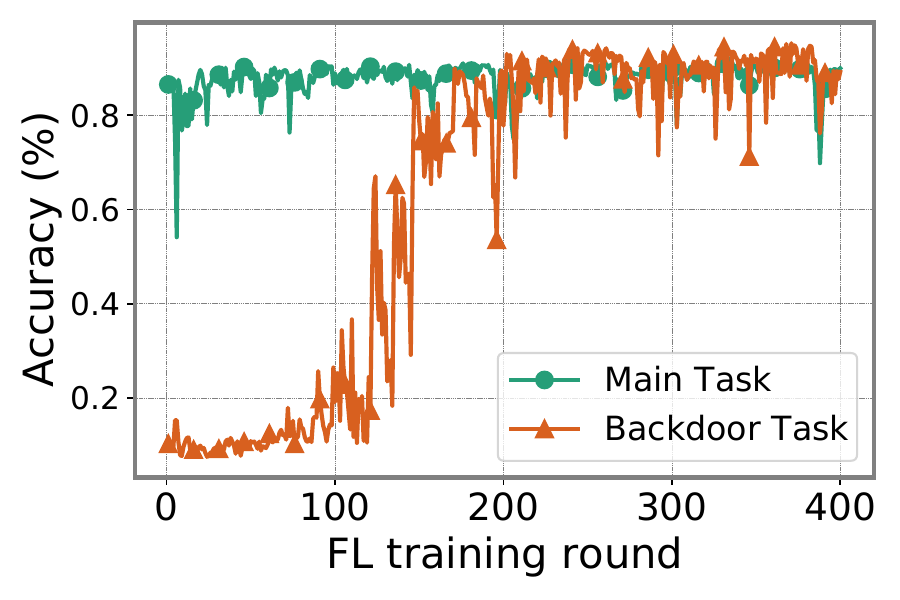}
    \caption{Neuroxin}
    \end{subfigure}
    \begin{subfigure}{0.19\textwidth}
    \includegraphics[width=\linewidth]{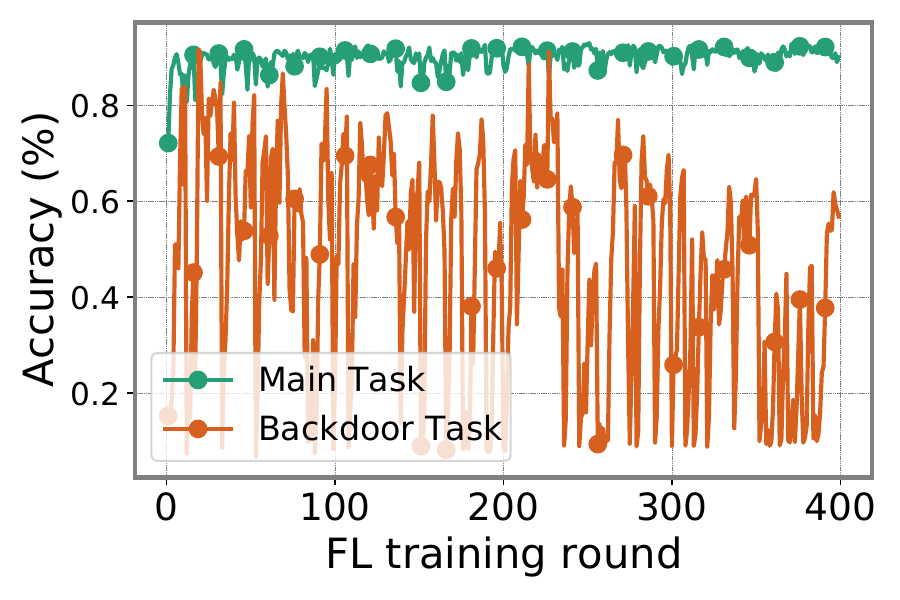}
    \caption{IBA}
    \end{subfigure}
    \begin{subfigure}{0.19\textwidth}
    \includegraphics[width=\linewidth]{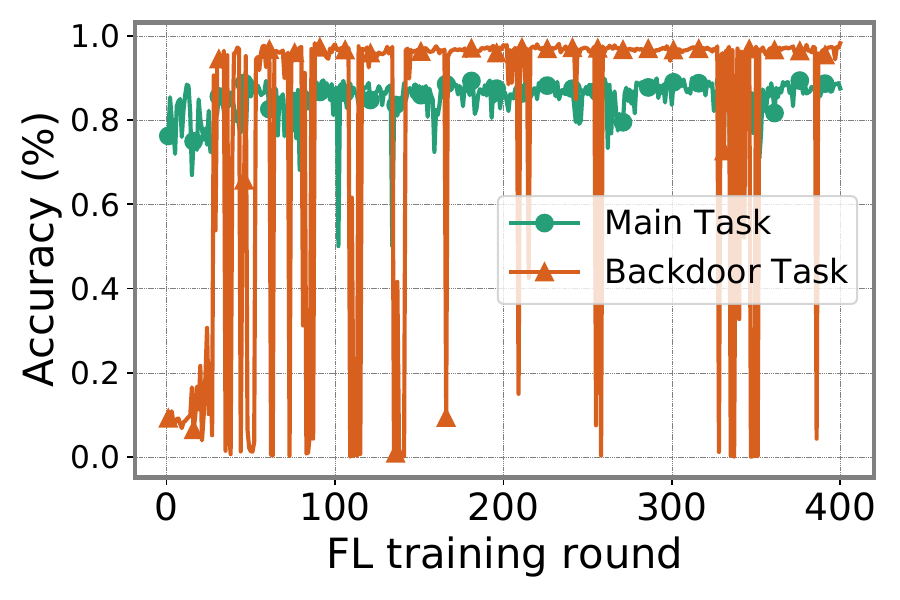}
    \caption{Chameleon}
    \end{subfigure}
    \caption{The comparison of EDBA and other attack methods under the Krum defense.}
    \label{Comparision_Krum}
    \end{figure*}

    \subsection{Results under Defense Mechanisms}
    We evaluate the resilience of EDBA against representative FL defenses using the Non-IID CIFAR10 dataset with 25\% compromised clients under the fixed-pool scenario (Fig.~\ref{defense_EDBA_CIFAR10}). Norm-based defenses such as NDC, which attempt to mitigate scaling attacks by clipping local updates, are bypassed by EDBA. Without requiring any parameter scaling, EDBA still achieves nearly 100\% BA. Second, aggregation-based defenses such as Krum and Multi-Krum, although designed to exclude anomalous updates, remain ineffective against EDBA in the Non-IID case. Because EDBA decouples the backdoor from the main task, its malicious updates remain statistically similar to benign ones and are thus selected by Krum. Moreover, under IID settings, Krum is more effective, suggesting that data heterogeneity weakens its filtering capability. Other aggregation-based defenses, including Median, FLAME, and FreqFed also fail to mitigate EDBA while preserving high MA. In all cases, BA converges close to 100\% without degrading MA, indicating that EDBA successfully separates the two tasks.  
    
    Table~\ref{defense} provides a broader comparison with existing attacks across defenses. Although methods such as IBA, Neurotoxin, and A3FL occasionally reach BA levels comparable to EDBA, their performance is inconsistent and easily degraded by defenses such as FLAME or FreqFed. By contrast, EDBA maintains a high BA and stable MA across all settings. 
    
    To further investigate the training stability, we compare convergence under Krum (Fig.~\ref{Comparision_Krum}). The comparison exhibits heavy fluctuations in BA during training. In contrast, EDBA achieves stable convergence, with BA and MA remaining steady throughout. This stability highlights that by dynamically optimizing the trigger and decoupling the backdoor task from the main task, EDBA minimizes mutual interference during training. As a result, the global model can maintain high performance on both tasks, even with only one client to update the global model.

    \begin{figure*}[t]
    \centering
    \begin{subfigure}{0.16\textwidth}
    \includegraphics[width=\linewidth]{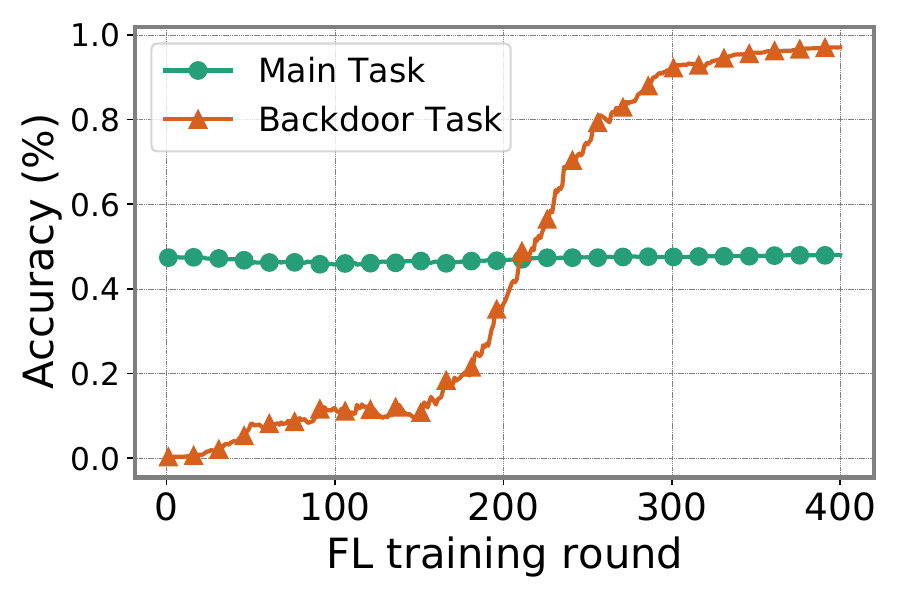}
    \caption{IID-NDC}
    \end{subfigure}
    \begin{subfigure}{0.16\textwidth}
    \includegraphics[width=\linewidth]{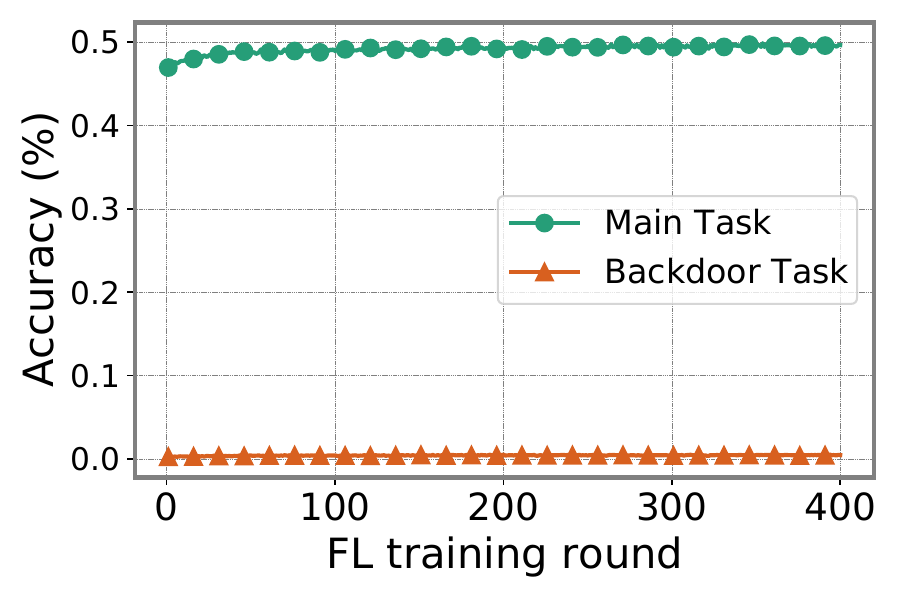}
    \caption{IID-Krum}
    \end{subfigure}
    \begin{subfigure}{0.16\textwidth}
    \includegraphics[width=\linewidth]{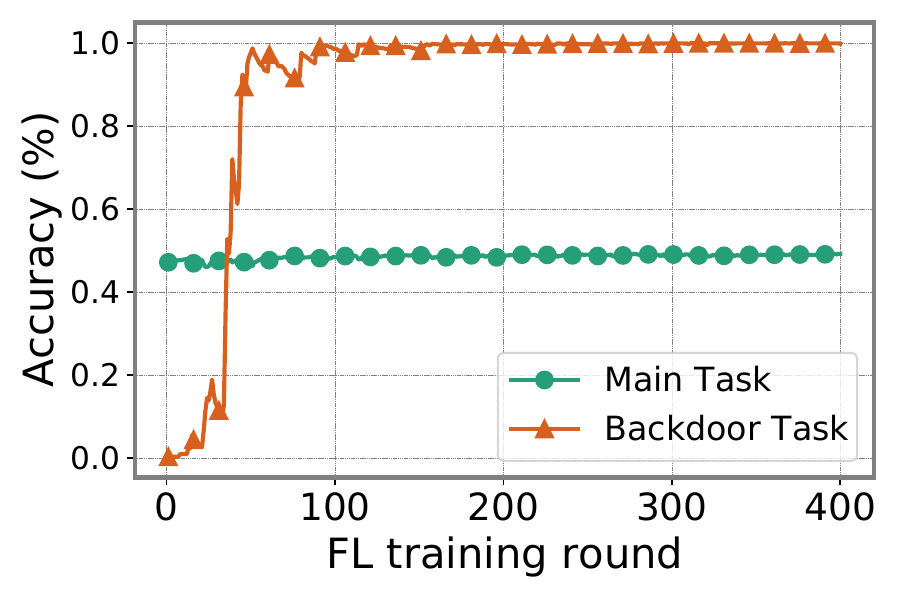}
    \caption{IID-MKrum}
    \end{subfigure}
    \begin{subfigure}{0.16\textwidth}
    \includegraphics[width=\linewidth]{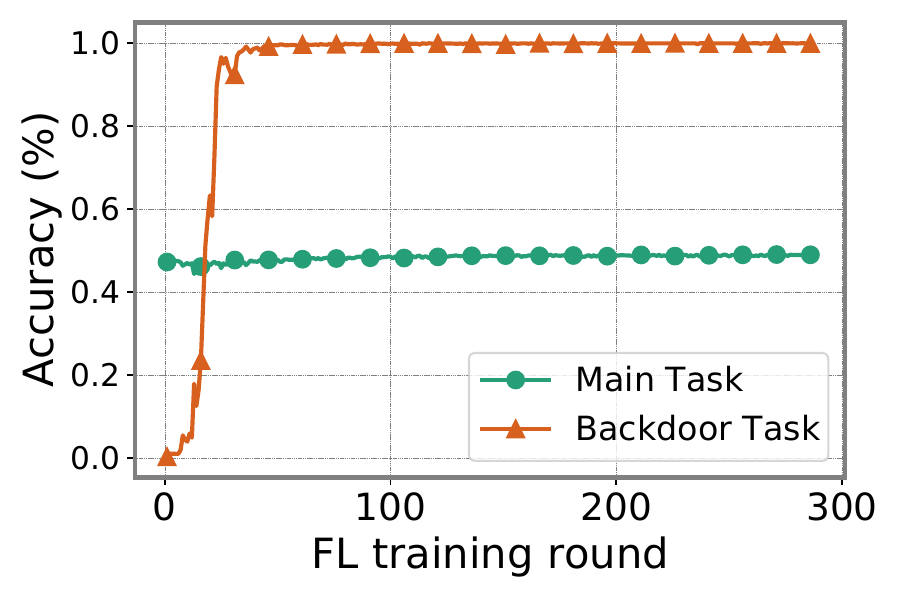}
    \caption{IID-Median}
    \end{subfigure}
    \begin{subfigure}{0.16\textwidth}
    \includegraphics[width=\linewidth]{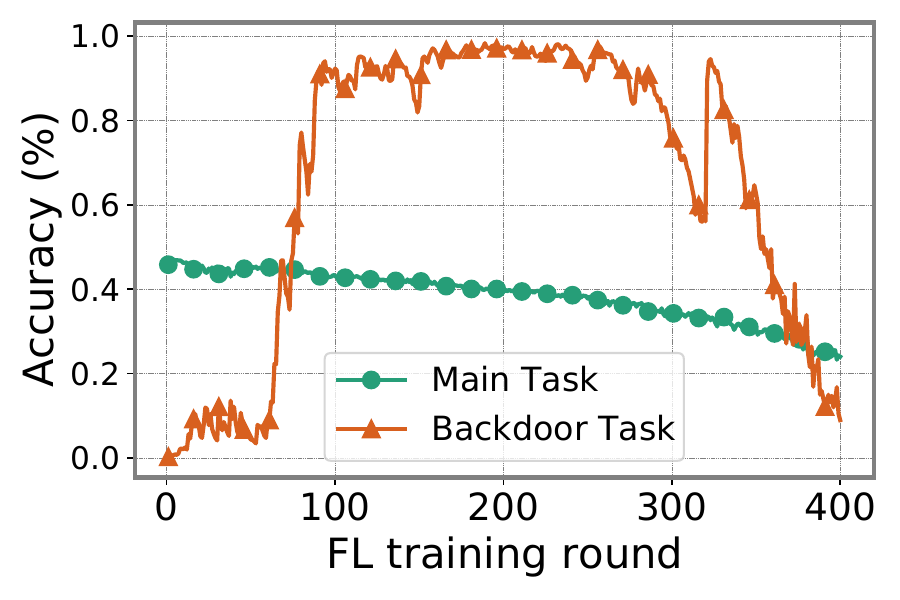}
    \caption{IID-FLAME}
    \end{subfigure}
    \begin{subfigure}{0.16\textwidth}
    \includegraphics[width=\linewidth]{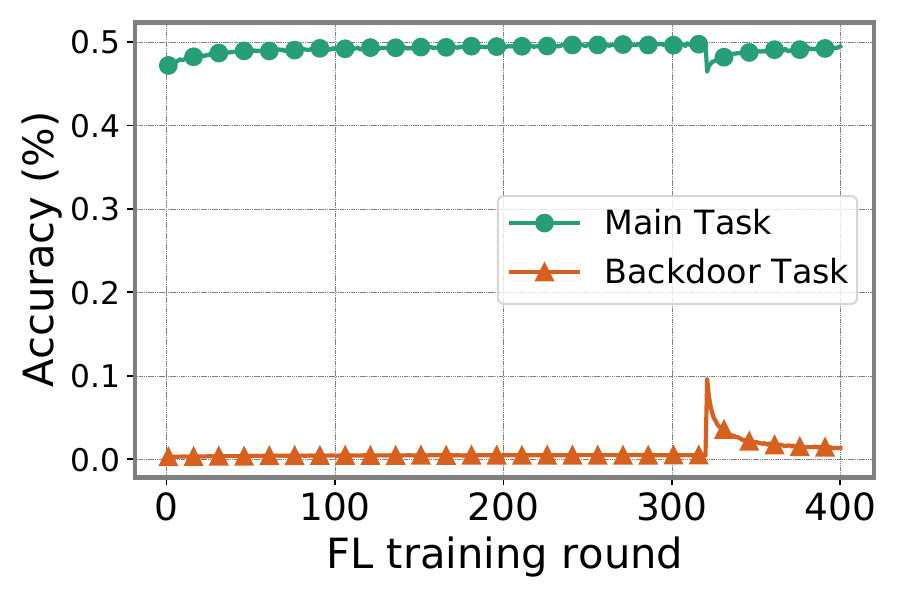}
    \caption{IID-FreqFed}
    \end{subfigure}
    \hfill
    \begin{subfigure}{0.16\textwidth}
    \includegraphics[width=\linewidth]{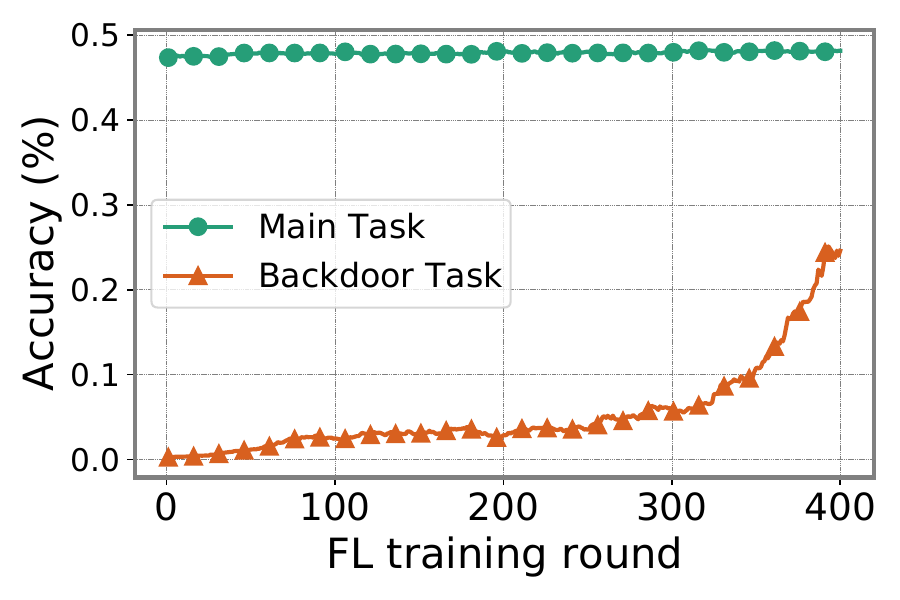}
    \caption{Non-NDC}
    \end{subfigure}
    \begin{subfigure}{0.16\textwidth}
    \includegraphics[width=\linewidth]{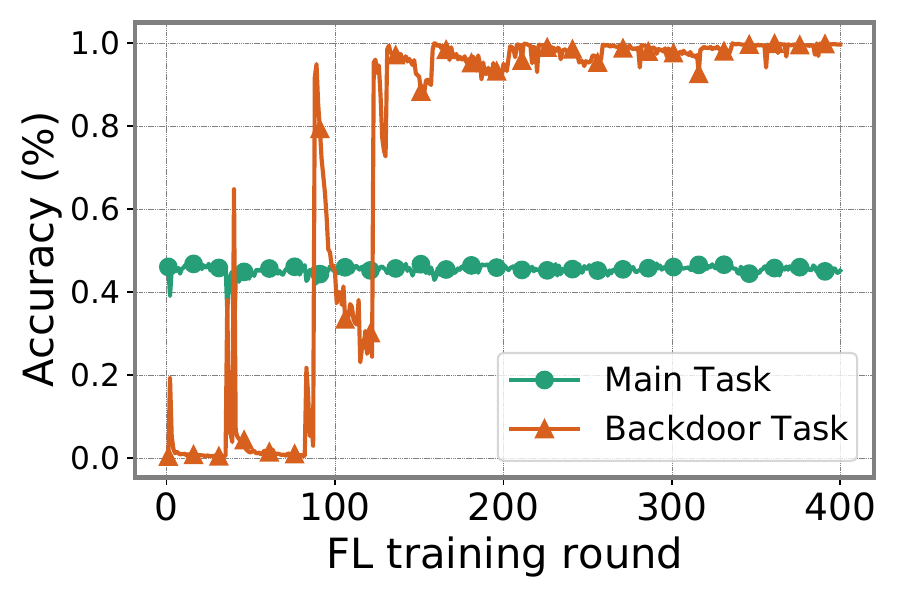}
    \caption{Non-Krum}
    \end{subfigure}
    \begin{subfigure}{0.16\textwidth}
    \includegraphics[width=\linewidth]{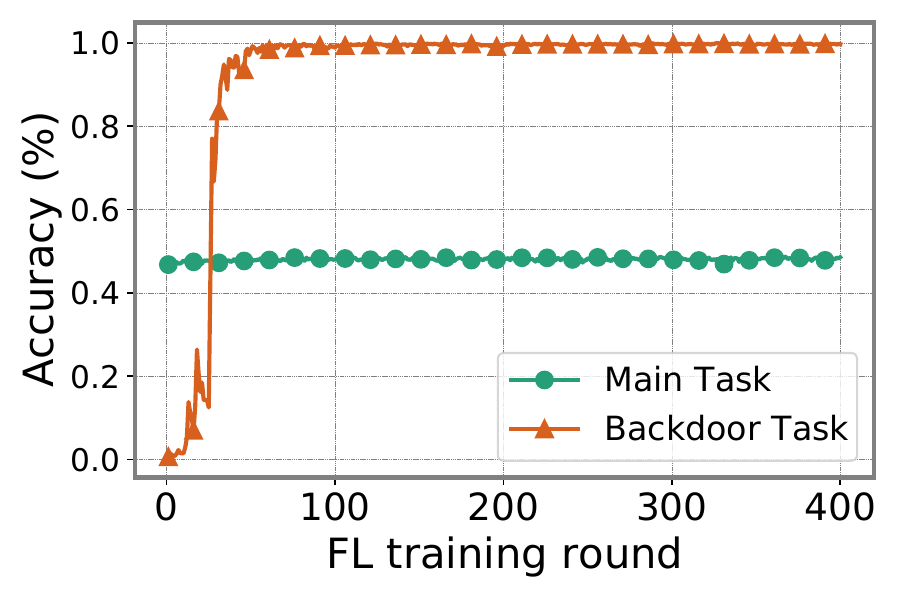}
    \caption{Non-MKrum}
    \end{subfigure}
    \begin{subfigure}{0.16\textwidth}
    \includegraphics[width=\linewidth]{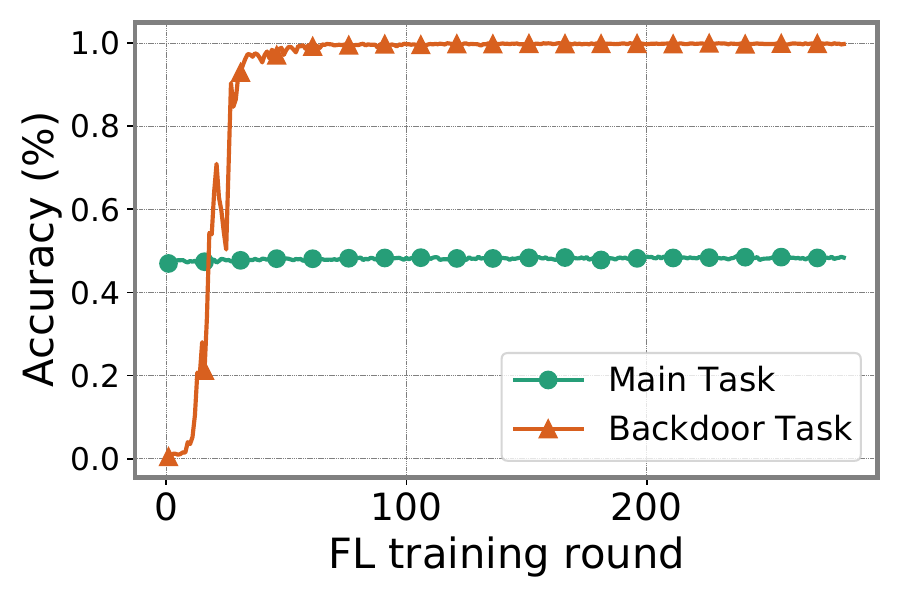}
    \caption{Non-Median}
    \end{subfigure}
    \begin{subfigure}{0.16\textwidth}
    \includegraphics[width=\linewidth]{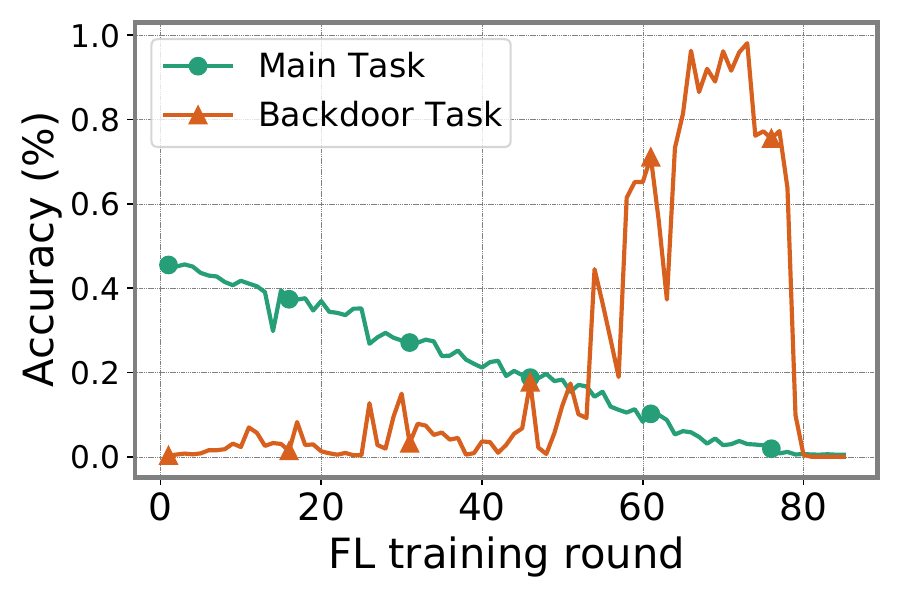}
    \caption{Non-FLAME}
    \end{subfigure}
    \begin{subfigure}{0.16\textwidth}
    \includegraphics[width=\linewidth]{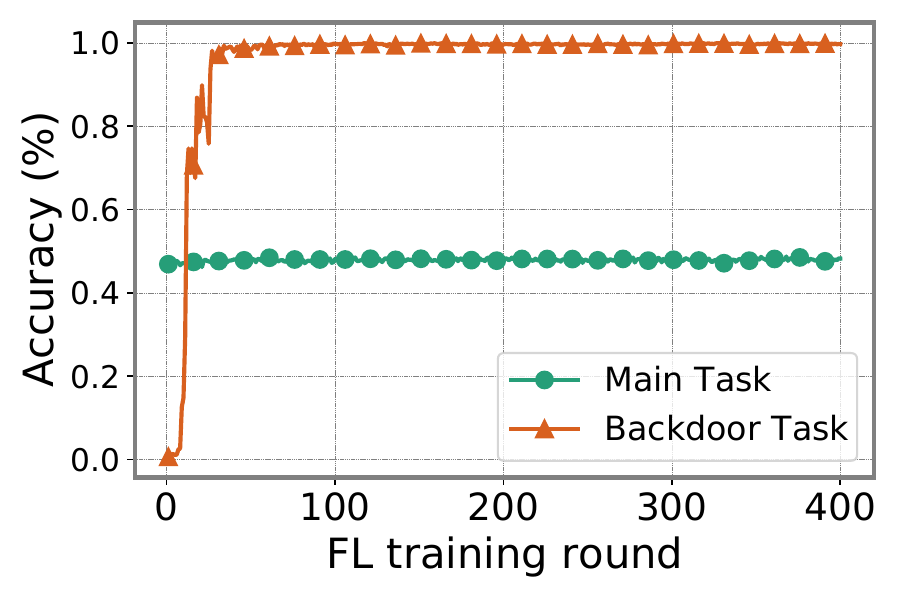}
    \caption{Non-FreqFed}
    \end{subfigure}
    \caption{The performance of EDBA under the fixed pool setting with 10\% compromised clients.}
    \label{defense_EDBA_CIFAR10_percent10}
    \end{figure*}
    
    \begin{figure*}[t]
    \centering
    \begin{subfigure}{0.16\textwidth}
    \includegraphics[width=\linewidth]{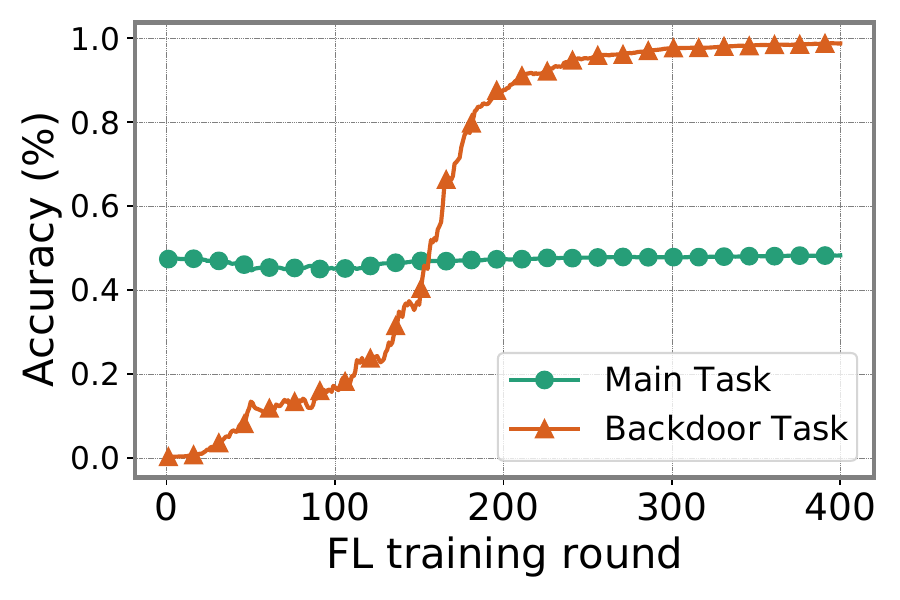}
    \caption{IID-NDC}
    \end{subfigure}
    \begin{subfigure}{0.16\textwidth}
    \includegraphics[width=\linewidth]{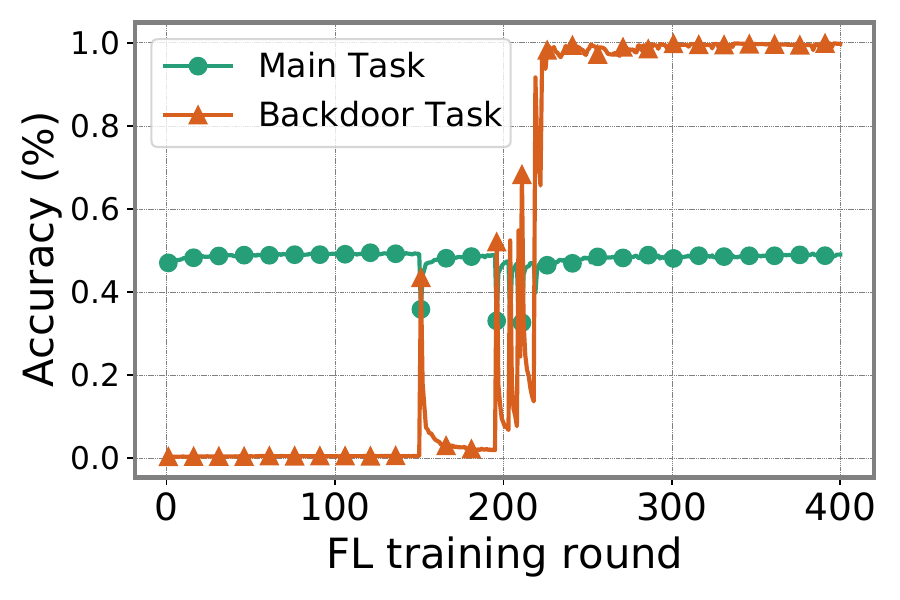}
    \caption{IID-Krum}
    \end{subfigure}
    \begin{subfigure}{0.16\textwidth}
    \includegraphics[width=\linewidth]{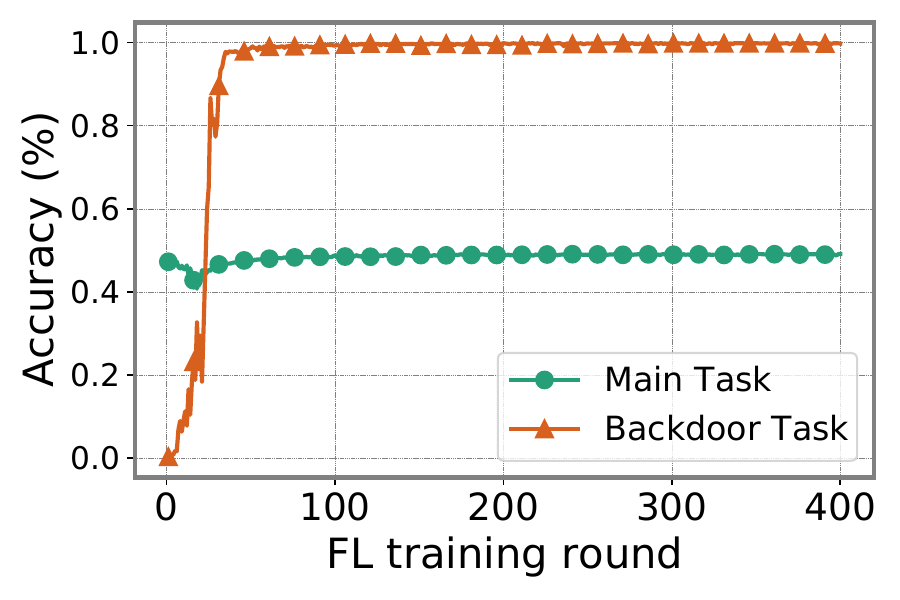}
    \caption{IID-MKrum}
    \end{subfigure}
    \begin{subfigure}{0.16\textwidth}
    \includegraphics[width=\linewidth]{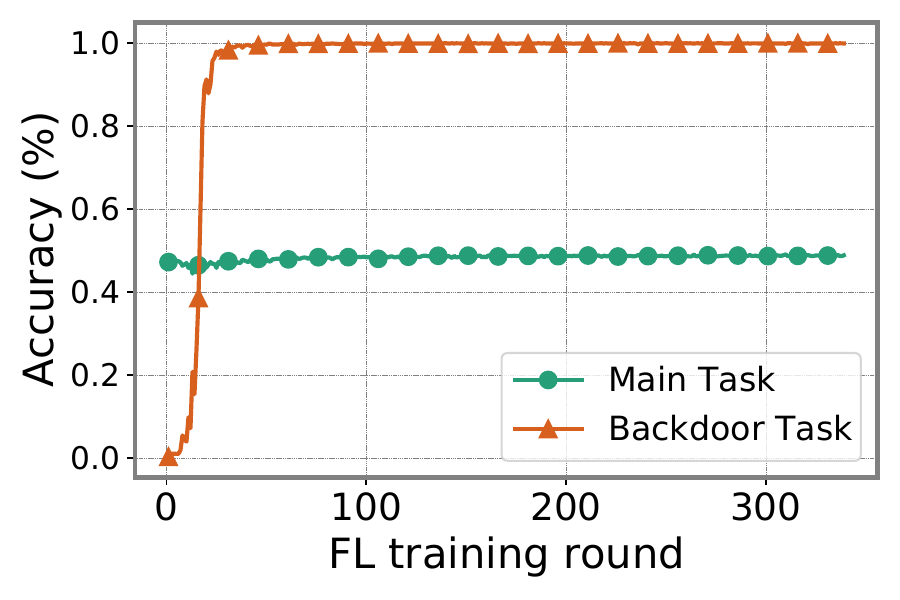}
    \caption{IID-Median}
    \end{subfigure}
    \begin{subfigure}{0.16\textwidth}
    \includegraphics[width=\linewidth]{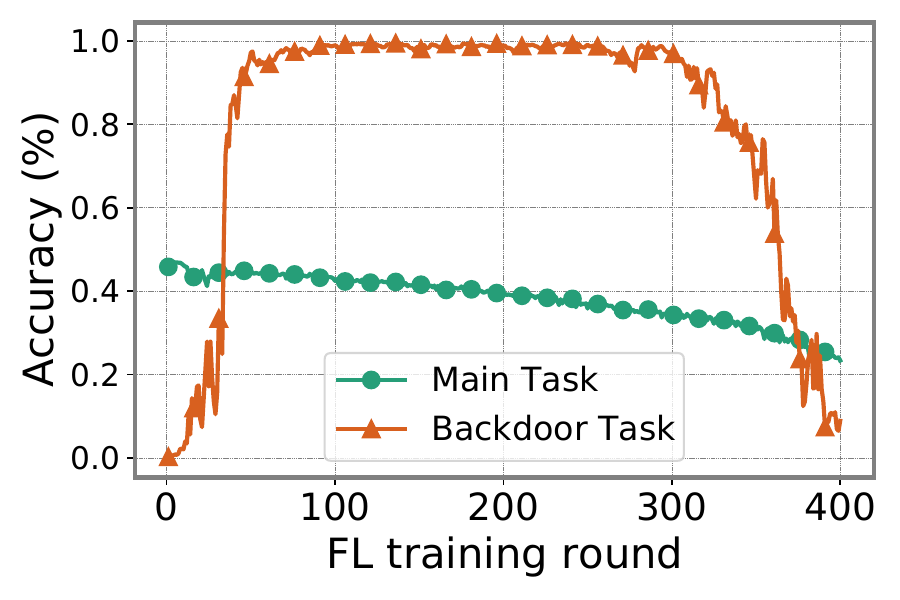}
    \caption{IID-FLAME}
    \end{subfigure}
    \begin{subfigure}{0.16\textwidth}
    \includegraphics[width=\linewidth]{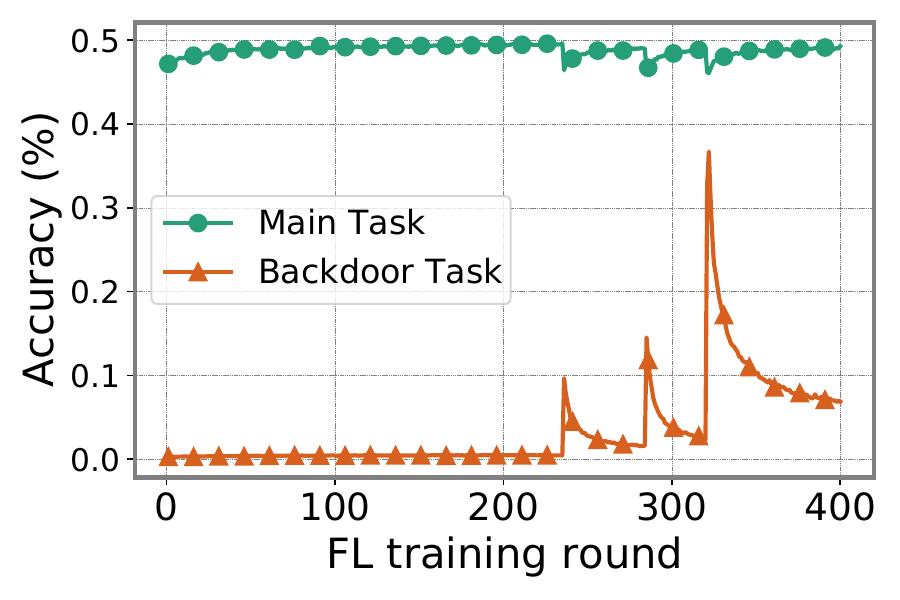}
    \caption{IID-FreqFed}
    \end{subfigure}
    \hfill
    \begin{subfigure}{0.16\textwidth}
    \includegraphics[width=\linewidth]{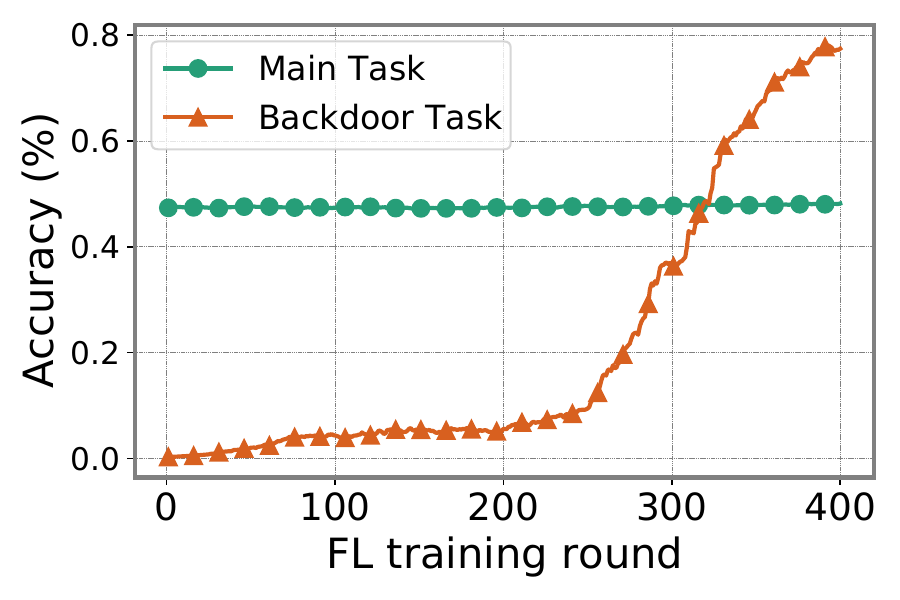}
    \caption{Non-NDC}
    \end{subfigure}
    \begin{subfigure}{0.16\textwidth}
    \includegraphics[width=\linewidth]{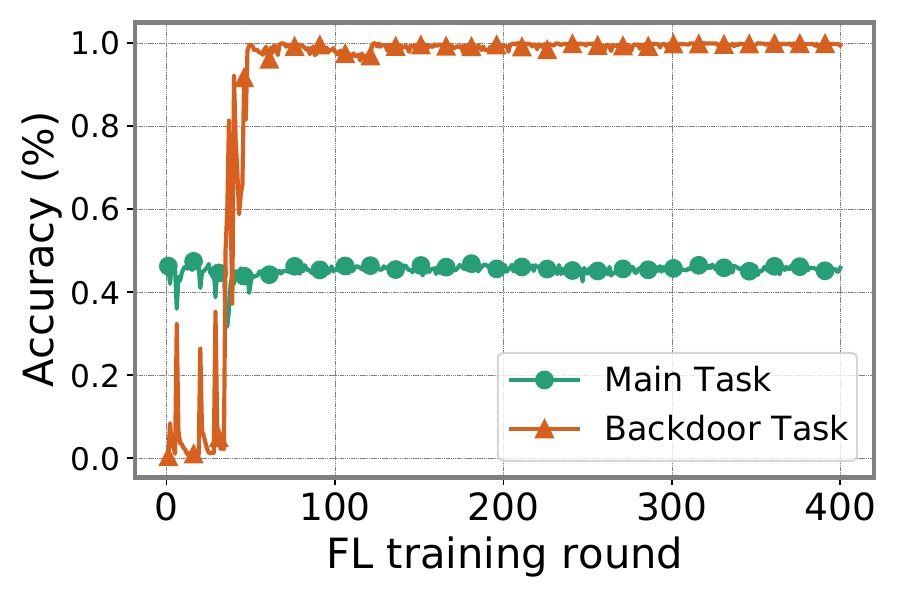}
    \caption{Non-Krum}
    \end{subfigure}
    \begin{subfigure}{0.16\textwidth}
    \includegraphics[width=\linewidth]{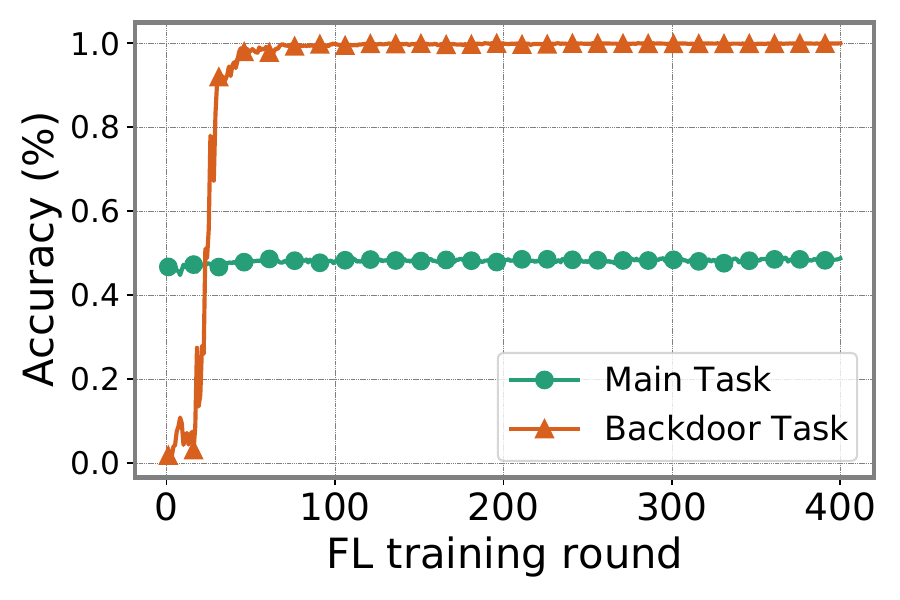}
    \caption{Non-MKrum}
    \end{subfigure}
    \begin{subfigure}{0.16\textwidth}
    \includegraphics[width=\linewidth]{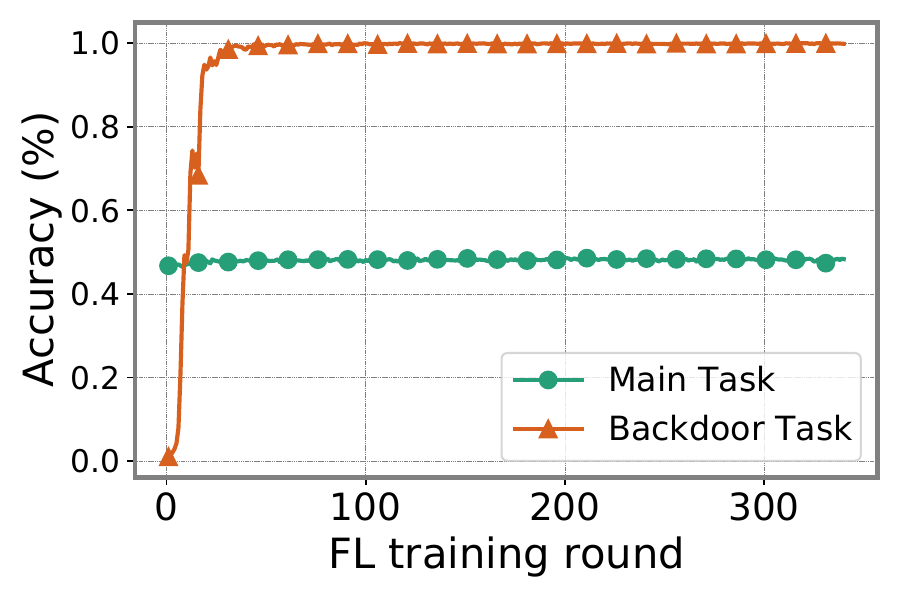}
    \caption{Non-Median}
    \end{subfigure}
    \begin{subfigure}{0.16\textwidth}
    \includegraphics[width=\linewidth]{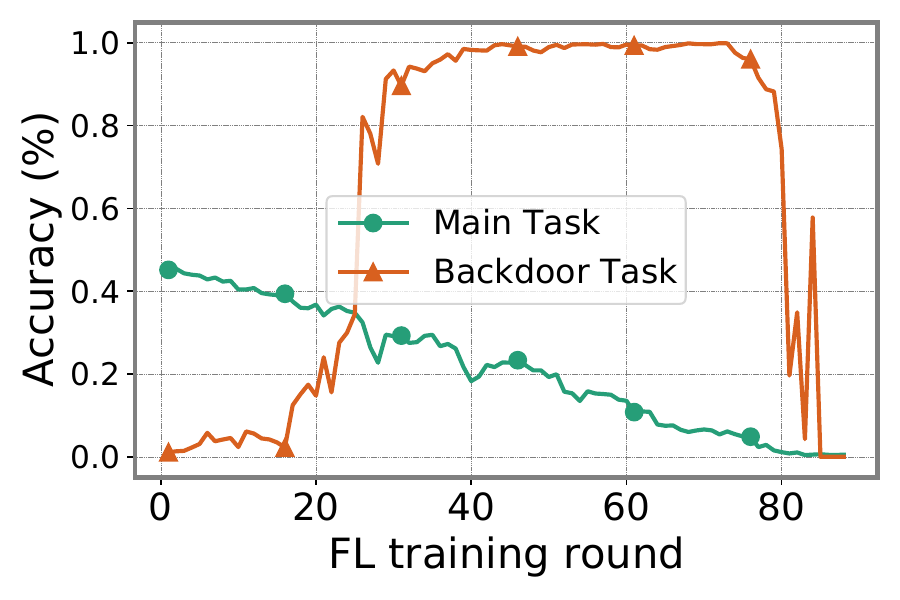}
    \caption{Non-FLAME}
    \end{subfigure}
    \begin{subfigure}{0.16\textwidth}
    \includegraphics[width=\linewidth]{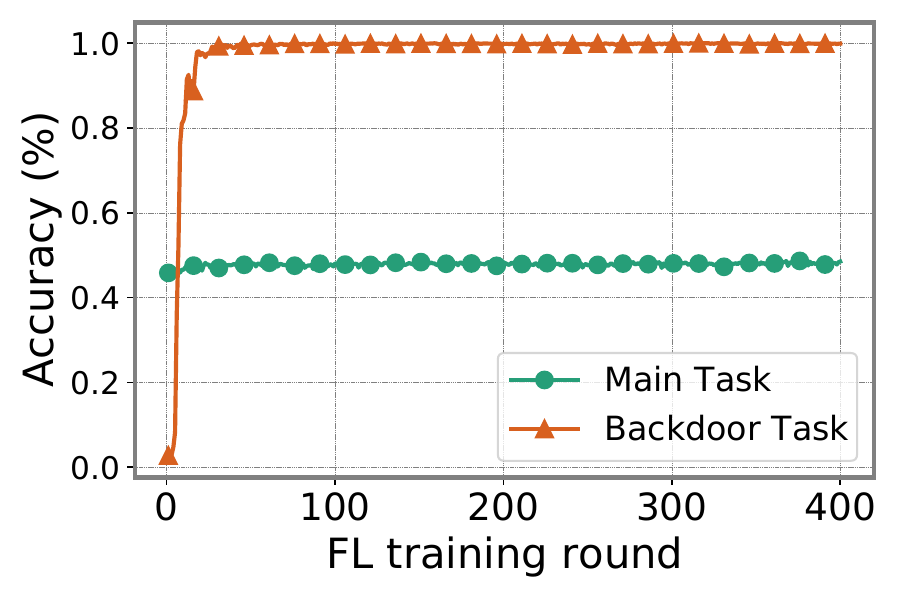}
    \caption{Non-FreqFed}
    \end{subfigure}
    \caption{The performance of EDBA under the fixed pool setting with 15\% compromised clients.}
    \label{defense_EDBA_TINY_percent15}
    \end{figure*}
    
    \begin{figure*}[t]
    \centering
    \begin{subfigure}{0.16\textwidth}
    \includegraphics[width=\linewidth]{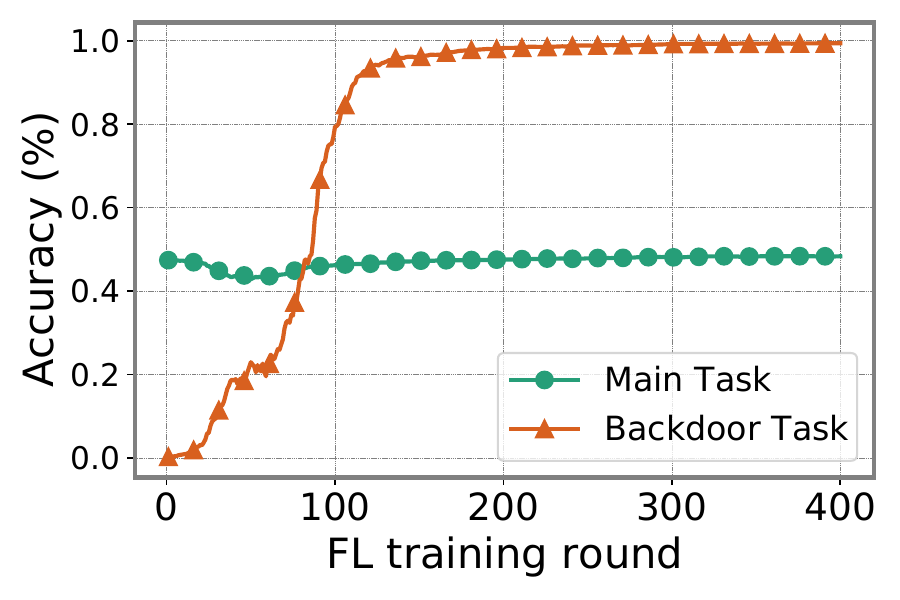}
    \caption{IID-NDC}
    \end{subfigure}
    \begin{subfigure}{0.16\textwidth}
    \includegraphics[width=\linewidth]{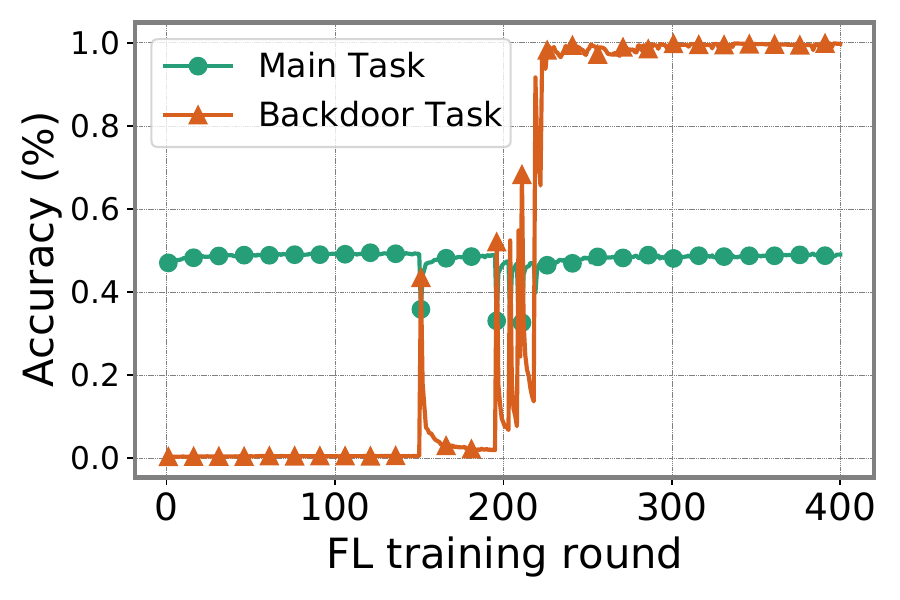}
    \caption{IID-Krum}
    \end{subfigure}
    \begin{subfigure}{0.16\textwidth}
    \includegraphics[width=\linewidth]{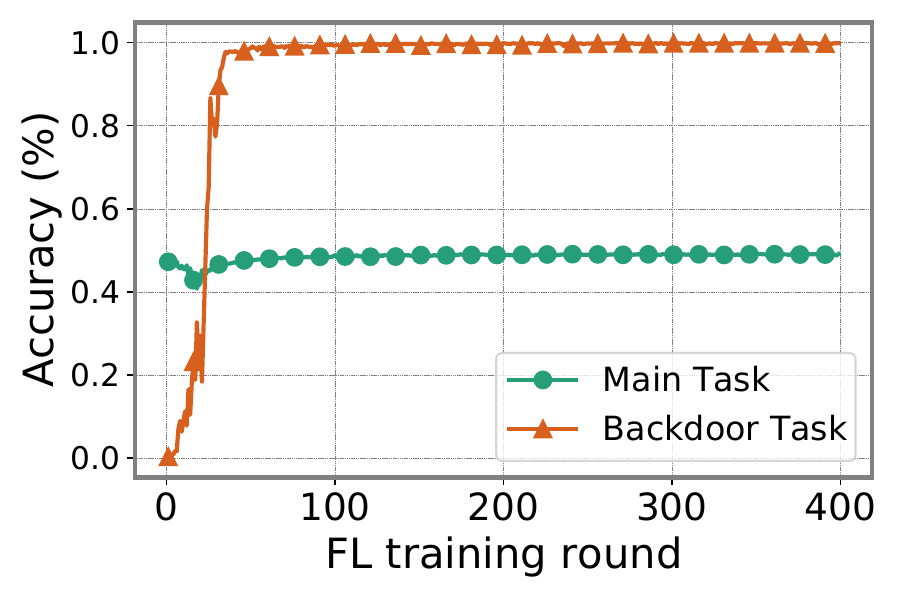}
    \caption{IID-MKrum}
    \end{subfigure}
    \begin{subfigure}{0.16\textwidth}
    \includegraphics[width=\linewidth]{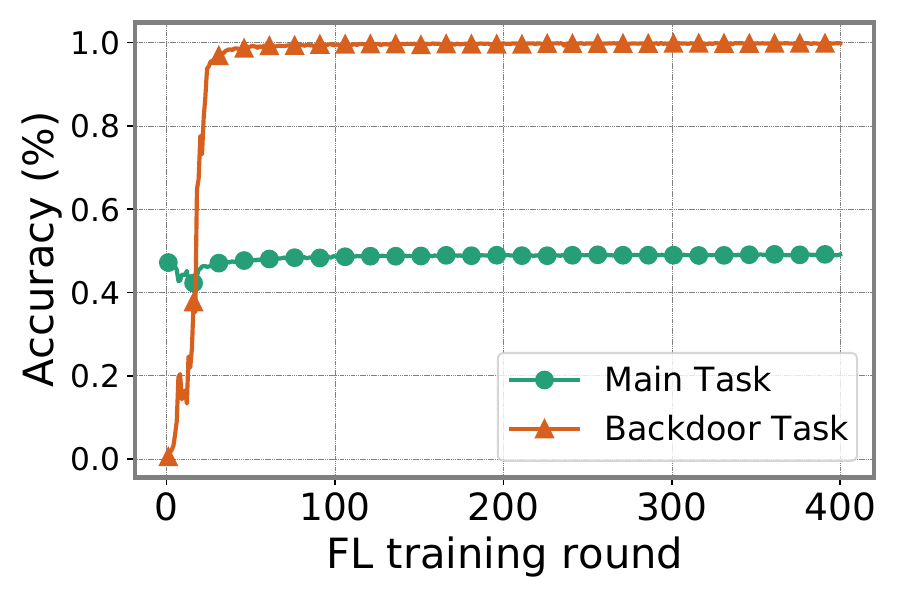}
    \caption{IID-Median}
    \end{subfigure}
    \begin{subfigure}{0.16\textwidth}
    \includegraphics[width=\linewidth]{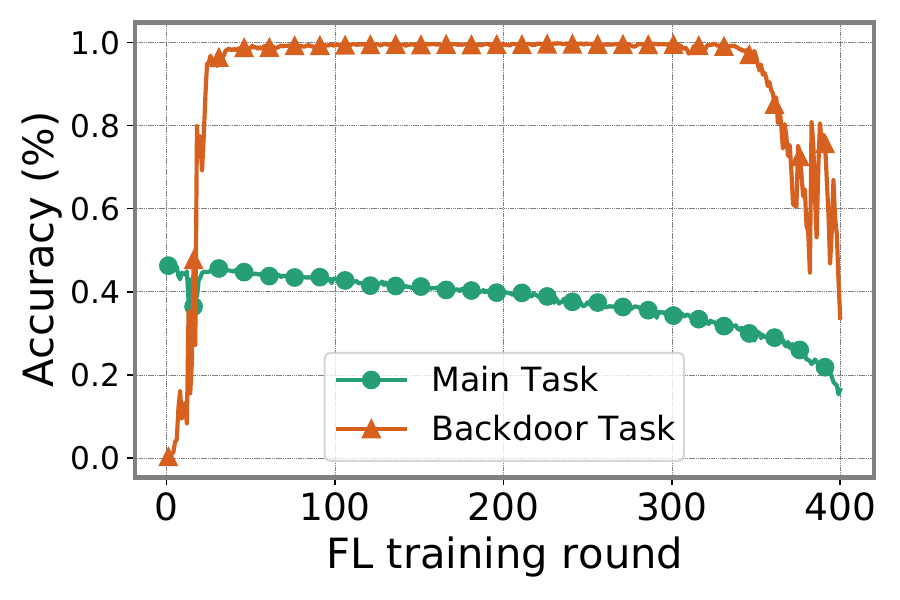}
    \caption{IID-FLAME}
    \end{subfigure}
    \begin{subfigure}{0.16\textwidth}
    \includegraphics[width=\linewidth]{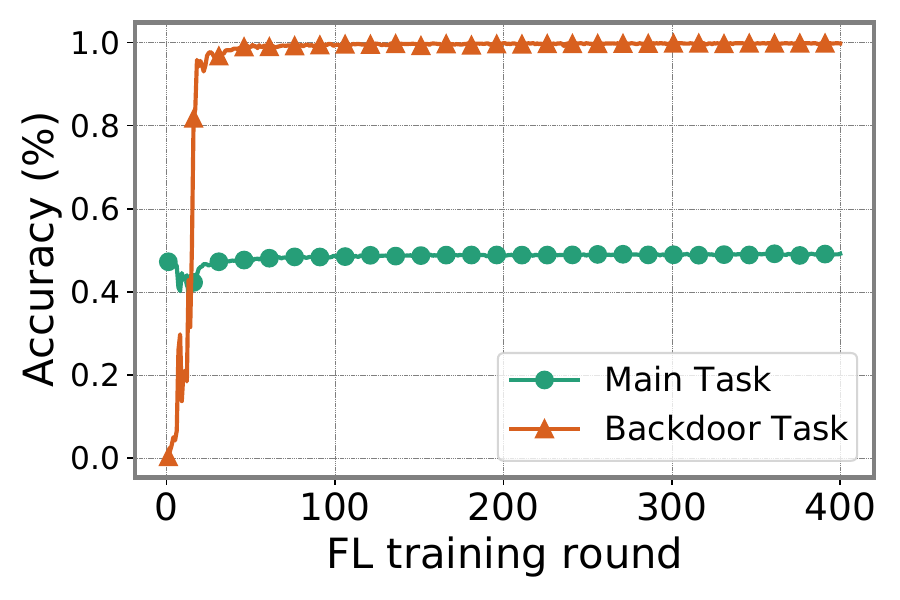}
    \caption{IID-FreqFed}
    \end{subfigure}
    \hfill
    \begin{subfigure}{0.16\textwidth}
    \includegraphics[width=\linewidth]{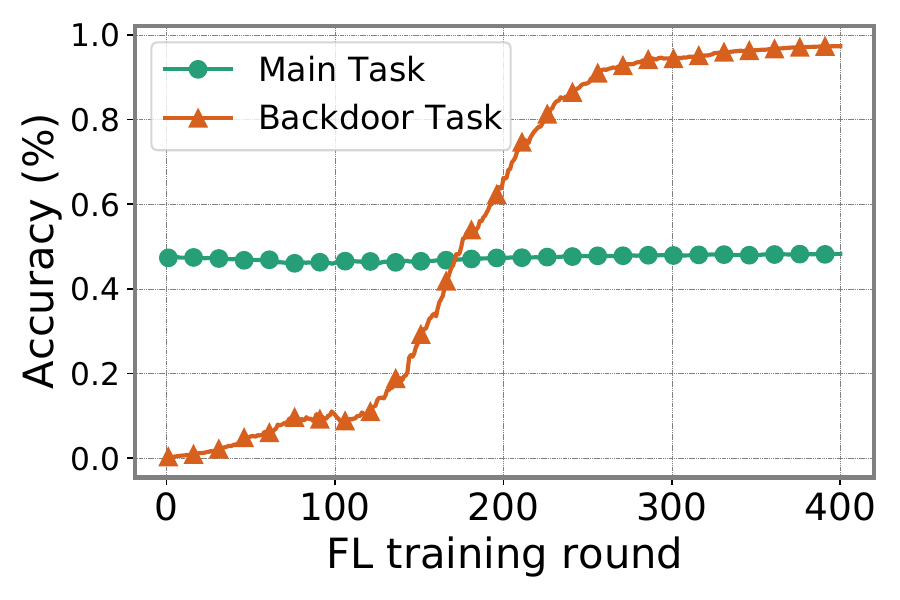}
    \caption{Non-NDC}
    \end{subfigure}
    \begin{subfigure}{0.16\textwidth}
    \includegraphics[width=\linewidth]{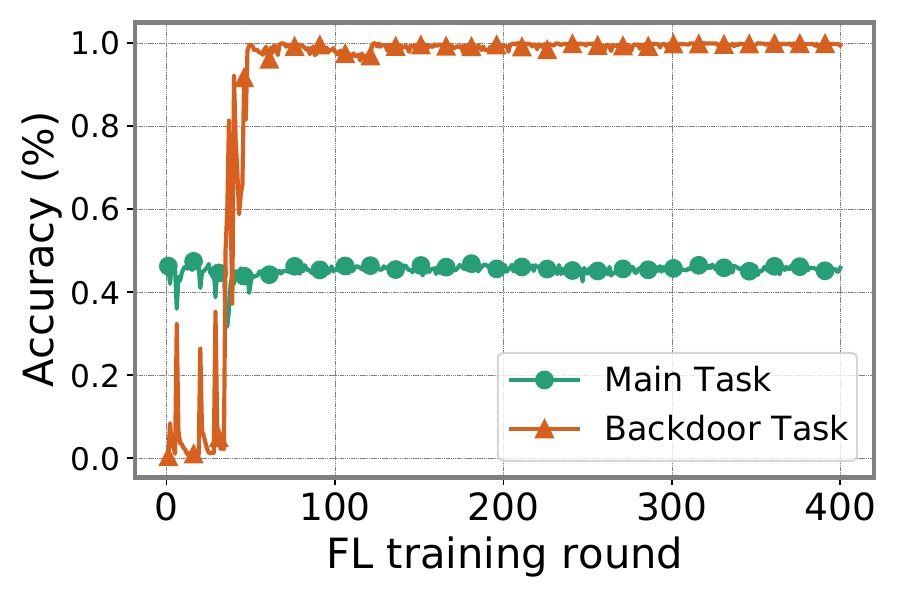}
    \caption{Non-Krum}
    \end{subfigure}
    \begin{subfigure}{0.16\textwidth}
    \includegraphics[width=\linewidth]{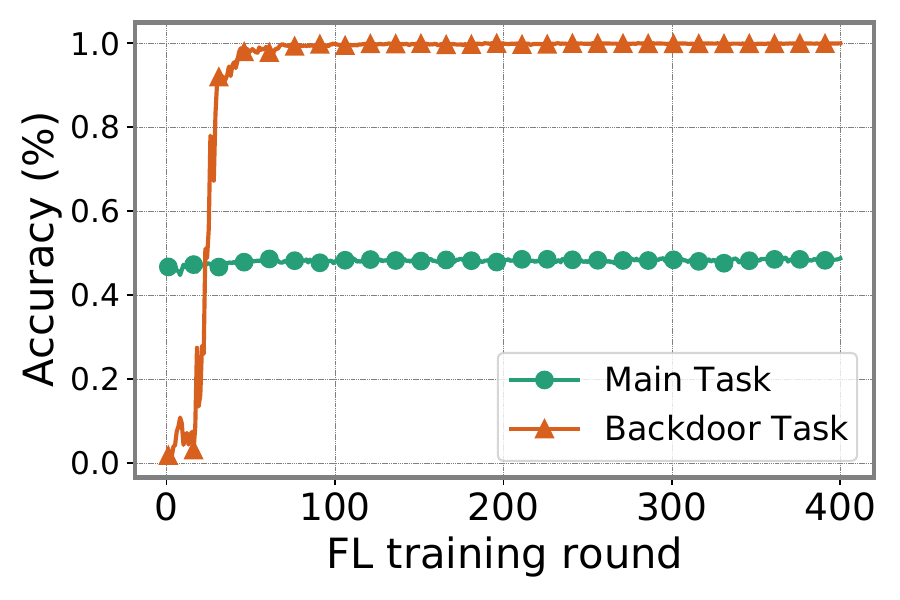}
    \caption{Non-MKrum}
    \end{subfigure}
    \begin{subfigure}{0.16\textwidth}
    \includegraphics[width=\linewidth]{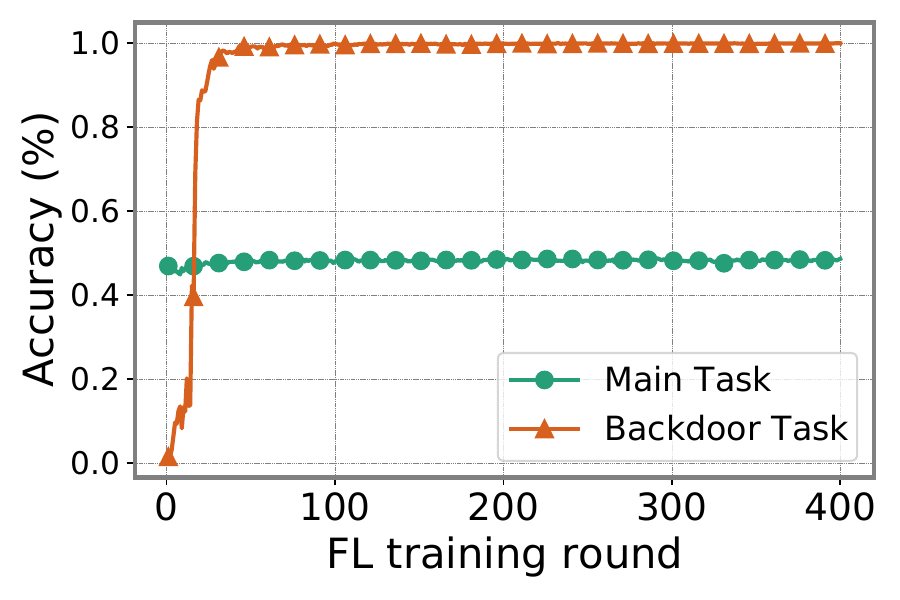}
    \caption{Non-Median}
    \end{subfigure}
    \begin{subfigure}{0.16\textwidth}
    \includegraphics[width=\linewidth]{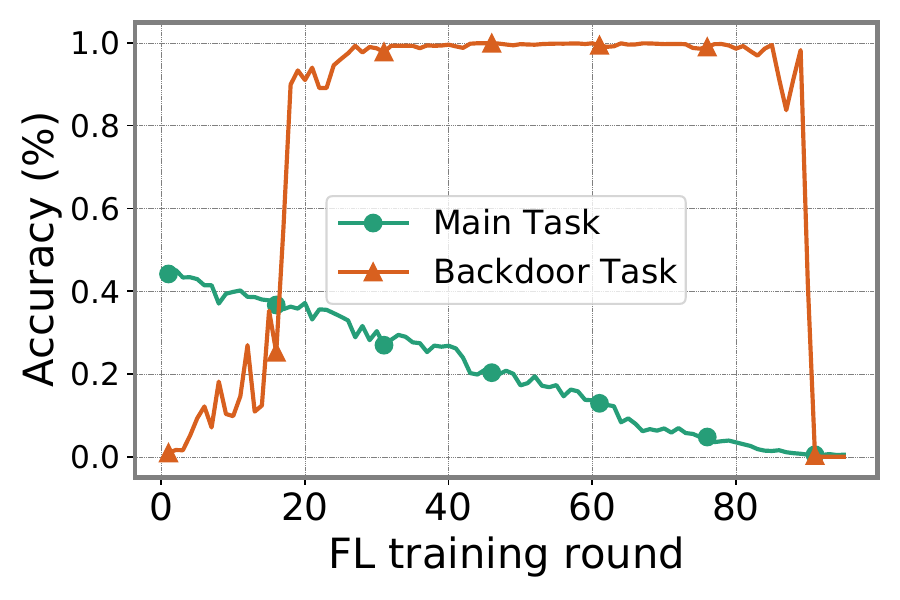}
    \caption{Non-FLAME}
    \end{subfigure}
    \begin{subfigure}{0.16\textwidth}
    \includegraphics[width=\linewidth]{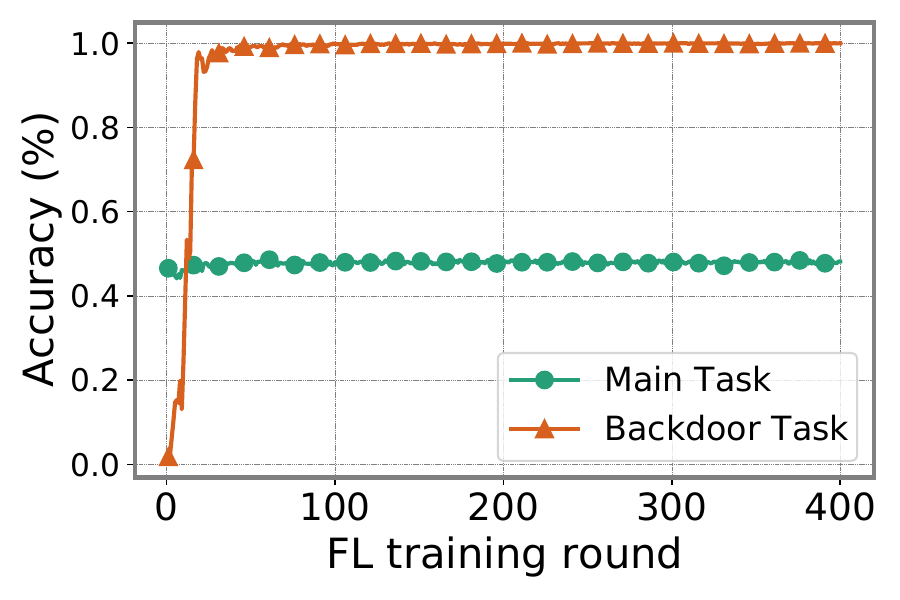}
    \caption{Non-FreqFed}
    \end{subfigure}
    \caption{The performance of EDBA under the fixed pool setting with 25\% compromised clients.}
    \label{defense_EDBA_TINY_percent25}
    \end{figure*}

    \subsection{Performance under Different Compromised Ratios}
    To further investigate the impact of attacker proportion, we evaluate EDBA on the Tiny-ImageNet dataset under the fixed-pool setting with different ratios of compromised clients (10\%, 15\%, and 25\%). The results are presented in Figs.~\ref{defense_EDBA_CIFAR10_percent10}–\ref{defense_EDBA_TINY_percent25}. 

    As the proportion of compromised clients increases, both the effectiveness and persistence of the implanted backdoor are enhanced. With 10\% malicious clients, EDBA already achieves strong BA while maintaining stable MA. At higher ratios such as 15\% and 25\%, BA rapidly converges to nearly 100\% and remains stable, highlighting the scalability of our attack. Second, the persistence of EDBA improves with larger attacker coalitions. Even after several training, the injected backdoor is not diluted by honest updates, suggesting that once the adversarial ratio exceeds a small threshold, the backdoor becomes effectively irreversible under standard aggregation. In practical FL deployments, it is often assumed that limiting the number of compromised clients is sufficient to preserve robustness. Our findings challenge this assumption: even with as few as 10\% attackers, EDBA can reliably implant durable backdoors, and the threat only amplifies as the attacker pool grows. This underscores the need for defenses that can detect and mitigate backdoors early, before adversaries accumulate sufficient influence in the aggregation process.

    \begin{figure*}[htbp]
    \centering
    \begin{subfigure}[b]{0.36\textwidth}
        \centering
        \raisebox{0.12\height}{\includegraphics[width=\linewidth]{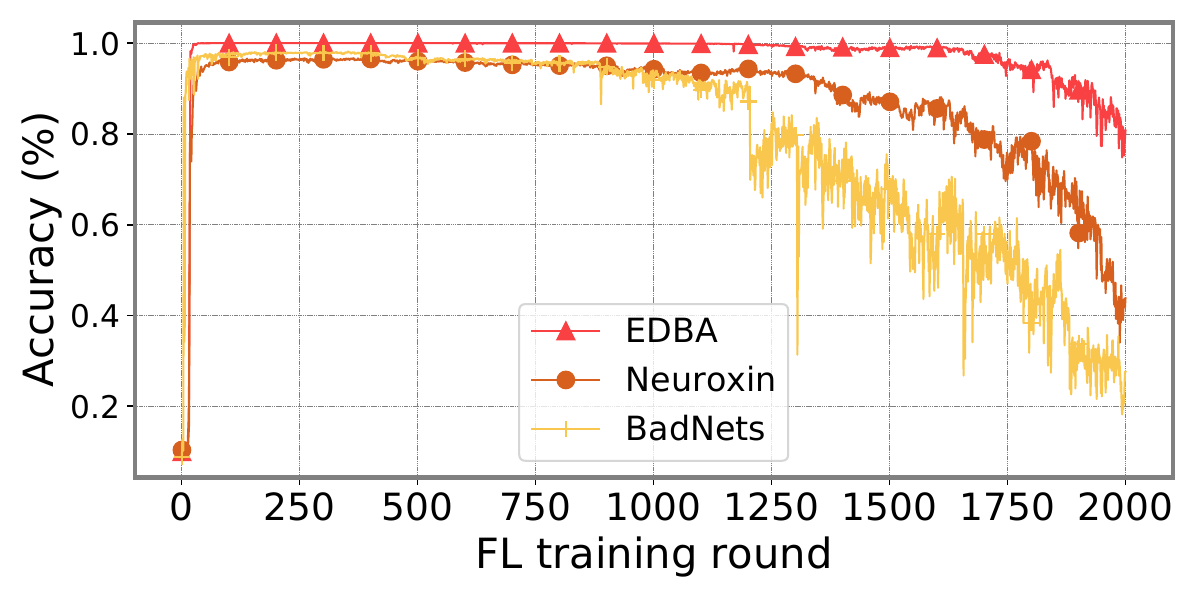}}
        \caption{Lifespan on CIFAR-10}
        \label{Durability}
    \end{subfigure}
    \hfill
    \begin{subfigure}[b]{0.31\textwidth}
        \centering
        \includegraphics[width=\linewidth]{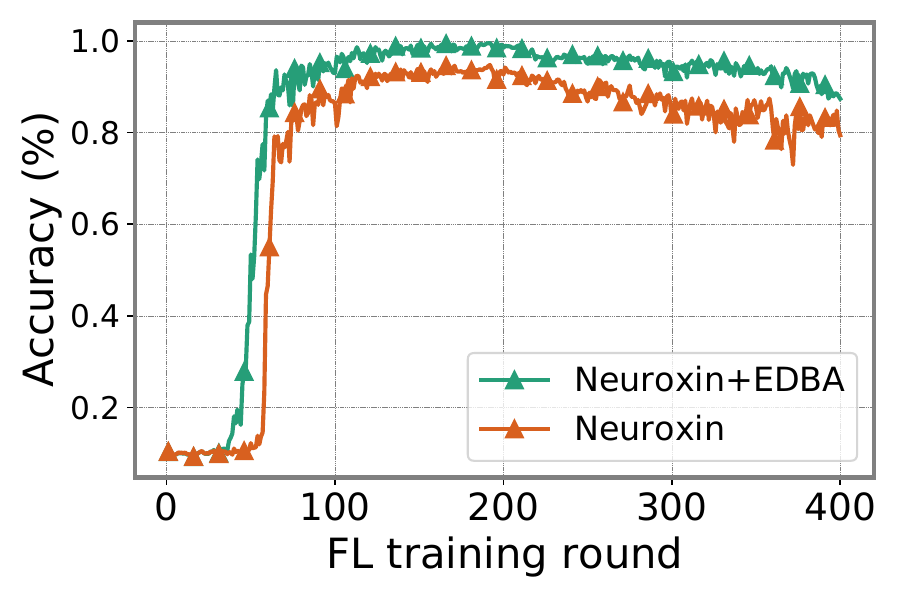}
        \caption{Integration with Neuroxin}
        \label{Integration_neuro}
    \end{subfigure}
    \begin{subfigure}[b]{0.31\textwidth}
        \centering
        \includegraphics[width=\linewidth]{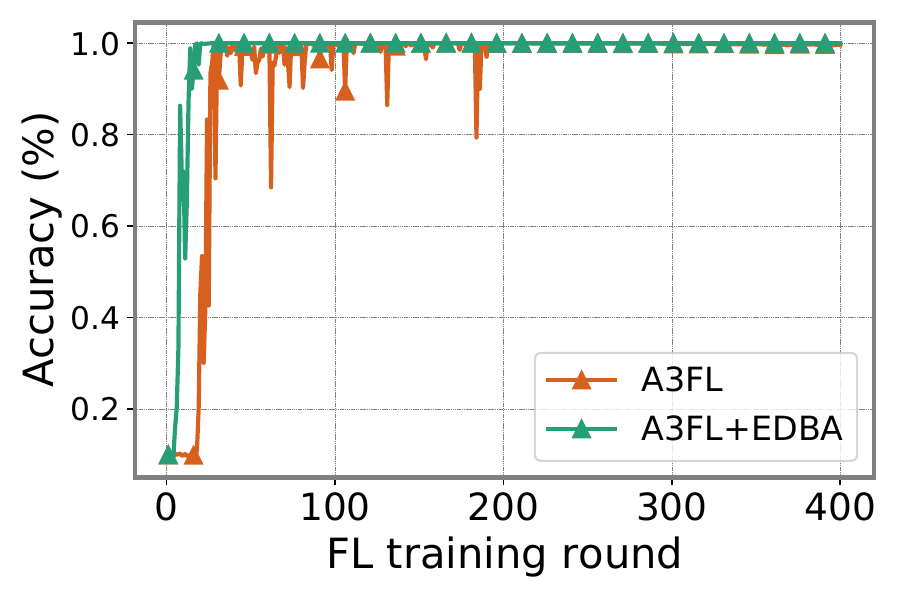}
        \caption{Integration with A3FL}
        \label{Integration_a3fl}
    \end{subfigure}
    \caption{(a) shows the EDBA has a longer lifespan compared with other baseline methods, with attackers removed from 400 epochs. (b) and (c) show the performance of EDBA integrated into the Neuroxin and A3FL training frameworks.}
    \end{figure*}
    
    \subsection{Durability Evaluation}
    Beyond BA and MA, durability is a critical property of backdoors in FL, since malicious clients may only participate temporarily. To evaluate this, we conduct experiments on the Non-IID CIFAR10 dataset under the FLAME defense. Malicious clients are allowed to participate during the first 400 communication rounds, after which they are removed, and the training continues solely with honest clients. This setup enables us to evaluate whether the implanted backdoor can persist without ongoing adversarial intervention. As shown in Fig.~\ref{Durability}, EDBA exhibits remarkable durability. Even after the removal of malicious clients, the backdoor accuracy remains high, and the global model does not revert to a clean state. In contrast, baseline methods such as BadNets and Neurotoxin experience a gradual decline in backdoor accuracy. Compared to these baselines, EDBA sustains stable performance for over 1000 communication rounds, highlighting its robustness against dilution by honest updates.  
    
    These results confirm that the backdoor generated by EDBA is not easily eliminated once implanted. By dynamically optimizing the trigger and decoupling the backdoor from the main task, EDBA creates malicious behaviors that are more persistent than those of prior methods. From a security perspective, this durability implies that even short-lived adversarial participation can introduce long-term vulnerabilities in federated models, underscoring the difficulty of defending against such attacks.
	
\subsection{Integration with Existing Attack Frameworks}
Many prior works focus either on designing effective trigger patterns or on proposing specialized training frameworks to improve backdoor persistence. In contrast, EDBA targets the separation between the main and backdoor tasks via dynamic trigger optimization, which is inherently modular and can be combined with existing attack techniques. To validate this claim, we integrate our trigger-generation module into two representative training frameworks: Neurotoxin~\cite{zhang2022neurotoxin} (implanting backdoors into parameters that are updated infrequently) and A3FL~\cite{zhang2024a3fl} (imitating global training dynamics to improve backdoor efficacy). In these experiments, adversaries are allowed to attack only during the first 200 communication rounds.

The results are shown in Figs.~\ref{Integration_neuro} and~\ref{Integration_a3fl}. When combined with Neurotoxin, EDBA's trigger module noticeably increases both ASR and durability compared to Neurotoxin alone, while preserving MA. Similarly, augmenting A3FL with our trigger-generation process yields faster backdoor convergence and higher sustained BA throughout subsequent rounds. These improvements indicate that EDBA's trigger optimization is compatible with and complementary to different injection strategies. By producing triggers that better decouple the backdoor objective from the main task, EDBA enhances the effectiveness and durability of a wide range of existing attacks without requiring additional attacker resources.

The ease of integration is concerning because defenders cannot assume that improvements in trigger design and improvements in injection frameworks are independent. Adversaries can modularly combine advances to create substantially more persistent and robust backdoors. 

\begin{figure}[ht]
\centering
\begin{subfigure}{0.15\textwidth}
    \includegraphics[width=\linewidth]{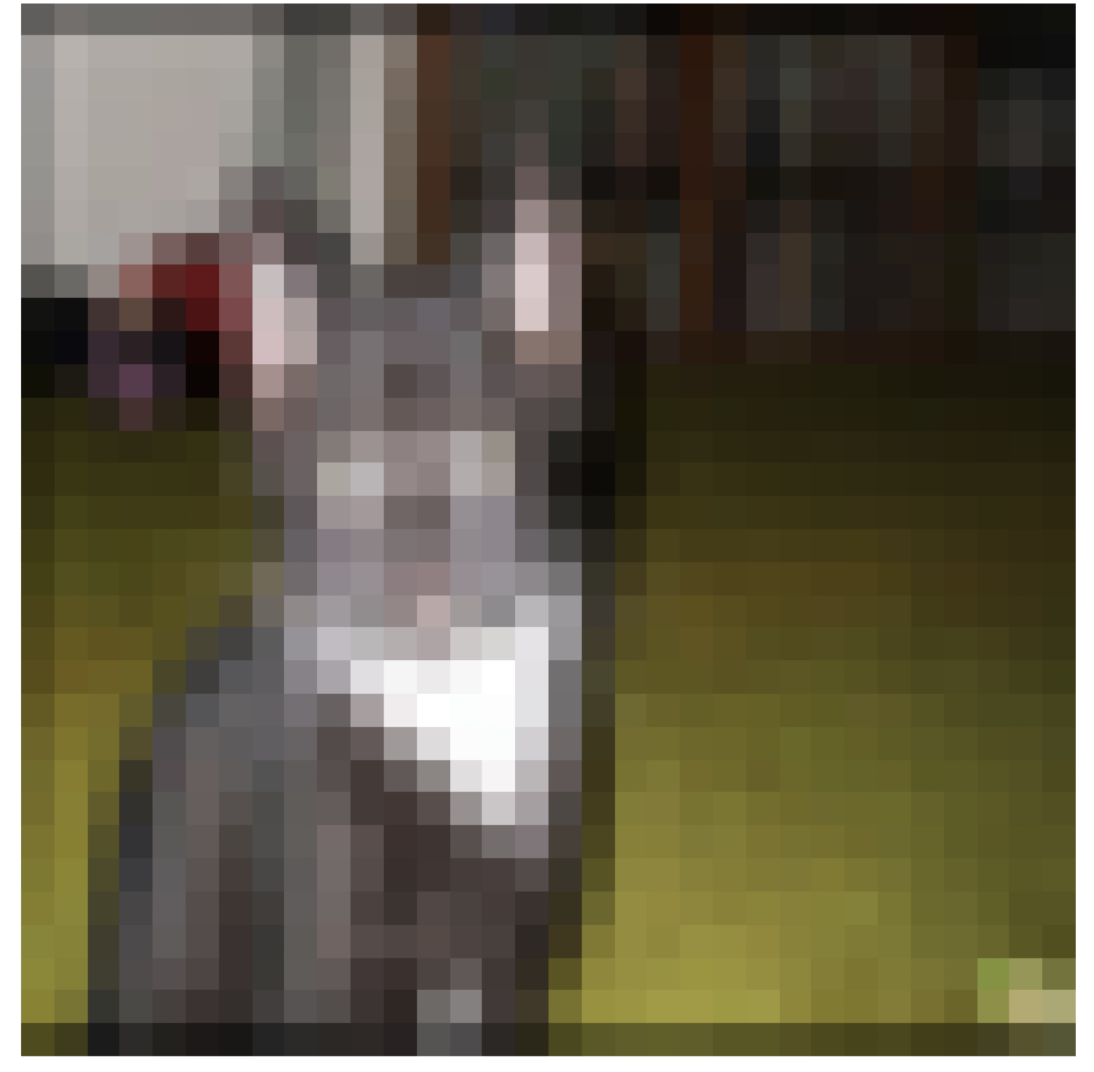}
    \caption{EDBA}
    \label{EDBA_image}
\end{subfigure}
\hfill
\begin{subfigure}{0.15\textwidth}
    \includegraphics[width=\linewidth]{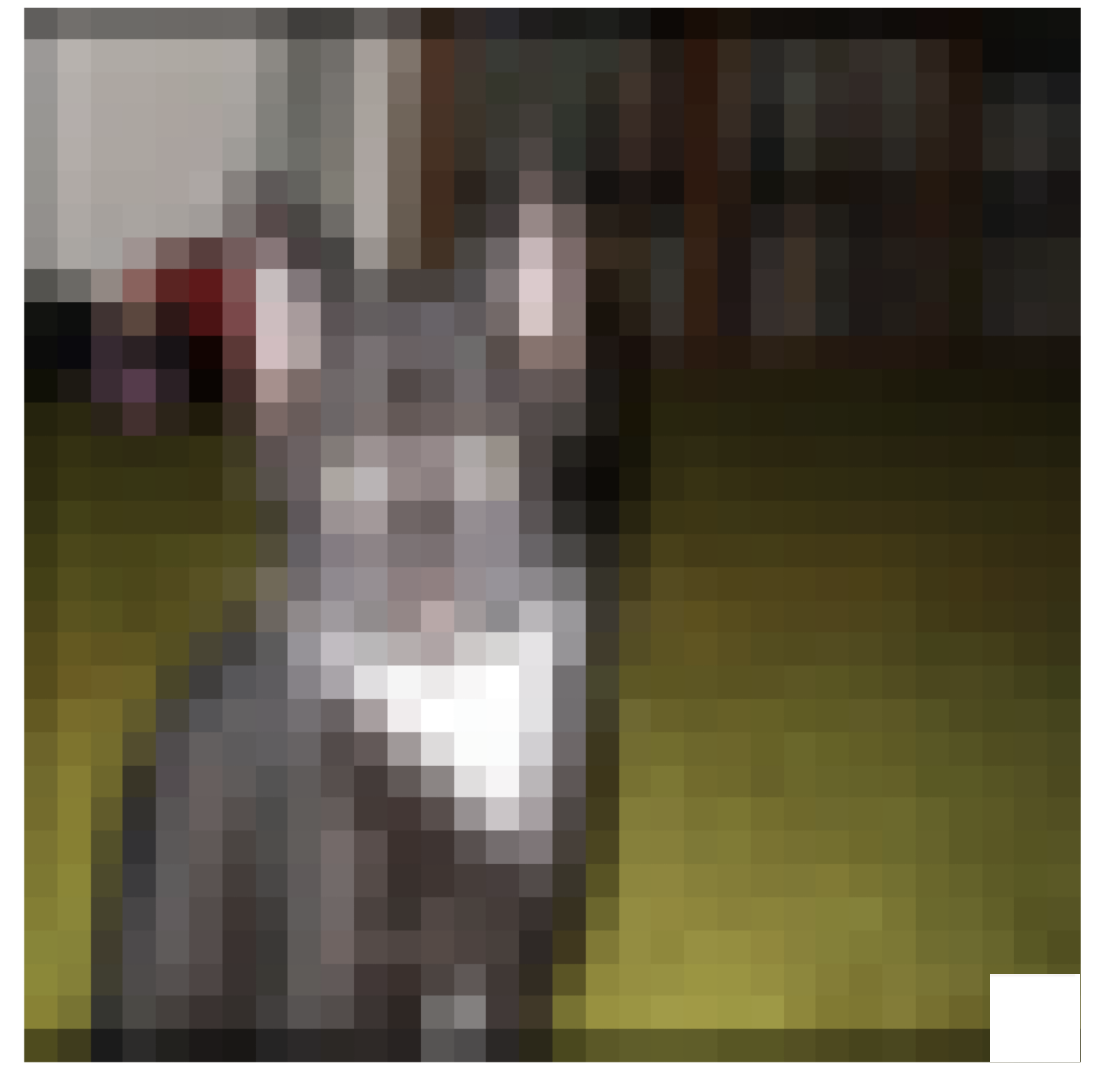}
    \caption{Pixel}
    \label{pixel}
\end{subfigure}
\hfill
\begin{subfigure}{0.15\textwidth}
    \includegraphics[width=\linewidth]{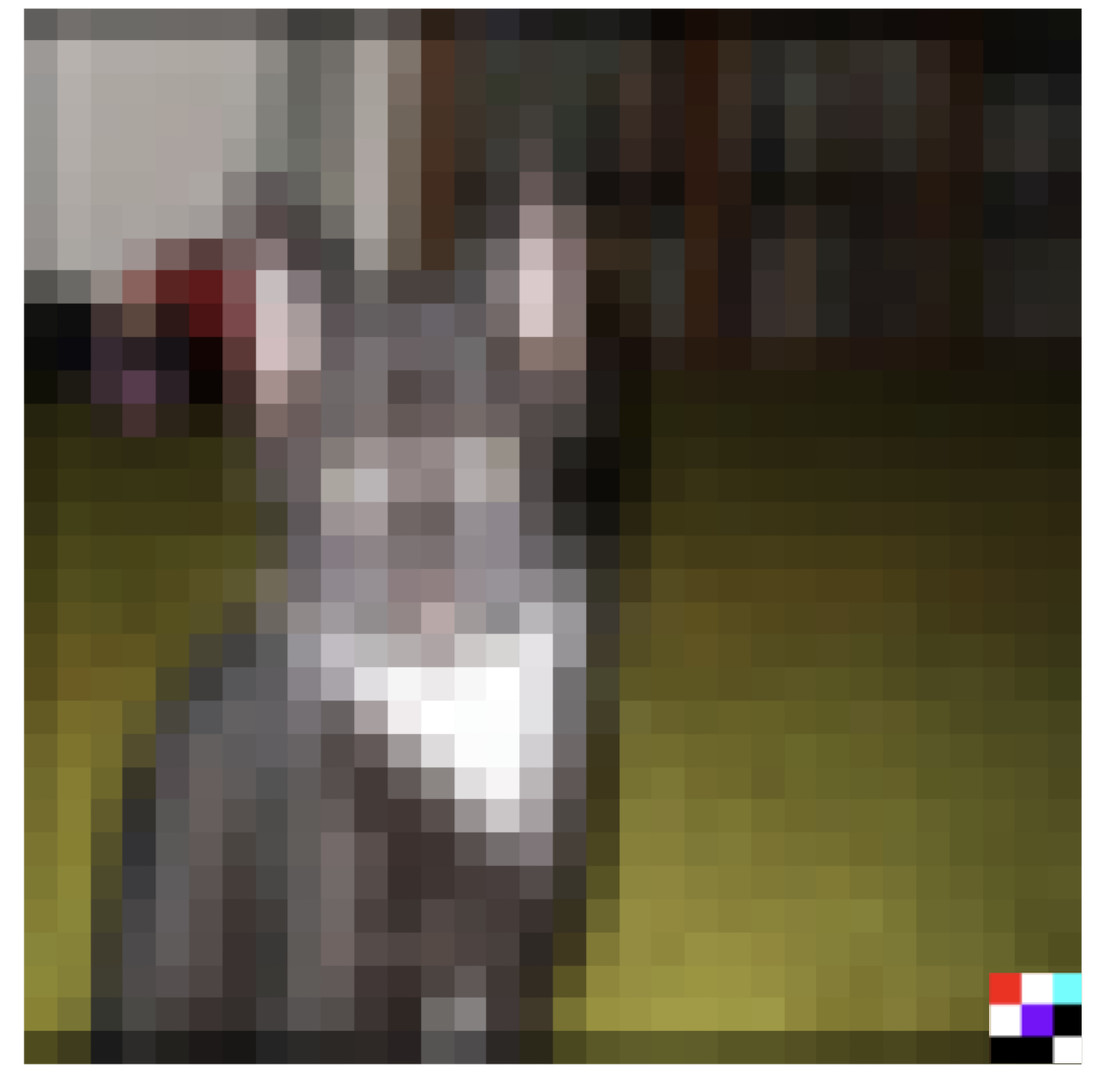}
    \caption{PGD}
    \label{pgd}
\end{subfigure}
\caption{Visualization of backdoor samples under different trigger generation methods.}
\label{visual}
\end{figure}

\subsection{Visualization of Benign and Backdoor Samples}
Fig.~\ref{visual} shows representative backdoor samples produced by three trigger-generation strategies: (\subref{EDBA_image}) for the EDBA, (\subref{pixel}) for the fixed-pixel patch, and (\subref{pgd}) for the PGD-style adversarial perturbation. The Pixel baseline inserts an explicit visible patch, while PGD applies adversarial-style noise aimed at flipping the classifier's decision. By contrast, EDBA produces input modifications that appear neither as a conspicuous patch nor as pure high-frequency adversarial noise. Instead, the EDBA perturbations are tailored to the model and thus employ subtle changes that are visually less distinctive than a fixed patch. Compared to adversarial-style perturbations, EDBA modifications tend to align more with the model's internal features, reflecting the method's goal of maximizing the distinction between backdoor and main tasks in representation space rather than merely forcing a misclassification.

\section{Security Implications and Discussion} \label{sec:discussion}
The results in this paper highlight that existing defenses in FL are insufficient against advanced backdoor attacks such as EDBA. Our dynamic trigger optimization and task-decoupling design enables the backdoor to persist even under widely adopted aggregation defenses, which raises several important security implications.

\paragraph{Limitations of existing defenses.} 
Traditional defenses such as norm clipping, robust aggregation, and anomaly detection assume that malicious updates will be statistically distinguishable from benign ones. As our experiments show, these assumptions do not hold once the backdoor is optimized to resemble benign updates. Behavior-based detection methods such as FLDetector~\cite{zhang2022fldetector}, which track the consistency of user updates across rounds, provide a more promising direction. However, direct application of FLDetector is less effective against EDBA, since the backdoor is successfully injected before the malicious participants are flagged and excluded.

\paragraph{Future directions.} 
To address this gap, we suggest complementing behavior-score tracking with model parameter archiving. By storing historical versions of aggregated models, the server can retrospectively remove the influence of malicious updates if a client is later identified as adversarial. Our preliminary exploration indicates that this rollback mechanism can effectively mitigate persistent backdoors introduced by EDBA. Nevertheless, this approach incurs non-trivial overhead, including storage costs for archived models and additional computation to revert model states, raising practical challenges for large-scale FL deployments. Addressing this challenge requires not only stronger defenses, but also new paradigms for model auditing and verification that can operate effectively under practical FL constraints.
	
\section{Conclusion} \label{conclusion}
In this work, we revisited the durability and effectiveness of backdoor attacks in FL and attribute it to the coupling between the main and backdoor tasks. To address this, we proposed EDBA, a dynamic min–max trigger optimization framework that explicitly maximizes the distinction between the two tasks. Unlike fixed-pattern or adversarial-perturbation triggers, EDBA enforces that the poisoned input produces an output different from the original prediction, thereby preventing benign updates from diluting the backdoor. Extensive experiments on both computer vision and natural language processing tasks demonstrate that EDBA consistently achieves higher attack success rates and stronger persistence than six state-of-the-art baselines, and remains effective under six widely used FL defenses. These findings reveal that task-decoupled backdoors constitute a more severe and persistent threat than previously recognized, highlighting blind spots in current defense assumptions and motivating the design of more robust detection and mitigation strategies.  

Despite these promising results, our framework has limitations. The inner maximization for dynamic trigger generation introduces additional computational overhead. Future work will explore more efficient optimization techniques to reduce the cost of trigger generation, and another important direction is to study joint defense, particularly how behavior-tracking and rollback mechanisms could be adapted to mitigate durable and modular attacks such as EDBA.  

Overall, our study demonstrates that separating main and backdoor tasks is a powerful paradigm for adversaries in FL, and underscores the urgent need for the community to develop defenses that explicitly consider persistence and task decoupling in federated environments.

\bibliographystyle{IEEEtran}
\bibliography{reference.bib}

\begin{thebibliography}{10}
\providecommand{\url}[1]{#1}
\csname url@samestyle\endcsname
\providecommand{\newblock}{\relax}
\providecommand{\bibinfo}[2]{#2}
\providecommand{\BIBentrySTDinterwordspacing}{\spaceskip=0pt\relax}
\providecommand{\BIBentryALTinterwordstretchfactor}{4}
\providecommand{\BIBentryALTinterwordspacing}{\spaceskip=\fontdimen2\font plus
\BIBentryALTinterwordstretchfactor\fontdimen3\font minus \fontdimen4\font\relax}
\providecommand{\BIBforeignlanguage}[2]{{%
\expandafter\ifx\csname l@#1\endcsname\relax
\typeout{** WARNING: IEEEtran.bst: No hyphenation pattern has been}%
\typeout{** loaded for the language `#1'. Using the pattern for}%
\typeout{** the default language instead.}%
\else
\language=\csname l@#1\endcsname
\fi
#2}}
\providecommand{\BIBdecl}{\relax}
\BIBdecl

\bibitem{mcmahan2017communication}
B.~McMahan, E.~Moore, D.~Ramage, S.~Hampson, and B.~A. y~Arcas, ``Communication-efficient learning of deep networks from decentralized data,'' in \emph{Artificial intelligence and statistics}.\hskip 1em plus 0.5em minus 0.4em\relax PMLR, 2017, pp. 1273--1282.

\bibitem{xue2023differentially}
R.~Xue, K.~Xue, B.~Zhu, X.~Luo, T.~Zhang, Q.~Sun, and J.~Lu, ``Differentially private federated learning with an adaptive noise mechanism,'' \emph{IEEE Transactions on Information Forensics and Security}, vol.~19, pp. 74--87, 2023.

\bibitem{chen2023privacy}
Z.~Chen, S.~Yu, M.~Fan, X.~Liu, and R.~H. Deng, ``Privacy-enhancing and robust backdoor defense for federated learning on heterogeneous data,'' \emph{IEEE Transactions on Information Forensics and Security}, vol.~19, pp. 693--707, 2023.

\bibitem{yazdinejad2024robust}
A.~Yazdinejad, A.~Dehghantanha, H.~Karimipour, G.~Srivastava, and R.~M. Parizi, ``A robust privacy-preserving federated learning model against model poisoning attacks,'' \emph{IEEE Transactions on Information Forensics and Security}, vol.~19, pp. 6693--6708, 2024.

\bibitem{konevcny2016federated}
J.~Kone{\v{c}}n{\`y}, H.~B. McMahan, D.~Ramage, and P.~Richt{\'a}rik, ``Federated optimization: Distributed machine learning for on-device intelligence,'' \emph{arXiv preprint arXiv:1610.02527}, 2016.

\bibitem{mcmahan2016federated}
H.~B. McMahan, F.~Yu, P.~Richtarik, A.~Suresh, D.~Bacon \emph{et~al.}, ``Federated learning: Strategies for improving communication efficiency,'' in \emph{Proceedings of the 29th Conference on Neural Information Processing Systems (NIPS), Barcelona, Spain}, 2016, pp. 5--10.

\bibitem{yang2019federated}
Q.~Yang, Y.~Liu, T.~Chen, and Y.~Tong, ``Federated machine learning: Concept and applications,'' \emph{ACM Transactions on Intelligent Systems and Technology (TIST)}, vol.~10, no.~2, pp. 1--19, 2019.

\bibitem{lyu2020threats}
L.~Lyu, H.~Yu, and Q.~Yang, ``Threats to federated learning: A survey,'' \emph{arXiv preprint arXiv:2003.02133}, 2020.

\bibitem{kairouz2021advances}
P.~Kairouz, H.~B. McMahan, B.~Avent, A.~Bellet, M.~Bennis, A.~N. Bhagoji, K.~Bonawitz, Z.~Charles, G.~Cormode, R.~Cummings \emph{et~al.}, ``Advances and open problems in federated learning,'' \emph{Foundations and trends{\textregistered} in machine learning}, vol.~14, no. 1--2, pp. 1--210, 2021.

\bibitem{rodriguez2023survey}
N.~Rodr{\'\i}guez-Barroso, D.~Jim{\'e}nez-L{\'o}pez, M.~V. Luz{\'o}n, F.~Herrera, and E.~Mart{\'\i}nez-C{\'a}mara, ``Survey on federated learning threats: Concepts, taxonomy on attacks and defences, experimental study and challenges,'' \emph{Information Fusion}, vol.~90, pp. 148--173, 2023.

\bibitem{gu2017badnets}
T.~Gu, B.~Dolan-Gavitt, and S.~Garg, ``Badnets: Identifying vulnerabilities in the machine learning model supply chain,'' \emph{arXiv preprint arXiv:1708.06733}, 2017.

\bibitem{szegedy2013intriguing}
C.~Szegedy, W.~Zaremba, I.~Sutskever, J.~Bruna, D.~Erhan, I.~Goodfellow, and R.~Fergus, ``Intriguing properties of neural networks,'' \emph{arXiv preprint arXiv:1312.6199}, 2013.

\bibitem{shokri2017membership}
R.~Shokri, M.~Stronati, C.~Song, and V.~Shmatikov, ``Membership inference attacks against machine learning models,'' in \emph{2017 IEEE symposium on security and privacy (SP)}.\hskip 1em plus 0.5em minus 0.4em\relax IEEE, 2017, pp. 3--18.

\bibitem{liu2025token}
J.~Liu, Z.~Wang, H.~Wang, C.~Tian, and Y.~Jin, ``Token-level constraint boundary search for jailbreaking text-to-image models,'' \emph{arXiv preprint arXiv:2504.11106}, 2025.

\bibitem{zhu2019deep}
L.~Zhu, Z.~Liu, and S.~Han, ``Deep leakage from gradients,'' \emph{Advances in neural information processing systems}, vol.~32, 2019.

\bibitem{he2023backdoor}
Y.~He, Z.~Shen, J.~Hua, Q.~Dong, J.~Niu, W.~Tong, X.~Huang, C.~Li, and S.~Zhong, ``Backdoor attack against split neural network-based vertical federated learning,'' \emph{IEEE Transactions on Information Forensics and Security}, vol.~19, pp. 748--763, 2023.

\bibitem{fang2023vulnerability}
P.~Fang and J.~Chen, ``On the vulnerability of backdoor defenses for federated learning,'' in \emph{Proceedings of the AAAI Conference on Artificial Intelligence}, vol.~37, no.~10, 2023, pp. 11\,800--11\,808.

\bibitem{cramer2015secure}
R.~Cramer, I.~B. Damg{\aa}rd \emph{et~al.}, \emph{Secure multiparty computation}.\hskip 1em plus 0.5em minus 0.4em\relax Cambridge University Press, 2015.

\bibitem{bonawitz2017practical}
K.~Bonawitz, V.~Ivanov, B.~Kreuter, A.~Marcedone, H.~B. McMahan, S.~Patel, D.~Ramage, A.~Segal, and K.~Seth, ``Practical secure aggregation for privacy-preserving machine learning,'' in \emph{proceedings of the 2017 ACM SIGSAC Conference on Computer and Communications Security}, 2017, pp. 1175--1191.

\bibitem{gong2023multi}
M.~Gong, Y.~Zhang, Y.~Gao, A.~K. Qin, Y.~Wu, S.~Wang, and Y.~Zhang, ``A multi-modal vertical federated learning framework based on homomorphic encryption,'' \emph{IEEE Transactions on Information Forensics and Security}, vol.~19, pp. 1826--1839, 2023.

\bibitem{bagdasaryan2020backdoor}
E.~Bagdasaryan, A.~Veit, Y.~Hua, D.~Estrin, and V.~Shmatikov, ``How to backdoor federated learning,'' in \emph{International conference on artificial intelligence and statistics}.\hskip 1em plus 0.5em minus 0.4em\relax PMLR, 2020, pp. 2938--2948.

\bibitem{xie2019dba}
C.~Xie, K.~Huang, P.-Y. Chen, and B.~Li, ``Dba: Distributed backdoor attacks against federated learning,'' in \emph{International conference on learning representations}, 2019.

\bibitem{wang2020attack}
H.~Wang, K.~Sreenivasan, S.~Rajput, H.~Vishwakarma, S.~Agarwal, J.-y. Sohn, K.~Lee, and D.~Papailiopoulos, ``Attack of the tails: Yes, you really can backdoor federated learning,'' \emph{Advances in Neural Information Processing Systems}, vol.~33, pp. 16\,070--16\,084, 2020.

\bibitem{blanchard2017machine}
P.~Blanchard, E.~M. El~Mhamdi, R.~Guerraoui, and J.~Stainer, ``Machine learning with adversaries: Byzantine tolerant gradient descent,'' \emph{Advances in neural information processing systems}, vol.~30, 2017.

\bibitem{pillutla2022robust}
K.~Pillutla, S.~M. Kakade, and Z.~Harchaoui, ``Robust aggregation for federated learning,'' \emph{IEEE Transactions on Signal Processing}, vol.~70, pp. 1142--1154, 2022.

\bibitem{sun2019can}
Z.~Sun, P.~Kairouz, A.~T. Suresh, and H.~B. McMahan, ``Can you really backdoor federated learning?'' \emph{arXiv preprint arXiv:1911.07963}, 2019.

\bibitem{nguyen2022flame}
T.~D. Nguyen, P.~Rieger, R.~De~Viti, H.~Chen, B.~B. Brandenburg, H.~Yalame, H.~M{\"o}llering, H.~Fereidooni, S.~Marchal, M.~Miettinen \emph{et~al.}, ``$\{$FLAME$\}$: Taming backdoors in federated learning,'' in \emph{31st USENIX Security Symposium (USENIX Security 22)}, 2022, pp. 1415--1432.

\bibitem{gong2023agramplifier}
Z.~Gong, L.~Shen, Y.~Zhang, L.~Y. Zhang, J.~Wang, G.~Bai, and Y.~Xiang, ``Agramplifier: Defending federated learning against poisoning attacks through local update amplification,'' \emph{IEEE Transactions on Information Forensics and Security}, vol.~19, pp. 1241--1250, 2023.

\bibitem{zhang2024flpurifier}
J.~Zhang, C.~Zhu, X.~Sun, C.~Ge, B.~Chen, W.~Susilo, and S.~Yu, ``Flpurifier: Backdoor defense in federated learning via decoupled contrastive training,'' \emph{IEEE Transactions on Information Forensics and Security}, vol.~19, pp. 4752--4766, 2024.

\bibitem{zhang2022neurotoxin}
Z.~Zhang, A.~Panda, L.~Song, Y.~Yang, M.~Mahoney, P.~Mittal, R.~Kannan, and J.~Gonzalez, ``Neurotoxin: Durable backdoors in federated learning,'' in \emph{International Conference on Machine Learning}.\hskip 1em plus 0.5em minus 0.4em\relax PMLR, 2022, pp. 26\,429--26\,446.

\bibitem{nguyen2024iba}
T.~D. Nguyen, T.~A. Nguyen, A.~Tran, K.~D. Doan, and K.-S. Wong, ``Iba: Towards irreversible backdoor attacks in federated learning,'' \emph{Advances in Neural Information Processing Systems}, vol.~36, 2024.

\bibitem{zhang2024a3fl}
H.~Zhang, J.~Jia, J.~Chen, L.~Lin, and D.~Wu, ``A3fl: Adversarially adaptive backdoor attacks to federated learning,'' \emph{Advances in Neural Information Processing Systems}, vol.~36, 2024.

\bibitem{li2023federated}
X.~Li, Z.~Song, and J.~Yang, ``Federated adversarial learning: A framework with convergence analysis,'' in \emph{International Conference on Machine Learning}.\hskip 1em plus 0.5em minus 0.4em\relax PMLR, 2023, pp. 19\,932--19\,959.

\bibitem{tan2022towards}
A.~Z. Tan, H.~Yu, L.~Cui, and Q.~Yang, ``Towards personalized federated learning,'' \emph{IEEE Transactions on Neural Networks and Learning Systems}, 2022.

\bibitem{karimireddy2020scaffold}
S.~P. Karimireddy, S.~Kale, M.~Mohri, S.~Reddi, S.~Stich, and A.~T. Suresh, ``Scaffold: Stochastic controlled averaging for federated learning,'' in \emph{International conference on machine learning}.\hskip 1em plus 0.5em minus 0.4em\relax PMLR, 2020, pp. 5132--5143.

\bibitem{zhu2019multi}
H.~Zhu and Y.~Jin, ``Multi-objective evolutionary federated learning,'' \emph{IEEE transactions on neural networks and learning systems}, vol.~31, no.~4, pp. 1310--1322, 2019.

\bibitem{tolpegin2020data}
V.~Tolpegin, S.~Truex, M.~E. Gursoy, and L.~Liu, ``Data poisoning attacks against federated learning systems,'' in \emph{Computer Security--ESORICS 2020: 25th European Symposium on Research in Computer Security, ESORICS 2020, Guildford, UK, September 14--18, 2020, Proceedings, Part I 25}.\hskip 1em plus 0.5em minus 0.4em\relax Springer, 2020, pp. 480--501.

\bibitem{baruch2019little}
G.~Baruch, M.~Baruch, and Y.~Goldberg, ``A little is enough: Circumventing defenses for distributed learning,'' \emph{Advances in Neural Information Processing Systems}, vol.~32, 2019.

\bibitem{lyu2023poisoning}
X.~Lyu, Y.~Han, W.~Wang, J.~Liu, B.~Wang, J.~Liu, and X.~Zhang, ``Poisoning with cerberus: Stealthy and colluded backdoor attack against federated learning,'' in \emph{Proceedings of the AAAI Conference on Artificial Intelligence}, vol.~37, no.~7, 2023, pp. 9020--9028.

\bibitem{lyu2022privacy}
L.~Lyu, H.~Yu, X.~Ma, C.~Chen, L.~Sun, J.~Zhao, Q.~Yang, and S.~Y. Philip, ``Privacy and robustness in federated learning: Attacks and defenses,'' \emph{IEEE transactions on neural networks and learning systems}, 2022.

\bibitem{huang2019neuroninspect}
X.~Huang, M.~Alzantot, and M.~Srivastava, ``Neuroninspect: Detecting backdoors in neural networks via output explanations,'' \emph{arXiv preprint arXiv:1911.07399}, 2019.

\bibitem{hou2021mitigating}
B.~Hou, J.~Gao, X.~Guo, T.~Baker, Y.~Zhang, Y.~Wen, and Z.~Liu, ``Mitigating the backdoor attack by federated filters for industrial iot applications,'' \emph{IEEE Transactions on Industrial Informatics}, vol.~18, no.~5, pp. 3562--3571, 2021.

\bibitem{nasr2018comprehensive}
M.~Nasr, R.~Shokri, and A.~Houmansadr, ``Comprehensive privacy analysis of deep learning,'' in \emph{Proceedings of the 2019 IEEE Symposium on Security and Privacy (SP)}, vol. 2018, 2018, pp. 1--15.

\bibitem{liu2021federaser}
G.~Liu, X.~Ma, Y.~Yang, C.~Wang, and J.~Liu, ``Federaser: Enabling efficient client-level data removal from federated learning models,'' in \emph{2021 IEEE/ACM 29th International Symposium on Quality of Service (IWQOS)}.\hskip 1em plus 0.5em minus 0.4em\relax IEEE, 2021, pp. 1--10.

\bibitem{rieger2022deepsight}
P.~Rieger, T.~D. Nguyen, M.~Miettinen, and A.-R. Sadeghi, ``Deepsight: Mitigating backdoor attacks in federated learning through deep model inspection,'' \emph{arXiv preprint arXiv:2201.00763}, 2022.

\bibitem{fung2018mitigating}
C.~Fung, C.~J. Yoon, and I.~Beschastnikh, ``Mitigating sybils in federated learning poisoning,'' \emph{arXiv preprint arXiv:1808.04866}, 2018.

\bibitem{panda2022sparsefed}
A.~Panda, S.~Mahloujifar, A.~N. Bhagoji, S.~Chakraborty, and P.~Mittal, ``Sparsefed: Mitigating model poisoning attacks in federated learning with sparsification,'' in \emph{International Conference on Artificial Intelligence and Statistics}.\hskip 1em plus 0.5em minus 0.4em\relax PMLR, 2022, pp. 7587--7624.

\bibitem{fereidooni2023freqfed}
H.~Fereidooni, A.~Pegoraro, P.~Rieger, A.~Dmitrienko, and A.-R. Sadeghi, ``Freqfed: A frequency analysis-based approach for mitigating poisoning attacks in federated learning,'' \emph{arXiv preprint arXiv:2312.04432}, 2023.

\bibitem{wang2019neural}
B.~Wang, Y.~Yao, S.~Shan, H.~Li, B.~Viswanath, H.~Zheng, and B.~Y. Zhao, ``Neural cleanse: Identifying and mitigating backdoor attacks in neural networks,'' in \emph{2019 IEEE Symposium on Security and Privacy (SP)}.\hskip 1em plus 0.5em minus 0.4em\relax IEEE, 2019, pp. 707--723.

\bibitem{liu2018fine}
K.~Liu, B.~Dolan-Gavitt, and S.~Garg, ``Fine-pruning: Defending against backdooring attacks on deep neural networks,'' in \emph{International symposium on research in attacks, intrusions, and defenses}.\hskip 1em plus 0.5em minus 0.4em\relax Springer, 2018, pp. 273--294.

\bibitem{li2024backdoorindicator}
S.~Li and Y.~Dai, ``Backdoorindicator: Leveraging ood data for proactive backdoor detection in federated learning,'' \emph{arXiv preprint arXiv:2405.20862}, 2024.

\bibitem{alam2022perdoor}
M.~Alam, E.~Sarkar, and M.~Maniatakos, ``Perdoor: Persistent non-uniform backdoors in federated learning using adversarial perturbations,'' \emph{arXiv preprint arXiv:2205.13523}, 2022.

\bibitem{lecun1995learning}
Y.~LeCun, L.~D. Jackel, L.~Bottou, C.~Cortes, J.~S. Denker, H.~Drucker, I.~Guyon, U.~A. Muller, E.~Sackinger, P.~Simard \emph{et~al.}, ``Learning algorithms for classification: A comparison on handwritten digit recognition,'' \emph{Neural networks: the statistical mechanics perspective}, vol. 261, no. 276, p.~2, 1995.

\bibitem{AlexKrizhevsky2009LearningML}
A.~Krizhevsky, G.~Hinton \emph{et~al.}, ``Learning multiple layers of features from tiny images,'' 2009.

\bibitem{deng2009imagenet}
J.~Deng, W.~Dong, R.~Socher, L.-J. Li, K.~Li, and L.~Fei-Fei, ``Imagenet: A large-scale hierarchical image database,'' in \emph{2009 IEEE conference on computer vision and pattern recognition}.\hskip 1em plus 0.5em minus 0.4em\relax Ieee, 2009, pp. 248--255.

\bibitem{KaimingHe2016IdentityMI}
K.~He, X.~Zhang, S.~Ren, and J.~Sun, ``Identity mappings in deep residual networks,'' in \emph{Computer Vision--ECCV 2016: 14th European Conference, Amsterdam, The Netherlands, October 11--14, 2016, Proceedings, Part IV 14}.\hskip 1em plus 0.5em minus 0.4em\relax Springer, 2016, pp. 630--645.

\bibitem{maas2011learning}
A.~Maas, R.~E. Daly, P.~T. Pham, D.~Huang, A.~Y. Ng, and C.~Potts, ``Learning word vectors for sentiment analysis,'' in \emph{Proceedings of the 49th annual meeting of the association for computational linguistics: Human language technologies}, 2011, pp. 142--150.

\bibitem{zhang2015character}
X.~Zhang, J.~Zhao, and Y.~LeCun, ``Character-level convolutional networks for text classification,'' \emph{Advances in neural information processing systems}, vol.~28, 2015.

\bibitem{vaswani2017attention}
A.~Vaswani, N.~Shazeer, N.~Parmar, J.~Uszkoreit, L.~Jones, A.~N. Gomez, {\L}.~Kaiser, and I.~Polosukhin, ``Attention is all you need,'' \emph{Advances in neural information processing systems}, vol.~30, 2017.

\bibitem{devlin2019bert}
J.~Devlin, M.-W. Chang, K.~Lee, and K.~Toutanova, ``Bert: Pre-training of deep bidirectional transformers for language understanding,'' in \emph{Proceedings of the 2019 Conference of the North American Chapter of the Association for Computational Linguistics: Human Language Technologies, Volume 1 (Long and Short Papers)}, 2019, pp. 4171--4186.

\bibitem{minka2000estimating}
T.~Minka, ``Estimating a dirichlet distribution,'' 2000.

\bibitem{dai2023chameleon}
Y.~Dai and S.~Li, ``Chameleon: Adapting to peer images for planting durable backdoors in federated learning,'' in \emph{International Conference on Machine Learning}.\hskip 1em plus 0.5em minus 0.4em\relax PMLR, 2023, pp. 6712--6725.

\bibitem{yin2018byzantine}
D.~Yin, Y.~Chen, R.~Kannan, and P.~Bartlett, ``Byzantine-robust distributed learning: Towards optimal statistical rates,'' in \emph{International Conference on Machine Learning}.\hskip 1em plus 0.5em minus 0.4em\relax Pmlr, 2018, pp. 5650--5659.

\bibitem{zhang2022fldetector}
Z.~Zhang, X.~Cao, J.~Jia, and N.~Z. Gong, ``Fldetector: Defending federated learning against model poisoning attacks via detecting malicious clients,'' in \emph{Proceedings of the 28th ACM SIGKDD conference on knowledge discovery and data mining}, 2022, pp. 2545--2555.

\end{thebibliography}

\end{document}